# A Neutrosophic Recommender System for Medical Diagnosis Based on Algebraic Neutrosophic Measures

**Mumtaz Ali [1], Nguyen Van Minh [2], Le Hoang Son [3 *]**

[1] Department of Mathematics, Quaid-i-azam University Islamabad, 45320, Pakistan

E-mail: mumtazali7288@gmail.com

[2] Faculty of Science, Hanoi University of Natural Resources and Environment, Vietnam

E-mail: nvminh@hunre.edu.vn

[3] VNU University of Science, Vietnam National University, Vietnam

E-mail: sonlh@vnu.edu.vn

*: Corresponding author. Official address: 334 Nguyen Trai, Thanh Xuan, Hanoi, Vietnam

**Abstract:** Neutrosophic set has the ability to handle uncertain, incomplete, inconsistent, indeterminate information in a more accurate way. In this paper, we proposed a neutrosophic recommender system to predict the diseases based on neutrosophic set which includes single-criterion neutrosophic recommender system (SC-NRS) and multi-criterion neutrosophic recommender system (MC-NRS). Further, we investigated some algebraic operations of neutrosophic recommender system such as union, complement, intersection, probabilistic sum, bold sum, bold intersection, bounded difference, symmetric difference, convex linear sum of min and max operators, Cartesian product, associativity, commutativity and distributive. Based on these operations, we studied the algebraic structures such as lattices, Kleen algebra, de Morgan algebra, Brouwerian algebra, BCK algebra, Stone algebra and MV algebra. In addition, we introduced several types of similarity measures based on these algebraic operations and studied some of their theoretic properties. Moreover, we accomplished a prediction formula using the proposed algebraic similarity measure. We also proposed a new algorithm for medical diagnosis based on neutrosophic recommender system. Finally to check the validity of the proposed methodology, we made experiments on the datasets Heart, RHC, Breast cancer, Diabetes and DMD. At the end, we presented the MSE and computational time by comparing the proposed algorithm with the relevant ones such as ICSM, DSM, CARE, CFMD, as well as other variants namely Variant 67, Variant 69, and Varian 71 both in tabular and graphical form to analyze the efficiency and accuracy. Finally we analyzed the strength of all 8 algorithms by ANOVA statistical tool.

## 1. Introduction

Medical diagnosis is process of investigation of a person's symptoms on the basis of diseases. From modern medical technology, a large amount of information available to medical experts due to whom medical diagnosis contains uncertain, inconsistent, indeterminate information and this information are mandatory in medical diagnosis. A characterized relationship between a symptom and a disease is usually based on these uncertain, inconsistent information which leads to us for decision making in a medical diagnosis. Mostly diagnosis problems have pattern recognition on the basis of which medical experts make their decision. Medical diagnosis has successful practical applications in several areas such as telemedicine, space medicine and rescue services etc. where access of human means of diagnosis is a difficult task. Thus, starting from the early time of Artificial Intelligence, medical diagnosis has got full attention from both computer science and computer applicable mathematics research society. In this regard, Kononenko [27] in 2001 proposed a process of medical diagnosis which is based on the probability or risk of a person who has a particular state of health in a specific time frame. This type of medical diagnosis is advantageous to reduce health problems such as bowel cancer, osteoporosis but on the other hand it can increase the ratio to the risk of other autoimmune problems and diseases. Davis et al., [10] conducted the study to predict individual diseases risk which is based on medical history. Their methodology depends on patient's clinical history to predict the possible risk of the disease. For this purpose, they used collaborative filtering and clustering algorithms to predict patient diseases based on their clinical history. In 1976, Sanchez [40] applied successfully methods of resolution of the fuzzy relations to the medical diagnosis problems which was further extended by De et al., [12] in 2001. This approach is highly relied on defuzzification method through which the most suitable disease can be determined. Treasure [52] in 2011 studied diagnosis and risk management in care. In the same year Kala et al. [24] carried out the research on diagnosis of breast cancer using modular evolutionary neural networks. Johnson et al. [23] discussed expertise and error in diagnostic problems in 1981. Mahdavi [29] in 2012 used recommender system for medial recognition and treatment. Parthiban and Subramanian [34] used CANFIS and genetic algorithm to construct a prediction system for heart disease in 2008. Similarly, Tan et al. [51] in 2003 applied evolutionary computing for knowledge discovery in medical diagnosis. Some theories on medical diagnosis and their applications can be referenced in [8, 11, 22, 30, 32, 33, 39, 41, 48, 49, 50, 56].

A medical diagnostic problem often contains a huge amount of uncertain, inconsistent, incomplete, indeterminate data which is very difficult in retrieving. Neutrosophic Set proposed by Smarandache [42] in 1998 can handle this type of information very accurately. A neutrosophic set can be characterized independently by a truth membership function, indeterminate membership function and false membership function respectively. Recently, Ye [59] applied improved cosine similarity measures of simplified neutrosophic set (subclass of neutrosophic set) in medical diagnosis. In other papers [58, 59, 60, 61, 62, 63, 64], Ye et al. applied neutrosophic sets to medical diagnosis problem. Their approaches were based on dice similarity, distance-based similarity, vector similarity, tangent similarity measures and trapezoidal numbers of neutrosophic sets. In 2014, Broumi and Deli [5] studied the applicability of neutrosophic refined sets through correlation in medical diagnosis problem. Kharal [26] extended the approach of Sanchez to neutrosophic set. Further Broumi and Smarandache [6] in 2015 took the approach of extended Hausdorff distance and similarity measure of refined neutrosophic sets with possible applications in medical diagnosis. Pramanik and Mondal [35] in 2015 studied rough neutrosophic sets with its applications in medical diagnosis. Guo et al. [18] studied neutrosophic sets in lung segmentation for image analysis in thoracic computed tomography in 2013. Some more applications of neutrosophic set in medical diagnosis can be referenced in [1, 7, 65, 66].

The previous literature shows that the focusing aspect is to determine the relationship of patients and diseases by considering symptoms and diseases as well as patients and symptoms in [26]. But this is not the case which always true as these are sometimes missing. Ignoring the history of patient diagnosis is another drawback of the previous approaches. The previous work depends on deneutrosophication process, similarity measures [59, 60, 61, 62, 63], correlation coefficients [5, 20, 21, 58], distance measure [6] etc. However, mathematical properties such as distance measure [6], similarity measures [59, 60, 61, 62, 63] etc. were discussed in the previous work but it was not explained why these operations have been taken in the medical diagnosis problem. The previous discussed neutrosophic methods did not provided accurate information in medical diagnosis.

From the above discussion, it is clear that there should be an appropriate methodology which can handle these pointed out issues in medical diagnosis. The purpose of this article is to introduce a new hybrid structure which based on the neutrosophic set and recommender systems. Recommender systems are basically decision support systems which provide a recommendation by decreasing information overload. Yager [57] in 2003 derived the distinction between the recommender system and targeted marketing. Recommender systems are attractive computer-based techniques which are highly applicable in several interdisciplinary fields which provide predictive rating or preference to select the most suitable item among others. Park et. al. [67] in 2012 conducted a survey on the applicability of Recommender Systems in books, images, documents, music, movies, shopping, and TV programs. Ghazanfar and Prügel-Bennett [17] derived a hybrid recommendation algorithm for gray-sheep users to reduce the error rate by maintaining computational performance. Davis et al. [10] developed a Recommendation Engine which uses patient medical history to predict future diseases risk. Hassan and Syed [22] build a combined filtering technique which evaluated the patient risk by comparing new and historical records and patient demographics. Duan et. al. [11] in 2011 discussed the plans of nursing care in a healthcare Recommender System. A huge amount of literature of the applications of recommender systems in medical diagnosis can be seen in [9, 11, 29, 43, 44, 45, 46, 47]. Therefore, recommender systems can be used successfully to predict patient diseases on historic records.

Motivated from the previous issues and the applications of neutrosophic sets in medical diagnosis, we observe that the neutrosophic recommender systems (NRS) could successfully handle all the pointed out issues of previous work. We focused the following areas in this article.

- First, we proposed the single-criterion neutrosophic recommender system (SC-NRS) and multi-criterion neutrosophic recommender system (MC-NRS).
- We investigated some algebraic operations of neutrosophic recommender system NRS such as union, complement, intersection, probabilistic sum, bold sum, bold intersection, bounded difference, symmetric difference, convex linear sum of min and max operators, and the Cartesian product. We gave explanation of these algebraic properties with illustrative examples.
- Moreover, we discussed the algebraic properties such associativity, commutativity and distributive of neutrosophic recommender system NRS and we discussed some theoretic properties of these operations and explain them with illustrative examples.

- In addition, we studied the algebraic structures such as lattices, Kleen algebra, de Morgan algebra, Brouwerian algebra, BCK algebra, Stone algebra and MV algebra. We also studied some theoretic properties of algebraic structures and explain them with illustrative examples.
- Further, we introduced several types of similarity measures of single-criteria neutrosophic system (SC-NRS) and multi-criteria neutrosophic recommender system (MC-NRS). These similarity measures based on the algebraic operations discussed in section 2.3. We also studied some theoretic properties of these similarity measures and explain them with illustrative examples.
- We also accomplished the formula on the basis of similarity measure for prediction.
- For medical diagnosis, we proposed a new algorithm based on neutrosophic recommender system.
- To check the validity of the proposed methodology, we made experiments on the datasets Heart, RHC, Breast cancer, Diabetes and DMD.
- Finally, we presented the MSE and computational time (Sec.) by comparing the proposed algorithm with ICSM, DSM, CARE, CFMD, Variant 67, Variant 69, and Varian 71 both in tabular and graphical form to analyze the efficiency and accuracy. We also analyzed the strength of all 8 algorithms by ANOVA test.

From the contributions and disadvantages of the previous research work, Neutrosophic Recommender System (NRS) has the following novelty and significance.

- The notions of Neutrosophic Recommender System (NRS) are completely distinct in the sense that Neutrosophic Recommender System (NRS) shares both the integrated features and characteristics of Neutrosophic Set (NS) and Recommender System (RS). This hybridization can be clearly seen in single-criterion neutrosophic recommender system (SC-NRS) and multi-criterion neutrosophic recommender system (MC-NRS). Both the (SC-NRS) and (MC-NRS) became the foundation of the neutrosophic collaborative filtering method (NCF). Moreover, Ye [59, 60, 61, 62], Broumi and Deli [5], Kharal [26], Broumi and Smarandache [6] and Pramanik and Mondal [35] used distance-based similarity measures to calculate the relationship of the patients and the diseases. But we used neutrosophic similarity measures based on algebraic operations to find out this relationship.

- Our proposed hybrid structure can efficiently handle the limitations of neutrosophic set NS concerning this missing information and historic diagnosis of the medical patient. It also covers the issues of crisp and training dataset in a Recommender System (NRS).
- The other significance of Neutrosophic Recommender System (NRS) is its generalization to all other currently existing Hybrid Recommender Systems. This novelty of Neutrosophic Recommender System (NRS) could improve accuracy of the related neutrosophic methods for medical diagnosis by a hybrid method with recommender systems.
- The significance and importance of the proposed work can be seen from both the theoretical and practical aspects. It can increase the accuracy of the algorithms of neutrosophic set (NS) as well as recommender system (RS). The proposed hybrid structure (NRS) based on a strong mathematical foundation which lacks in the previous work. On the other hand, this work can contribute to medical diagnosis as well as some other extended areas of relevant applications.
- The significance can also be seen in the construction of several types of algebras of neutrosophic recommender system studied in section 3.3.
- The novelty of neutrosophic recommender systems can also be seen in the similarity measures. These similarity measures are mainly based on the algebraic operations discussed in section 3.2.
- The significance of the proposed algorithm can be seen over the traditional ones on different types of medical datasets.

## 2. Background

In this section, we discussed the literature review about medical diagnosis, neutrosophic set; medical diagnosis based on neutrosophic set, recommender system and studied some of their basic properties which will be used in our later pursuit.

### 2.1 Literature Review

To handle uncertain, incomplete, vague, inconsistent, indeterminate information, Smarandache [42] initiated the concept of Neutrosophic sets in 1998 which is basically inspired by Neutrosophic philosophy. A neutrosophic set has three independent components association (truth membership) degree, non-association (false membership) degree and indeterminacy degree. The potential applications of neutrosophic sets can be seen in decision making theory [13, 37, 65, 66], relational database [54, 55], pattern recognition [2], image analysis [28], signal processing [3] and so on. In medical diagnosis, the neutrosophic sets also played a significant role. Broumi and Deli [5] showed the applicability of neutrosophic refined sets through correlation in medical diagnosis. Recently, Ye and Fu [62] studied multi-period medical diagnosis problems by applying neutrosophic sets. Applicability of neutrosophic set in medical AI has been discussed by Ansari et al. [4] in 2011. In some more paper, Ye et al. [59, 60, 61, 63, 64] applied neutrosophic sets to medical diagnosis problem. Further, (Kandasamy and Smarandache [25] in 2004 applied neutrosophic relational maps to HIV/AIDS. Gaber et al. [15] in 2015 discussed thermogram breast cancer detection based on neutrosophic set. Similarly, Kharal [26], Broumi and Smarandache [6] and Pramanik and Mondal [35] used distance-based similarity measures to calculate the relationship of the patients and the diseases. Guerram et al. [19] applied neutrosophic cognitive maps to viral infection. Pramanik and Mondal [35] studied rough neutrosophic sets with its applications in medical diagnosis. Guo et al. [18] in 2013 studied neutrosophic sets in lung segmentation for image analysis in thoracic computed tomography. In 2013, Mohan et al. [31] introduced a new filtering technique based on neutrosophic set for MRI de-noising. Some theory of neutrosophic set in medical diagnosis can be referenced in [1, 7, 65, 66]. Medical diagnosis of neutrosophic set can be explained in Example 1 by using distance-based similarity measures.

**Table 1. Relation between patients and the symptoms-** $\Re_{\wp S}$

| | Temperature ($s_1$) | Cough ($s_2$) | Throat pain ($s_3$) | Headache ($s_4$) | Body pain ($s_5$) |
|---|---|---|---|---|---|
| $p_1$ | <(0.8, 0.6, 0.5), (0.3, 0.2, 0.1), (0.4, 0.2, 0.1)> | <(0.5, 0.4, 0.3), (0.4, 0.4, 0.3), (0.6, 0.3, 0.4)> | <(0.2, 0.1, 0.0), (0.3, 0.2, 0.2), (0.8, 0.7, 0.7)> | <(0.7, 0.6, 0.5), (0.3, 0.2, 0.1), (0.4, 0.3, 0.2)> | <(0.4,0.3, 0.2), (0.6, 0.5, 0.5), (0.6, 0.4, 0.4)> |
| $p_2$ | <(0.5, 0.4, 0.3), (0.3, 0.3, 0.2), (0.5, 0.4, 0.4)> | <(0.9, 0.8, 0.7), (0.2, 0.1, 0.1), (0.2, 0.2, 0.1)> | <(0.6, 0.5, 0.4), (0.3, 0.2, 0.2), (0.4, 0.3, 0.3)> | <(0.6, 0.4, 0.3), (0.3, 0.1, 0.1), (0.7, 0.7, 0.3)> | <(0.8,0.7, 0.5), (0.4, 0.3, 0.1), (0.3, 0.2, 0.1)> |
| $p_3$ | <(0.2, 0.1, 0.1), (0.3, 0.2, 0.2), (0.8, 0.7, 0.6)> | <(0.3, 0.2, 0.2), (0.4, 0.2, 0.2), (0.7, 0.6, 0.5)> | <(0.8, 0.8, 0.7), (0.2, 0.2, 0.2), (0.1, 0.1, 0.0)> | <(0.3, 0.2, 0.2), (0.3, 0.3, 0.3), (0.7, 0.6, 0.6) | <(0.4,0.4, 0.3), (0.4, 0.3, 0.2), (0.7, 0.7, 0.5)> |
| $p_4$ | <(0.5, 0.5, 0.4), (0.3, 0.2, 0.2), (0.4, 0.4, 0.3)> | <(0.4,0.3, 0.1), (0.4, 0.3, 0.2), (0.7, 0.5, 0.3)> | <(0.2,0.1, 0.0), (0.4, 0.3, 0.3), (0.7, 0.7, 0.6)> | <(0.6, 0.5, 0.3), (0.2, 0.2, 0.1), (0.6, 0.4, 0.3)> | <(0.5,0.4, 0.4), (0.3, 0.3, 0.2), (0.6, 0.5, 0.4)> |

**Example 1:** Consider the dataset from [60]. Let $\wp = \{p_1, p_2, p_3, p_4\}$ be a set of four patients, $D = \{$Viral fever, Tuberculosis, Typhoid, Throat disease$\}$, and $S = \{$Temperature, Cough, Throat pain, Headache, Body pain$\}$ be a set of symptoms respectively which are illustrated in **Table 1** and **Table 2.** The relationship between patients and symptoms is shown in **Table 1**, while **Table 2** expressed the relationship of symptoms and diseases. The values obtained in **Table 3** are the largest similarity measure which indicates the proper diagnose of the patient.

**Table 2. Relation between the symptoms and the diseases-** $\Re_{SD}$

| | Temperature ($s_1$) | Cough ($s_2$) | Throat pain ($s_3$) | Headache ($s_4$) | Body pain ($s_5$) |
|---|---|---|---|---|---|
| Viral fever ($d_1$) | <0.8, 0.1, 0.1> | <0.2,0.7,0.1> | <0.3, 0.5, 0.2> | (0.5, 0.3, 0.2) | <0.5,0.4,0.1> |
| Tuberculosis ($d_2$) | <0.2, 0.7, 0.1> | <0.9,0.0,0.1> | <0.7, 0.2, 0.1> | (0.6, 0.3, 0.1) | <0.7,0.2,0.1> |
| Typhoid ($d_3$) | <0.5, 0.3, 0.2> | <0.3,0.5,0.2> | <0.2, 0.7, 0.1> | (0.2, 0.6, 0.2) | <0.4,0.4,0.2> |
| Throat disease($d_4$) | <0.1, 0.7, 0.2> | <0.3,0.6,0.1> | <0.8, 0.1, 0.1> | (0.1, 0.8, 0.1) | <0.1,0.8,0.1> |

**Table 3. The relation between the patients and the diseases by distance-based similarity measure**

| | Viral fever ($d_1$) | Tuberculosis ($d_2$) | Typhoid ($d_3$) | Throat disease($d_4$) |
|---|---|---|---|---|
| $p_1$ | 0.7358 | 0.6101 | 0.7079 | 0.5815 |
| $p_2$ | 0.6884 | 0.7582 | 0.6934 | 0.5964 |
| $p_3$ | 0.6159 | 0.6141 | 0.6620 | 0.6294 |
| $p_4$ | 0.7199 | 0.6167 | 0.7215 | 0.5672 |

**2.2 Medical Diagnosis**

**Definition 1 [27]:** Let $\wp=\{p_1,p_2,...,p_n\}$, $\Gamma=\{s_1,s_2,...,s_m\}$ and $D=\{d_1,d_2,...,d_k\}$ be three lists of patients, symptoms and diseases, respectively such that $n,m,k\in N^+$ be the numbers of patients, symptoms and diseases respectively. Let $\Re_{\wp\Gamma}=\left\{\Re^{\wp\Gamma}\left(p_i,s_j\right):\forall i=1,2,..,n;\ j=1,2,...,m\right\}$ be the set the relation between patients and symptoms where $\Re^{\wp\Gamma}\left(p_i,s_j\right)$ is the level of the patient $p_i$ who acquires the symptom $s_j$. The value of $\Re^{\wp\Gamma}\left(p_i,s_j\right)$ is either numeric number or a neutrosophic number which depends on the proposed domain of the problem. Similarly, $\Re_{\Gamma D}=\left\{\Re^{\Gamma D}\left(s_i,d_j\right):\forall i=1,2,..,\mathrm{m};\ j=1,2,...,\mathrm{k}\right\}$ be the set which represents the relation between the symptoms and the diseases where $\Re^{\Gamma D}\left(s_i,d_j\right)$ reveals the possibility of the symptom $s_i$ leads to the disease $d_j$. The purpose of the medical diagnosis is to determine the relationship between the patients and the diseases and this can be described as $\Re_{\wp D}=\left\{\Re^{\wp D}\left(p_i,d_j\right):\forall i=1,2,..,\mathrm{m};\ j=1,2,...,\mathrm{k}\right\}$ where the value of $\Re^{\wp D}\left(p_i, d_j\right)$ is either 0 or 1 which demonstrate that the patient $p_i$ acquired the disease $d_j$ or not. Mathematically the problem of medical diagnosis is an implication operator given by the map $\left\{\Re_{\wp\Gamma}, \Re_{\Gamma D}\right\} \to \Re_{\wp D}$.

### 2.3 Neutrosophic Set NS and Simplified Neutrosophic Set SNS

**Definition 2 [42]:** Let $X$ be a non-empty set and $x \in X$. A neutrosophic set $A$ in $X$ is characterized by a truth membership function $T_A$, an indeterminacy membership function $I_A$, and a falsehood membership function $F_A$. Here $T_A(x)$, $I_A(x)$ and $F_A(x)$ are real standard or non-standard subsets of $\left]0^-, 1^+\right[$ such that $T_A, I_A, F_A : X \to \left]0^-, 1^+\right[$. There is no restriction on the sum of $T_A(x), I_A(x)$ and $F_A(x)$, so, $0^- \le T_A(x) + I_A(x) + F_A(x) \le 3^+$. From philosophical point view, the neutrosophic set takes the value from real standard or non-standard subsets of $\left]0^-, 1^+\right[$. Thus it is necessary to take the interval $[0,1]$ instead of $\left]0^-, 1^+\right[$ for technical applications because it is difficult to use $\left]0^-, 1^+\right[$ in the real life applications such as engineering and scientific problems.

If the functions $T_A(x)$, $I_A(x)$ and $F_A(x)$ are singleton subinterval/subsets of the real standard such that with $T_A(x) : \mathrm{X} \to [0,1], \mathrm{I}_A(x) : \mathrm{X} \to [0,1], F_A(x) : X \to \in [0,1]$. Then a simplification of the neutrosophic set $A$ is denoted by

$$A = \left\{\left(x, T_A(x), I_A(x), F_A(x)\right) : x \in X\right\} \qquad (1)$$

with $0 \le T_A(x) + I_A(x) + F_A(x) \le 3$. It is a subclass of neutrosophic set and called simplified neutrosophic set. A simplified neutrosophic set SNS [59] contains the concept of interval neutrosophic set INS [54], and single valued neutrosophic set SVNS [55]. In our paper, we will use simplified neutrosophic set.

Some operations of NS are defined as follows: For two NS

$A_1 = \{\langle x; T_1(x); I_1(x); F_1(x)\rangle \mid \mathrm{x} \in \mathrm{X}\}$ and $\mathrm{A}_2 = \{\langle x; T_2(x); I_2(x); F_2(x)\rangle \mid \mathrm{x} \in \mathrm{X}\}$

1. $A_1 \subseteq A_2$ if and only if $T_1(x) \le T_2(x); I_1(x) \ge I_2(x); F_1(x) \ge F_2(x)$ ;

2. $A_1 = A_2$ if and only if $T_1(x) = T_2(x); I_1(x) = I_2(x); F_1(x) = F_2(x)$ ;

3. $A_1^c = \{\langle x; F_1(x); I_1(x); T_1(x) \rangle \mid x \in X\}$;

4. $A_1 \cap A_2 = \{\langle x; \min\{T_1(x); T_2(x)\}; \max\{I_1(x); I_2(x)\}; \max\{F_1(x); F_2(x)\} \rangle \mid x \in X\}$;

5. $A_1 \cup A_2 = \{\langle x; \max\{T_1(x); T_2(x)\}; \min\{I_1(x); I_2(x)\}; \min\{F_1(x); F_2(x)\} \rangle \mid x \in X\}$.

### 2.3.1: Neutrosophication and Deneutrosophication Process

**Definition 3** [54]: **Neutrosophication**

The main purpose of neutrosophication is to map input variables into neutrosophic input sets. If $x$ is a crisp input, then

$$T(x) = \begin{cases} \dfrac{x - a_1}{a_2 - a_1}, & a_1 \le x < a_2, \\ \dfrac{a_3 - x}{a_3 - a_2}, & a_2 \le x < a_3, \\ \dfrac{x - a_3}{a_4 - a_3}, & a_3 \le x < a_4 \\ 0, & \text{otherwise}. \end{cases} \tag{1a}$$

$$I(x) = \begin{cases} \dfrac{b_2 - x}{b_2 - b_1}, & b_1 \le x < b_2, \\ \dfrac{b_3 + x - b_2}{b_3 - b_2}, & b_2 \le x < b_3, \\ \dfrac{b_3 + b_4 - x}{b_4 - b_3}, & b_3 \le x < b_4, \\ 1, & \text{otherwise}. \end{cases} \tag{2a}$$

$$F(x)=\begin{cases}\dfrac{c_2-x}{c_2-c_1}, & c_1\le x<c_2,\\ \dfrac{x}{c_3} & c_2\le x<c_3,\\ \dfrac{c_4+c_3-x}{c_4-c_3}, & c_3\le x<c_4\\ 1, & \text{otherwise}.\end{cases} \qquad (3a)$$

Where $x\in X$ and $a_j\le x\le a_k$ for truth membership, $b_j\le x\le b_k$ for indeterminacy membership and $c_j\le x\le c_k$ for falsehood membership respectively and $j,k=1,2,3,4$.

**Definition 4 [54]: Deneutrosophication**

This step is similar to defuzzification of George and Bo **[16]** in 1995. This step involves in the following two stages:

**Stage 1: Synthesization**

In this stage, we transform a neutrosophic set $H^k$ in to fuzzy set $\mathrm{B}$ by the following function:

$$f\left(T_{H^k}(y),I_{H^k}(y),F_{H^k}(y)\right):[0,1]\times[0,1]\times[0,1]\rightarrow[0,1] \qquad (1b)$$

Here f is defined by

$$T_B(y)=\alpha*T_{H^k}(y)+\beta*\frac{F_{H^k}(y)}{4}+\gamma*\frac{I_{H^k}(y)}{2} \qquad (2b)$$

Where $0\le\alpha,\beta,\gamma\le 1$ such that $\alpha+\beta+\gamma=1$.

**Stage 2: Typical neutrosophic value**

In this stage, we can calculate a typical deneutrosophicated value $den\left(T_{\mathrm{B}}(y)\right)$ by the centroid or center of gravity method which is given below:

$$den\left(T_{\mathrm{B}}(y)\right)=\frac{\int_a^b T_{\mathrm{B}}(y)\,y\,dy}{\int_a^b T_{\mathrm{B}}(y)\,dy} \tag{3b}$$

**2.4 Recommender Systems RS**

**Definition 5 [38]:** Single-criteria recommender Systems (SC-RS)

Suppose U is a set of all users and $\Omega$ is the set of items in the system. The utility function $\Re$ is a mapping specified on $U_1 \subset U$ and $\Omega_1 \subset \Omega$ as follows:

$$\begin{aligned}&\Re : U_1 \times \Omega_1 \to \wp\\&(u_1;\omega_1) \mapsto \Re(u_1;\omega_1)\end{aligned} \tag{2}$$

where $\Re(u_1;\omega_1)$ is a non-negative integer or a real number within a certain range. $\Re$ is a set of available ratings in the system. Thus, RS is the system that provides two basic functions below.

(a) Prediction: determine $\Re(u^*;\omega^*)$ for any $(u^*,\omega^*) \in (U,\Omega) \setminus (U_1;\Omega_1)$ ;

(b) Recommendation: choose $\omega^* \in \Omega$ satisfying $\omega^* = \arg\max_{i\in I} \Re(u,\omega)$ for all u $\in U$

**Definition 6 [38]:** Multi-criteria recommender Systems (MC-RS)

MCRS are the systems providing similar basic functions with RS but following by multiple criteria. In the other words, the utility function is defined below

$$\begin{aligned}&\Re : U_1 \times \Omega_1 \to \wp_1 \times \wp_2 \times \ldots \times \wp_k,\\&(u_1;\omega_1) \mapsto (\Re_1, \Re_2, \ldots, \Re_k)\end{aligned} \tag{3}$$

where $\Re_i\,(i=1,2,\ldots,k)$ is the rating of user $u_1 \in U_1$ for item $\omega_1 \in \Omega_1$ following by criteria i in this case, the recommendation is performed according to a given criteria.

**Example 2.** Suppose that U = {John, David, Jenny, Marry} and I = {Titanic, Hulk, Scallet}. The set of criteria of a movie is $\wp$ = {Story, Visual effects}. The ratings are assigned numerically from 1 (bad) to 5 (excellent). **Table 4** describes the utility function. From this table, it is clear that MCRS can help us to predict the ratings of users (Marry) to a movie that was not rated by her beforehand (Titanic). This kind of systems also recommends her favorite movie through available ratings. In cases that there is only a criterion in P, MCRS returns to the traditional RS.

**Table 4. Movies rating**

| User | Movie | Story | Visual effects |
|---|---|---|---|
| **John** | Hulk | 4 | 3 |
| **John** | Scallet | 2 | 2 |
| **David** | Titanic | 4 | 2 |
| **David** | Hulk | 3 | 1 |
| **David** | Scallet | 1 | 4 |
| **Jenny** | Hulk | 2 | 3 |
| **Jenny** | Titanic | 1 | 2 |
| **Marry** | Hulk | 3 | 5 |
| **Marry** | Titanic | ? | ? |

## 3. Neutrosophic Recommender System NRS

In this section, we introduced single-criteria neutrosophic recommender system (SC-NRS) and multi-criteria neutrosophic recommender system (MC-NRS). Further, we introduced some algebraic operations of neutrosophic recommender system as well as algebraic structures (algebras). Finally, we presented some similarity measures based on these algebraic operations.

### 3.1: Single-criteria Neutrosophic Recommender System (SC-NRS) and Multi-criteria Neutrosophic Recommender System (MC-NRS)

Let $\wp = \{p_1, p_2, ..., p_n\}$, $\Gamma = \{s_1, s_2, ..., s_m\}$ and $D = \{d_1, d_2, ..., d_k\}$ be three lists of patients, symptoms and diseases, respectively such that $n, m, k \in N^+$ be the numbers of patients, symptoms and diseases respectively where $p_i$ and $s_j$ have some features and characteristics respectively such that $i = 1, 2, ..., n$ and $j = 1, 2, ..., m$. Further, we consider that the features of the patient and characteristics of the symptoms are denoted by $\mathrm{X}$ and $\Upsilon$ which consist of $s$ neutrosophic linguistic labels. Similarly, disease $d_i$ also has $s$ neutrosophic linguistic labels where $i = 1, 2, ..., k$.

**Definition 7:** Single-criteria Neutrosophic recommender System (SC-NRS)

The (SC-NRS) is a utility function $\Re$ which is a mapping defined on $(\mathrm{X}, \Upsilon)$ as follows:

$$\Re : \mathrm{X} \times \Upsilon \to D$$

$$\left\langle \begin{array}{l} (T_{1\mathrm{X}}(x), I_{1\mathrm{X}}(x), F_{1\mathrm{X}}(x)), \\ (T_{2\mathrm{X}}(x), I_{2\mathrm{X}}(x), F_{2\mathrm{X}}(x)), \\ \cdots \\ (T_{s\mathrm{X}}(x), I_{s\mathrm{X}}(x), F_{s\mathrm{X}}(x)) \end{array} \right\rangle \times \left\langle \begin{array}{l} (T_{1\Upsilon}(y), I_{1\Upsilon}(y), F_{1\Upsilon}(y)), \\ (T_{2\Upsilon}(y), I_{2\Upsilon}(y), F_{2\Upsilon}(y)), \\ \cdots \\ (T_{s\Upsilon}(y), I_{s\Upsilon}(y), F_{s\Upsilon}(y)) \end{array} \right\rangle \to \left\langle \begin{array}{l} (T_{1D}(d), I_{1D}(d), F_{1D}(d)), \\ (T_{2D}(d), I_{2D}(d), F_{2D}(d)), \\ \cdots \\ (T_{sD}(d), I_{sD}(d), F_{sD}(d)) \end{array} \right\rangle, \quad (4)$$

where $T_{i\mathrm{X}}(x), I_{i\mathrm{X}}(x), F_{i\mathrm{X}}(x)$ is the truth membership function, indeterminate membership function and false membership function of the patient to the linguistic label $ith$ of the feature $\mathrm{X}$ such that $i = 1, 2, ..., s$ and $T_{i\mathrm{X}}(x), I_{i\mathrm{X}}(x), F_{i\mathrm{X}}(x) \in [0,1]$. Similarly, $T_{j\Upsilon}(y), I_{j\Upsilon}(y), F_{j\Upsilon}(y)$ is the truth membership function, indeterminate membership function and false membership function of the symptom to the linguistic label $jth$ of the feature $\Upsilon$ where $j = 1, 2, ..., s$ and $T_{j\Upsilon}(y), I_{j\Upsilon}(y), F_{j\Upsilon}(y) \in [0,1]$. Additionally, $T_{lD}(d), I_{lD}(d), F_{lD}(d)$ is the truth membership function, indeterminate membership function and false membership function of the disease $D$ to the linguistic label $lth$ where such that $l = 1, 2, ..., s$ and $T_{lD}(d), I_{lD}(d), F_{lD}(d) \in [0,1]$.

The single-criteria neutrosophic recommender system (SC-NRS) is potentially applicable in the real life to explain the uncertainties, ambiguities, indeterminacies, incompleteness, falsities etc. of choices of human decision. For illustration, consider **Table 4** of Movies rating of different viewers. John comments about the story and visual effect of the movie hulk can be bitterly captured by applying (SC-NRS). For example John is in favor of story about 0.8 but he may be disagreeing or undecided up to some extent say 0.5 and 0.4 respectively. Similarly he likes the visual effect of the movie up to 0.6 but he may be dislike the visual effect or unsure due to number of reasons such as low quality of visual effect tools, print, graphics, mood, surrounding environment etc. However, in this type of complicated situations, the (SC-NRS) are extremely beneficial to represent the human decision accurately.

SC-NRS depicts the following assertions:

1. Prediction: Compute the values of $\left(T_{lD}(d), I_{lD}(d), F_{lD}(d)\right)$ for all $l = 1, 2, ..., s$.

2. Recommendation 1: select $\overset{*}{i} \in [1, s]$ which satisfies

$$\overset{*}{i} = \arg\max_{i=1}^{s}\left\{T_{lD}(d) + T_{lD}(d)\left(3 - T_{lD}(d) - I_{lD}(d) - F_{lD}(d)\right)\right\}.$$

3. Recommendation 2: select $\overset{*}{i} \in [1, s]$ which satisfies

$$\overset{*}{i} = \arg\max_{i=1}^{s}\left\{T_{lD}(d) + T_{lD}(d)\left(2\left(1 - T_{lD}(d)\right) - I_{lD}(d) - F_{lD}(d)\right)\right\}.$$

The formula of recommendation 1 and recommendation 2 gives different results for some $\overset{*}{i} \in [1, s]$. This means the neutrosophic recommender system the ability of depicting more than one choice for recommendation.

**Remark 1:**

a) It is clear from Definition 7 and Eq. 4 that the medical diagnosis which is denoted by the implication $(Patient, Symptom) \rightarrow Disease$ identical to the Definition 1. Consequently, The SC-NRS is another form of medical diagnosis which follows the philosophy of recommender system RS.

b) SC-NRS in Definition 7 could be seen as the extension of the RS in Definition 5 in the following cases:

- There exist $i; T_{i\mathrm{X}}(x) = 1 \wedge I_{i\mathrm{X}}(x) \wedge F_{i\mathrm{X}}(x) = 0;\ \forall j \neq i; T_{j\mathrm{X}}(x) = 0 \wedge I_{j\mathrm{X}}(x) \wedge F_{j\mathrm{X}}(x) = 1,$
- There exist $i; T_{i\Upsilon}(y) = 1 \wedge I_{i\Upsilon}(y) \wedge F_{i\Upsilon}(y) = 0;\ \forall j \neq i; T_{j\Upsilon}(y) = 0 \wedge I_{j\Upsilon}(y) \wedge F_{j\Upsilon}(y) = 1,$

- There exist $i;T_{iD}(d)=1\wedge I_{iD}(d)\wedge F_{iD}(d)=0;\ \forall j\neq i;T_{jD}(d)=0\wedge I_{jD}(d)\wedge F_{jD}(d)=1,$
- There exist $i;T_{i\mathrm{X}}(x)=1\wedge I_{i\mathrm{X}}(x)=0;\ \forall j\neq i;T_{j\mathrm{X}}(x)=0\wedge I_{j\mathrm{X}}(x)=1,$
- There exist $i;T_{i\mathrm{X}}(x)=1\wedge F_{i\mathrm{X}}(x)=0;\ \forall j\neq i;T_{j\mathrm{X}}(x)=0\wedge F_{j\mathrm{X}}(x)=1,$
- There exist $i;T_{i\Upsilon}(y)=1\wedge I_{i\Upsilon}(y)=0;\ \forall j\neq i;T_{j\Upsilon}(y)=0\wedge I_{j\Upsilon}(y)=1,$
- There exist $i;T_{i\Upsilon}(y)=1\wedge F_{i\Upsilon}(y)=0;\ \forall j\neq i;T_{j\Upsilon}(y)=0\wedge F_{j\Upsilon}(y)=1,$
- There exist $i;T_{iD}(d)=1\wedge I_{iD}(d)=0;\ \forall j\neq i;T_{jD}(d)=0\wedge I_{jD}(d)=1,$
- There exist $i;T_{iD}(d)=1\wedge F_{iD}(d)=0;\ \forall j\neq i;T_{jD}(d)=0\wedge F_{jD}(d)=1.$

Alternatively, the mapping in Eq. 4 can be written as

$$\Re:\wp\times\Gamma\to D$$

$$\left((p,\mathrm{X}),(s,\Upsilon)\right)\to\Re_{\wp D} \tag{5}$$

Next, we extend single-criterion neutrosophic recommender system SC-NRS to multi-criteria neutrosophic recommender system MC-NRS which can handle multiple diseases $D=\{d_1,d_2,\cdots,d_k\}$.

**Definition 8:** Multi-criteria Neutrosophic recommender System (MC-NRS). The (MC-NRS) is a utility function $\Re$ which is a mapping defined on $(\mathrm{X},\Upsilon)$ as follows:

$$\Re:\mathrm{X}\times\Upsilon\to D_1\times D_2\times\cdots\times D_k$$

$$\left\langle\begin{matrix}(T_{1\mathrm{X}}(x),I_{1\mathrm{X}}(x),F_{1\mathrm{X}}(x)),\\(T_{2\mathrm{X}}(x),I_{2\mathrm{X}}(x),F_{2\mathrm{X}}(x)),\\\cdots\\(T_{s\mathrm{X}}(x),I_{s\mathrm{X}}(x),F_{s\mathrm{X}}(x))\end{matrix}\right\rangle\times\left\langle\begin{matrix}(T_{1\Upsilon}(y),I_{1\Upsilon}(y),F_{1\Upsilon}(y)),\\(T_{2\Upsilon}(y),I_{2\Upsilon}(y),F_{2\Upsilon}(y)),\\\cdots\\(T_{s\Upsilon}(y),I_{s\Upsilon}(y),F_{s\Upsilon}(y))\end{matrix}\right\rangle, \tag{4}$$

$$\to\left\langle\begin{matrix}(T_{1D}(d_1),I_{1D}(d_1),F_{1D}(d_1)),\\(T_{2D}(d_1),I_{2D}(d_1),F_{2D}(d_1)),\\\cdots\\(T_{sD}(d_1),I_{sD}(d_1),F_{sD}(d_1))\end{matrix}\right\rangle\times\left\langle\begin{matrix}(T_{1D}(d_2),I_{1D}(d_2),F_{1D}(d_2)),\\(T_{2D}(d_2),I_{2D}(d_2),F_{2D}(d_2)),\\\cdots\\(T_{sD}(d_2),I_{sD}(d_2),F_{sD}(d_2))\end{matrix}\right\rangle\times\cdots\times\left\langle\begin{matrix}(T_{1D}(d_k),I_{1D}(d_k),F_{1D}(d_k)),\\(T_{2D}(d_k),I_{2D}(d_k),F_{2D}(d_k)),\\\cdots\\(T_{sD}(d_k),I_{sD}(d_k),F_{sD}(d_k))\end{matrix}\right\rangle$$

The multi-criteria neutrosophic recommender systems (MC-NRS) can effectively explain the viewer's ratings about the multiple movies simultaneously in Table 4. John rating of story line about movies Hulk and Scallet could be true up to 0.8 and 0.2 respectively but He may be dislike or undecided up to some extent as human judgement is varying. So in short, the (MC-NRS) can handle the human ratings in an accurate way by considering the truth membership function, indeterminate membership function and falsity membership function.

The MC-NRS defines the following two functions:

1. Prediction: Compute the values of $\left(T_{lD}\left(d_i\right), I_{lD}\left(d_i\right), F_{lD}\left(d_i\right)\right)$ for all $l = 1, 2, ..., s$ and $i = 1, 2, ..., \mathrm{k}$
2. Recommendation 1: select $\overset{*}{i} \in [1, s]$ which satisfies

$$\overset{*}{i} = \arg\max_{i=1}^{s} \left\{ \sum_{j=1}^{k} w_j \left( T_{iD}\left(d_j\right) + T_{iD}\left(d_j\right) \left(3 - T_{iD}\left(d_j\right) - I_{iD}\left(d_j\right) - F_{iD}\left(d_j\right)\right)\right)\right\},$$

where $w_j$ is the weight of $d_j$ which belongs to $[0,1]$ and satisfying the constraint $\sum_{j=1}^{k} w_j = 1$.

3. Recommendation 2: select $\overset{*}{i} \in [1, s]$ which satisfies

$$\overset{*}{i} = \arg\max_{i=1}^{s} \left\{ \sum_{j=1}^{k} w_j \left( T_{iD}\left(d_j\right) + T_{iD}\left(d_j\right) \left(2\left(1 - T_{iD}\left(d_j\right)\right) - I_{iD}\left(d_j\right) - F_{iD}\left(d_j\right)\right)\right)\right\},$$

where $w_j$ is the weight of $d_j$ which belongs to $[0,1]$ and satisfying the constraint $\sum_{j=1}^{k} w_j = 1$.

**Example 3:** Consider 4 patients in a medical diagnosis process who have age (feature) $\mathrm{X}$ which consists of 3 linguistic labels $\{young, middle, old\}$ where $s = 1, 2, 3$. Similarly, the "Temperature" is symptoms (characteristic) $\Upsilon$ which comprise 3 linguistic labels $\{cold, medium, hot\}$. The disease $\{d_1\}$ is $\{Fever\}$ which also comprise 3 linguistic levels $\{L_1, L_2, L_3\}$. We will use the trapezoidal neutrosophic number - $TNN$ [64] to find which age of patient and types of temperature cause the relevant disease. The truth membership functions, indeterminate membership functions and false membership functions respectively of the patients to the linguistic label $ith$ of the age (feature) X are as follows:

$$T_{young}(x)\begin{cases} 1 & if \quad 5 \le x < 25, \\ {45 - x}/{20} & if \quad 25 \le x < 45, \\ 0 & if \quad 45 \le x > 50 \end{cases} \tag{5}$$

$$I_{young}(x)\begin{cases} 0 & if \quad 5 \le x < 25, \\ {35 - x}/{30} & if \quad 25 \le x < 35, \\ 1 & if \quad x > 35, \end{cases} \tag{6}$$

$$F_{young}(x)\begin{cases} 0 & if \quad 5 \le x < 25, \\ {x - 25}/{20} & if \quad 25 \le x < 35, \\ 1 & if \quad x > 35, \end{cases} \tag{7}$$

$$T_{middle}(x)\begin{cases} 0 & if \quad x > 70, \\ {x - 25}/{20} & if \quad 25 < x \le 35, \\ 1 & if \quad 35 \le x < 50, \\ {70 - x}/{20} & if \quad 50 \le x \le 70, \end{cases} \tag{8}$$

$$I_{middle}(x)\begin{cases} 1 & if \quad x > 70, \\ {35 - x}/{20} & if \quad 25 < x \le 35, \\ {70 - x}/{50} & if \quad 35 < x \le 50, \\ 0 & if \quad 50 < x \le 70, \end{cases} \tag{9}$$

$$F_{middle}(x)\begin{cases} 1 & if \quad x > 70, \\ {40 - x}/{20} & if \quad 25 < x \le 35, \\ 0 & if \quad 35 < x \le 50, \\ {x - 50}/{20} & if \quad 50 < x \le 70, \end{cases} \tag{10}$$

$$T_{old}(x)\begin{cases} 0 & if \quad 25 < x \leq 70, \\ x-35/20 & if \quad 45 < x \leq 50, \\ 1 & if \quad 35 < x \leq 45, \\ x-35/50 & if \quad 50 < x \leq 70, \end{cases} \tag{11}$$

$$I_{old}(x)\begin{cases} 1 & if \quad 50 < x \leq 70 \\ 50-x/25 & if \quad 25 < x \leq 35 \\ 0 & if \quad 35 < x \leq 50 \end{cases} \tag{12}$$

$$F_{old}(x)\begin{cases} 1 & if \quad 25 < x \leq 45, \\ 60-x/20 & if \quad 45 < x \leq 50, \\ 0 & if \quad x < 25, 55 < x \geq 70, \\ x-35/25 & if \quad 50 < x \leq 55, \end{cases} \tag{13}$$

From Eq. (5)-(13), we can calculate the following information about the patients:

$$Alex(30t) = \langle old(0, 0.8, 1); middle(0.25, 0.25, 0.5), young(0.75, 0.16, 0.25)\rangle, \tag{14}$$

$$Linda(40t) = \langle old(1, 0, 1); middle(1, 0.6, 0); young(0.25, 1, 1)\rangle, \tag{15}$$

$$Bill(50t) = \langle old(0.75, 0, 0.5); middle(1, 0.4, 0); young(0, 1, 1)\rangle, \tag{16}$$

$$John(55t) = \langle old(0.4, 1, 0.8); middle(0.75, 0, 0.25); young(0, 1, 1)\rangle. \tag{17}$$

Similarly, the truth membership functions, indeterminacy membership functions and falsity membership functions of the symptoms to the linguistic label $jth$ of characteristic $\Upsilon$ are given below:

$$T_{cold}(x)\begin{cases} {20-x}/{25} & if \quad x \le 5, \\ {x-5}/{15} & if \quad 5 < x \le 15, \\ {30-x}/{15} & if \quad 15 < x \le 30, \\ 0 & if \quad x > 30, \end{cases} \tag{18}$$

$$I_{cold}(x)\begin{cases} {10-x}/{15} & if \quad x \le 5, \\ {x-5}/{15} & if \quad 5 < x \le 15, \\ {x-15}/{30} & if \quad 15 < x \le 30, \\ 1 & if \quad x > 30, \end{cases} \tag{19}$$

$$F_{cold}(x)\begin{cases} {5-x}/{10} & if \quad 5 \le x < 10, \\ {x-5}/{20} & if \quad 10 \le x < 20, \\ {30-x}/{20} & if \quad 20 \le x \le 30, \\ 1 & if \quad x > 30, \end{cases} \tag{20}$$

$$T_{medium}(x)\begin{cases} {15-x}/{15} & if \quad 5 \le x < 10, \\ {20-x}/{20} & if \quad 10 \le x < 20, \\ {30-x}/{10} & if \quad 20 \le x \le 30, \\ 0 & if \quad x > 30, \end{cases} \tag{21}$$

$$I_{medium}(x)\begin{cases} {9-x}/{15} & if \quad 5 \le x < 9, \\ {x-5}/{18} & if \quad 9 \le x < 18, \\ {30-x}/{30} & if \quad 18 \le x \le 30, \\ 1 & if \quad x > 30, \end{cases} \tag{22}$$

$$F_{medium}(x)\begin{cases} 7-x\big/5 & if \quad 5 \le x < 7, \\ x-7\big/14 & if \quad 7 \le x < 14, \\ 30-x\big/14 & if \quad 14 \le x \le 30, \\ 1 & if \quad x > 30, \end{cases} \tag{23}$$

$$T_{hot}(x)\begin{cases} 10-x\big/10 & if \quad 5 \le x < 10, \\ x-7\big/17 & if \quad 10 \le x < 17, \\ 30-x\big/17 & if \quad 17 \le x \le 30, \\ 0 & if \quad x > 30, \end{cases} \tag{24}$$

$$I_{hot}(x)\begin{cases} 8-x\big/5 & if \quad 5 \le x < 8, \\ x-8\big/16 & if \quad 8 \le x < 16, \\ 30-x\big/16 & if \quad 16 \le x \le 30, \\ 1 & if \quad x > 30, \end{cases} \tag{25}$$

$$F_{hot}(x)\begin{cases} 5-x\big/11 & if \quad 5 \le x < 11, \\ x-11\big/18 & if \quad 11 \le x < 18, \\ 30-x\big/18 & if \quad 18 \le x \le 30, \\ 1 & if \quad x > 30, \end{cases} \tag{26}$$

From Eqs. (18)- (26), we can compute the information of symptoms as follows:

$$\left(4^{\circ}C\right) = \left\langle cold\left(0.64, 0.4, 0.1\right); medium\left(0.73, 0.33, 0.6\right); hot\left(0.6, 0.8, 0.09\right)\right\rangle \tag{27}$$

$$\left(15^{\circ}C\right) = \left\langle cold\left(0.66, 0.66, 0.5\right); medium\left(0.25, 0.55, 0.57\right); hot\left(0.47, 0.43, 0.22\right)\right\rangle \tag{28}$$

$$\left(22^{\circ}C\right) = \left\langle cold\left(0.53, 0.23, 0.4\right); medium\left(0.8, 0.26, 0.57\right); hot\left(0.47, 0.5, 0.44\right)\right\rangle \tag{29}$$

$$\left(28^{\circ}C\right) = \left\langle cold\left(0.13, 0.43, 0.1\right); medium\left(0.2, 0.06, 0.14\right); hot\left(0.11, 0.12, 0.11\right)\right\rangle \tag{30}$$

Eqs. (14)-(17) and Eqs. (27)-(30) can be written in the following tabular form.

**Table 5. A neutrosophic recommender system for medical diagnosis**

| Age | Temperature | Fever Level 100 M/l |
|---|---|---|
| $Alex(30t)=\left\langle \begin{matrix} old(0,0.8,1); \\ middle(0.25,0.25,0.5); \\ young(0.75,0.16,0.25) \end{matrix} \right\rangle$ | $(4^{\circ}C)=\left\langle \begin{matrix} cold(0.64,0.4,0.1); \\ medium(0.73,0.33,0.6); \\ hot(0.6,0.8,0.09) \end{matrix} \right\rangle$ | $L_1=\langle 0.5,0.3,0.5\rangle$, $L_2=\langle 0.4,0.7,0.1\rangle$, $L_3=\langle 0.7,0,0\rangle$ |
| $Linda(40t)=\left\langle \begin{matrix} old(1,0,1); \\ middle(1,0.6,0); \\ young(0.25,1,1) \end{matrix} \right\rangle$ | $(15^{\circ}C)=\left\langle \begin{matrix} cold(0.66,0.66,0.5); \\ medium(0.25,0.55,0.57); \\ hot(0.47,0.43,0.22) \end{matrix} \right\rangle$ | $L_1=\langle 0.9,0.1,0.3\rangle$, $L_2=\langle 0,0,0.8\rangle$, $L_3=\langle 0.7,0,0.5\rangle$ |
| $Bill(50t)=\left\langle \begin{matrix} old(0.75,0,0.5); \\ middle(1,0.4,0); \\ young(0,1,1) \end{matrix} \right\rangle$ | $(22^{\circ}C)=\left\langle \begin{matrix} cold(0.53,0.23,0.4); \\ medium(0.8,0.26,0.57); \\ hot(0.47,0.5,0.44) \end{matrix} \right\rangle$ | $L_1=\langle 0.15,0.03,0.01\rangle$, $L_2=\langle 0.24,0.75,0.16\rangle$, $L_3=\langle 0.8,0.3,0.1\rangle$ |
| $John(55t)=\left\langle \begin{matrix} old(0.4,1,0.8); \\ middle(0.75,0,0.25), \\ young(0,1,1) \end{matrix} \right\rangle$ | $(28^{\circ}C)=\left\langle \begin{matrix} cold(0.13,0.43,0.1); \\ medium(0.2,0.06,0.14); \\ hot(0.11,0.12,0.11) \end{matrix} \right\rangle$ | $L_1=\langle 0.55,0,0\rangle$, $L_2=\langle 0,0.7,0.9\rangle$, $L_3=\langle 0.4,0.4,0.4\rangle$ |

### 3.2 Algebraic Operations on Neutrosophic Recommender System NRS

In this section, we introduced some algebraic operations of neutrosophic recommender system NRS and studied some of their basic properties. We now proceed to define set theoretic operations of NRS.

Suppose that we have three subsets of $NRS=\{\mathrm{X},\mathrm{Y},\{\mathrm{D}_k\}\,|\,\mathrm{k}=1,2,...,n\}$ which are given below:

$NRS_1=\{\mathrm{X}_1;\mathrm{Y}_1;\{\mathrm{D}_i^1\}\,|\,\mathrm{i}=1,2,...,n\}$; $NRS_2=\{\mathrm{X}_2;\mathrm{Y}_2;\{\mathrm{D}_i^2\}\,|\,\mathrm{i}=1,2,...,n\}$; and

$NRS_3=\{\mathrm{X}_3;\mathrm{Y}_3;\{\mathrm{D}_i^3\}\,|\,\mathrm{i}=1,2,...,n\}$; where, $X_i\subset X;Y_i\subset Y$ and

$D_i^j=\left\{\mathrm{R}_i^j;\ \mathrm{T}_i^j;\ \mathrm{F}_i^j;\ \mathrm{I}_i^j\right\}=\left\{\left(\mathrm{R}_{iq}^j;\ \mathrm{T}_{iq}^j;\ \mathrm{F}_{iq}^j;\ \mathrm{I}_{iq}^j\right)|\ \mathrm{q}=1,2,...,\mathrm{r};\mathrm{j}=1,2,3;\right\}$.

Some algebraic operations of neutrosophic recommender system (NRS) can be defined below:

**(a) Union:** $NRS_1 \cup NRS_2 = NRS_{12}$, where

$$NRS_{12} = \{X_{12}; Y_{12}; \{D_l^{12}\} \mid l=1,2,...,k\}$$
$$X_{12} = X_1 \cup X_2$$
$$Y_{12} = Y_1 \cup Y_2$$

$$\begin{aligned} &\{D_l^{12}\} = (R_l^{12}; T_l^{12}; F_l^{12}; I_l^{12}) = \{(R_{lq}^{12}; T_{lq}^{12}; F_{lq}^{12}; I_{lq}^{12}) \mid q=1,2,...,r,\ l \in N;\ k \in N\} \\ &T_{lq}^{12} = \max\{T_{lq}^1; T_{lq}^2\};\ F_{lq}^{12} = \min\{F_{lq}^1; F_{lq}^2\};\ I_{lq}^{12} = \min\{I_{lq}^1; I_{lq}^2\} \end{aligned} \tag{31}$$

**(b) Intersection:** $NRS_1 \cap NRS_2 = NRS_{12}$, where

$$\begin{aligned} &NRS_{12} = \{X_{12}; Y_{12}; \{D_l^{12}\} \mid l=1,2,...,k\} \\ &X_{12} = X_1 \cap X_2 \\ &Y_{12} = Y_1 \cap Y_2 \\ &\{P_l^{12}\} = \{R_l^{12}; T_l^{12}; F_l^{12}; I_l^{12}\} = \{(R_{lq}^{12}; T_{lq}^{12}; F_{lq}^{12}; I_{lq}^{12}) \mid q=1,2,...,r,\ l \in N;\ k \in N\} \\ &T_{lq}^{12} = \min\{T_{lq}^1; T_{lq}^2\};\ F_{lq}^{12} = \max\{F_{lq}^1; F_{lq}^2\};\ I_{lq}^{12} = \max\{I_{lq}^1; I_{lq}^2\} \end{aligned} \tag{32}$$

(c) **Complement:** $NRS_1^c = \{X_1^C; Y_1^C; \{D_i^{1^C}\} \mid i=1,2,...,n\};$ where

$$\begin{aligned} &X_1^C = X \setminus X_1, \\ &Y_1^C = Y \setminus Y_1, \\ &\{D_i^{1^C}\} = \{R_i^{1^C}; T_i^{1^C}; F_i^{1^C}; I_i^{1^C}\} = \{\left(R_{iq}^{1^C}; T_{iq}^{1^C}; F_{iq}^{1^C}; I_{iq}^{1^C}\right) \mid q=1,2,...,r;\ i=1,2,...,n\} \\ &T_{iq}^{1^C} = F_{iq}^1;\quad F_{iq}^{1^C} = I_{iq}^1;\quad I_{iq}^{1^C} = T_{iq}^1 \end{aligned} \tag{33}$$

Here we derived some properties of these algebraic operations.

**(a) Commutative:**

$$\begin{aligned} &NRS_1 \cup NRS_2 = NRS_2 \cup NRS_1, \\ &NRS_1 \cap NRS_2 = NRS_2 \cap NRS_1. \end{aligned} \tag{34}$$

**(b) Associative:**

$$(NRS_1 \cup NRS_2) \cup \mathrm{NRS}_3 = NRS_1 \cup (NRS_2 \cup NRS_3),$$
$$(NRS_1 \cap NRS_2) \cap \mathrm{NRS}_3 = NRS_1 \cap (NRS_2 \cap NRS_3). \quad (35)$$

**(c) Distributive property:**

$$(NRS_1 \cup NRS_2) \cap \mathrm{NRS}_3 = (NRS_1 \cap NRS_2) \cup (NRS_2 \cap NRS_3)$$

We proved the first commutative property in Equation (34). Other properties can be proved analogously.

The fact that, we set

$$NRS_1 \cup NRS_2 = NRS_{12},$$

$$\begin{aligned}
&NRS_{12} = \{\mathrm{X}_{12};\mathrm{Y}_{12};\{\mathrm{D}_l^{12}\}|\ \mathrm{l}=1,2,...,\mathrm{k}\},\text{where}\\
&\mathrm{X}_{12} = X_1 \cup X_2\\
&Y_{12} = Y_1 \cup Y_2\\
&\{P_l^{12}\} = \{\mathrm{R}_l^{12};\mathrm{T}_l^{12};\ \mathrm{F}_l^{12};\ \mathrm{I}_l^{12}\ \}=\{(\mathrm{R}_{lq}^{12};\mathrm{T}_{lq}^{12};\ \mathrm{F}_{lq}^{12};\ \mathrm{I}_{lq}^{12})|\ \mathrm{q}=1,2,...,\mathrm{r},\ \mathrm{l} \in \mathrm{N};\ \mathrm{k} \in \mathrm{N}\}\\
&\mathrm{T}_{lq}^{12} = \max\{\mathrm{T}_{lq}^1;\ \mathrm{T}_{lq}^2\};\ \mathrm{F}_{lq}^{12} = \min\{\mathrm{F}_{lq}^1;\mathrm{F}_{lq}^2\};\ \mathrm{I}_{lq}^{12} = \min\{\mathrm{I}_{lq}^1;\mathrm{I}_{lq}^2\}.
\end{aligned} \quad (36)$$

Similarly, we have

$$NRS_2 \cup NRS_1 = NRS_{21},$$

$$\begin{aligned}
&NRS_{21} = \{\mathrm{X}_{21};\mathrm{Y}_{21};\{\mathrm{D}_l^{21}\}|\ \mathrm{l}=1,2,...,\mathrm{k}\}\\
&\mathrm{X}_{21} = X_2 \cup X_1\\
&Y_{21} = Y_2 \cup Y_1\\
&\{P_l^{21}\} = \{\mathrm{R}_l^{21};\mathrm{T}_l^{21};\ \mathrm{F}_l^{21};\ \mathrm{I}_l^{21}\ \}=\{(\mathrm{R}_{lq}^{21};\mathrm{T}_{lq}^{21};\ \mathrm{F}_{lq}^{21};\ \mathrm{I}_{lq}^{21})|\ \mathrm{q}=1,2,...,\mathrm{r};\mathrm{l}=1,2,...,\mathrm{k}\}\\
&\mathrm{T}_{lq}^{21} = \max\{\mathrm{T}_{lq}^2;\ \mathrm{T}_{lq}^1\};\ \mathrm{F}_{lq}^{21} = \min\{\mathrm{F}_{lq}^2;\mathrm{F}_{lq}^1\};\ \mathrm{I}_{lq}^{21} = \min\{\mathrm{I}_{lq}^2;\mathrm{I}_{lq}^1\}.
\end{aligned} \quad (37)$$

Thus, $NRS_{12} = NRS_{21}$ such that $NRS_2 \cup NRS_1 = NRS_1 \cup NRS_2$ are the subsets of

$X = \{\mathrm{x}_1;\mathrm{x}_2;\mathrm{x}_3\};\mathrm{Y} = \{\mathrm{y}_1;\mathrm{y}_2;\mathrm{y}_3\}$ ; $X_1; X_2 \subset X; Y_1; Y_2 \subset Y$ . □

**Definition 9:** Let $NRS_1$ and $NRS_2$ be two neutrosophic recommender systems. The probabilistic sum of $NRS_1$ and $NRS_2$ denoted as $NRS_1 \hat{+} NRS_2$ and is defined as:

$$NRS_1 \hat{+} NRS_2 = \left\{ X_{12}; Y_{12}; \{D_l^{12}\} \right\}, X_{12} = X_1 \cup X_2; Y_{12} = Y_1 \cup Y_2, \tag{38}$$

$$\{D_l^{12}\} = \{R_l^{12}; T_l^{12}; F_l^{12}; I_l^{12}\} = \{(R_{lq}^{12}; T_{lq}^{12}; F_{lq}^{12}; I_{lq}^{12}) | q = \overline{1;r}, l \in N; k \in N\} \tag{39}$$

$$\begin{aligned} T_{lq}^{12} &= T_{lq}^1(x) + T_{lq}^2(x) - T_{lq}^1(x).T_{lq}^2(x); \\ I_{lq}^{12} &= I_{lq}^1(x) + I_{lq}^2(x) - I_{lq}^1(x).I_{lq}^2(x); \\ F_{lq}^{12} &= F_{lq}^1(x) + F_{lq}^2(x) - F_{lq}^1(x).F_{lq}^2(x) \end{aligned} \tag{40}$$

where $T_{lq}^{12}; I_{lq}^{12}; F_{lq}^{12}$ are their truth membership functions, indeterminacy membership functions and falsity membership functions respectively.

**Definition 10.** The bold sum of $NRS_1$ and $NRS_2$ is defined as following:

$$NRS_1 \oplus NRS_2 = \left\{ X_{12}; Y_{12}; \{D_l^{12}\} \right\}, X_{12} = X_1 \cup X_2; Y_{12} = Y_1 \cup Y_2 \tag{41}$$

$$\{D_l^{12}\} = \{R_l^{12}; T_l^{12}; F_l^{12}; I_l^{12}\} = \{(R_{lq}^{12}; T_{lq}^{12}; F_{lq}^{12}; I_{lq}^{12}) | q = 1,2,...,r, l \in N; k \in N\} \tag{42}$$

$$\begin{aligned} &T_{lq}^{12}(x) = \min\{1; T_{lq}^1(x) + T_{lq}^2(x)\}; I_{lq}^{12}(x) = \min\{1; I_{lq}^1(x) + I_{lq}^2(x)\}; \\ &F_{lq}^{12}(x) = \min\{1; F_{lq}^1(x) + F_{lq}^2(x)\} \end{aligned} \tag{43}$$

**Definition 11:** The bold intersection of $NRS_1$ and $NRS_2$ can be defined as following:

$$NRS_1 \cap NRS_2 = \left\{ X_{12}; Y_{12}; \{D_l^{12}\} \right\}, X_{12} = X_1 \cap X_2; Y_{12} = Y_1 \cap Y_2 \tag{44}$$

$$\{D_l^{12}\} = \{R_l^{12}; T_l^{12}; F_l^{12}; I_l^{12}\} = \{(R_{lq}^{12}; T_{lq}^{12}; F_{lq}^{12}; I_{lq}^{12}) | q = 1,2,...,r, l \in N; k \in N\} \tag{45}$$

$$\begin{aligned} &T_{lq}^{12}(x) = \max\{0; T_{lq}^1(x) + T_{lq}^2(x) - 1\}; I_{lq}^{12}(x) = \max\{0; I_{lq}^1(x) + I_{lq}^2(x) - 1\}; \\ &F_{lq}^{12}(x) = \max\{0; F_{lq}^1(x) + F_{lq}^2(x) - 1\} \end{aligned} \tag{46}$$

**Definition 12:** The bounded difference of $NRS_1$ and $NRS_2$ is defined as:

$$NRS_1 |-| NRS_2 = \left\{X_{12};Y_{12};\{D_l^{12}\}\right\}, X_{12} = X_1 \setminus X_2; Y_{12} = Y_1 \setminus Y_2 \tag{47}$$

$$\{D_l^{12}\} = \{R_l^{12};T_l^{12};F_l^{12};I_l^{12}\} = \{(R_{lq}^{12};T_{lq}^{12};F_{lq}^{12};I_{lq}^{12}) | q=1,2,...,r, l \in N; k \in N\} \tag{48}$$

$$\begin{aligned} &T_{lq}^{12}(x) = \max\{0;T_{lq}^1(x) - T_{lq}^2(x)\}; I_{lq}^{12}(x) = \max\{0;I_{lq}^1(x) - I_{lq}^2(x)\}; \\ &F_{lq}^{12}(x) = \max\{0;F_{lq}^1(x) - F_{lq}^2(x)\} \end{aligned} \tag{49}$$

**Definition 13:** The symmetrical difference of two neutrosophic recommender systems is defined as:

$$NRS_1 \nabla NRS_2 = \left\{X_{12};Y_{12};\{D_l^{12}\}\right\}, X_{12} = X_1 \cup X_2; Y_{12} = Y_1 \cup Y_2 \tag{50}$$

$$\{D_l^{12}\} = \{R_l^{12};T_l^{12};F_l^{12};I_l^{12}\} = \{(R_{lq}^{12};T_{lq}^{12};F_{lq}^{12};I_{lq}^{12}) | q=1,2,...,r, l \in N; k \in N\} \tag{51}$$

$$\begin{aligned} &T_{lq}^{12}(x) = | T_{lq}^1(x) - T_{lq}^2(x) |; I_{lq}^{12}(x) = | I_{lq}^1(x) - I_{lq}^2(x) |; \\ &F_{lq}^{12}(x) = | F_{lq}^1(x) - F_{lq}^2(x) | \end{aligned} \tag{52}$$

**Definition 14:** The convex linear sum of min and max of $NRS_1$ and $NRS_2$ is defined as following:

$$NRS_1 \|_{\lambda} NRS_2 = \left\{X_{12};Y_{12};\{D_l^{12}\}\right\}, X_{12} = X_1 \cup X_2; Y_{12} = Y_1 \cup Y_2 \tag{53}$$

$$\{D_l^{12}\} = \{R_l^{12};T_l^{12};F_l^{12};I_l^{12}\} = \{(R_{lq}^{12};T_{lq}^{12};F_{lq}^{12};I_{lq}^{12}) | q=1,2,...,r, l \in N; k \in N\} \tag{54}$$

where

$$\begin{aligned} &T_{lq}^{12}(x) = \lambda \min\{T_{lq}^1(x);T_{lq}^2(x)\} + (1-\lambda)\max\{T_{lq}^1(x);T_{lq}^2(x)\}, \\ &I_{lq}^{12}(x) = \lambda \min\{I_{lq}^1(x),I_{lq}^2(x)\} + (1-\lambda)\max\{I_{lq}^1(x);I_{lq}^2(x)\}, \\ &F_{lq}^{12}(x) = \lambda \min\{F_{lq}^1(x),F_{lq}^2(x)\} + (1-\lambda)\max\{F_{lq}^1(x);F_{lq}^2(x)\}, \end{aligned} \tag{55}$$

and $\lambda \in [0;1]$.

**Definition 15:** The Cartesian product of $NRS_1$ and $NRS_2$ is defined as:

$$NRS_1 \times_1 NRS_2 = \left\{X_{12}; Y_{12}; \{D_l^{12}\}\right\}, X_{12} = X_1 \times X_2; Y_{12} = Y_1 \times Y_2 \tag{56}$$

$$D_l^{12} = \left\{\left\langle (x;y), \mathrm{T}_{lq}^1(x) T_{lq}^2(y), \mathrm{F}_{lq}^1(x).F_{lq}^2(x), \mathrm{I}_{lq}^1(x).I_{lq}^2(x)\right\rangle \mid \mathrm{X}_{12}; y \in Y_{12}\right\}; \tag{57}$$

$$NRS_1 \times_2 NRS_2 = \left\{X_{12}; Y_{12}; \{D_l^{12}\}\right\}, X_{12} = X_1 \times X_2; Y_{12} = Y_1 \times Y_2, \tag{58}$$

$$D_l^{12} = \left\{\left\langle \begin{array}{l} (x;y), \min\{\mathrm{T}_{lq}^1(x), T_{lq}^2(y)\}, \\ \min\{\mathrm{F}_{lq}^1(x), F_{lq}^2(x)\}, \\ \max\{\mathrm{I}_{lq}^1(x), I_{lq}^2(x)\} \end{array}\right\rangle \mid x \in X_{12}; y \in Y_{12}\right\}. \tag{59}$$

**Remark 3**: All of the above operators satisfy commutative and associative property.

**Example 4.** Suppose that we have,

$X_1 = \{(\mathrm{x}_1; 0,3; 0,5; 0,8); (\mathrm{x}_2; 0; 1,0; 0); (x_3; 0,5; 0,2; 0,6)\}$, $X_2 = \{(\mathrm{x}_1; 0,4; 0,3; 0,7); (\mathrm{x}_2; 0; 1,0; 0); (x_3; 0,8; 0,0; 0,5)\}$;

$Y_1 = \{(y_1; 0; 0,7; 0); (y_2; 0,4; 0,8; 0,6); (y_3; 0,2; 0,7; 0,4)\}$, $Y_2 = \{(y_1; 0,4; 0,5; 0,8); (y_2; 0,3; 0,4; 0,7); (y_3; 0; 0,8; 0)\}$,

$D = \{\mathrm{d}_1; \mathrm{d}_2\}$ where $D_X = \left\{\begin{array}{l} (d_{1x_1}; 0; 1; 0.5), (d_{1x_2}; 1; 0; 0.5), (\mathrm{d}_{1x_3}; 0.4; 1; 0.4), (d_{2x_1}; 0.3; 0.4; 0.1) \\ , (d_{2x_2}; 0.5; 0.2; 0.4), (d_{2x_3}; 0.6; 0.5; 0.2) \end{array}\right\}$

$$D_Y = \left\{\begin{array}{l} (d_{1y_1}; 0; 0.6; 0.2), (d_{1y_2}; 0; 0.8; 0.5), (\mathrm{d}_{1y_3}; 0.8; 0; 0.4), \\ (d_{2y_1}; 0.2; 0.7; 0.1), (d_{2y_2}; 0.8; 0.2; 0.1), (d_{2y_3}; 0.8; 0.5; 0.2) \end{array}\right\}$$

**Table 6. Algebraic operations**

| $NRS_1 \hat{+} NRS_2$ | $NRS_1 \oplus NRS_2$ | $NRS_1 \hat{\cap} NRS_2$ |
|---|---|---|
| $X_1 \cup X_2 =$<br>$\{(x_1;0.4;0.3;0.7);$<br>$(x_2;0;1;0);$<br>$(x_3;0.8;0.2;0.5)\}$<br>$Y_1 \cup Y_2 = \{(y_1;0.4;0.5;0)$<br>$;(y_2;0.4;0.4;0.6);$<br>$(y_3;0.2;0.7;0)\}$<br>$D_{\hat{+}}^{12} = \{(\mathrm{d}_{x_1}^{12};0.3;1;0.55);$<br>$(d_{x_2}^{12};1;0.2;0.7);$<br>$(d_{x_3}^{12};0.76;1;0.52);$<br>$(d_{y_1}^{12};0.2;0.88;0.28);$<br>$(d_{y_2}^{12};0.8;0.84;0.55)$<br>$(d_{y_3}^{12};0.96;0.5;0.52)\}$ | $X_1 \cup X_2 =$<br>$\{(x_1;0.4;0.3;0.7);$<br>$(x_2;0;1;0);$<br>$(x_3;0.8;0.2;0.5)\}$<br>$Y_1 \cup Y_2 = \{(y_1;0.4;0.5;0)$<br>$;(y_2;0.4;0.4;0.6);$<br>$(y_3;0.2;0.7;0)\}$<br>$D_{\hat{+}}^{12} = \{(\mathrm{d}_{x_1}^{12};0.3;1;0.6);$<br>$(d_{x_2}^{12};1;0.2;0.9);$<br>$(d_{x_3}^{12};1;1;0.6);$<br>$(d_{y_1}^{12};0.2;1;0.3);$<br>$(d_{y_2}^{12};0.2;1;0.6)$<br>$(d_{y_3}^{12};1;0.5;0.6)\}$ | $X_1 \cap X_2 =$<br>$\{(x_1;0.3;0.5;0.8);$<br>$(x_2;0;1;0);$<br>$(x_3;0.5;0.2;0.6)\}$<br>$Y_1 \cap Y_2 =$<br>$\{(y_1;0;0.7;0.8);$<br>$(y_2;0.3;0.8;0.7);$<br>$(y_3;0;0.8;0.4)\}$<br>$D_{\hat{\cap}}^{12} = \{(d_{x_1}^{12};0;0.4;0);$<br>$(d_{x_2}^{12};0.5;0;0)$<br>$;(d_{x_3}^{12};0;1;0.4);$<br>$(d_{y_1}^{12};0;0.3;0);$<br>$(d_{y_2}^{12};0;0;0);$<br>$(d_{y_3}^{12};0.6;0;0)\}$ |

| $NRS_1 \dot{\mid - \mid} NRS_2$ | $NRS_1 \dot{\nabla} NRS_2$ | $NRS_1 \dot{\|\|_{\lambda}} NRS_2$ |
|---|---|---|
| $X_1 \setminus X_2 =$<br>$\{(x_1;0;0.2;0.1);$<br>$(x_2;0;0;0);$<br>$(x_3;0;0.2;0.1)\}$<br>$Y_1 \setminus Y_2 = \{(y_1;0;0.2;0)$<br>$;(y_2;0.1;0.4;0);$<br>$(y_3;0.2;0;0.4)\}$ | $X_1 \cup X_2 =$<br>$\{(x_1;0.4;0.3;0.7);$<br>$(x_2;0;1;0);$<br>$(x_3;0.8;0.2;0.5)\}$<br>$Y_1 \cup Y_2 = \{(y_1;0.4;0.5;0)$<br>$;(y_2;0.4;0.4;0.6);$<br>$(y_3;0.2;0.7;0)\}$ | $X_1 \cup X_2 =$<br>$\{(x_1;0.4;0.3;0.7);$<br>$(x_2;0;1;0);$<br>$(x_3;0.8;0.2;0.5)\}$<br>$Y_1 \cup Y_2 = \{(y_1;0.4;0.5;0)$<br>$;(y_2;0.4;0.4;0.6);$<br>$(y_3;0.2;0.7;0)\}$ |
| $D_{\cup}^{12}=\{(d_{x_1}^{12};0;0.6;0.4);$<br>$(d_{x_2}^{12};0.5;0;0.1);$<br>$(d_{x_3}^{12};0;0.5;0.2);$<br>$(d_{y_1}^{12};0;0;0.1);$<br>$(d_{y_2}^{12};0;0.6;0.4)$<br>$(d_{y_3}^{12};0;0;0.2)\}$ | $D_{\nabla}^{12}=\{(d_{x_1}^{12};0.3;0.3;0.4);$<br>$(d_{x_2}^{12};0.5;0.2;0.1)$<br>$;(d_{x_3}^{12};0.2;0.5;0.2);$<br>$(d_{y_1}^{12};0.2;0.1;0.1);$<br>$(d_{y_2}^{12};0.8;0.6;0.4);$<br>$(d_{y_3}^{12};0;0.5;0.2)\}$ | $D^{12}=\{(d^{12}{}_{x_1};(1-\lambda)0.3;1-0.6\lambda;0.5-0.4\lambda);$<br>$(d_{x_2}^{12};0;\lambda;0.5-0.1\lambda);$<br>$(d_{x_3}^{12};0.6-0.2\lambda;1-0.5\lambda;0.4-0.2\lambda);$<br>$(d_{y_1}^{12};0.2(1-\lambda);0.7-0.1\lambda;0.2-0.1\lambda)$<br>$;(d_{y_2}^{12};0.8(1-\lambda);0.8-0.6\lambda;0.5-0.4\lambda)$<br>$;(d_{y_3}^{12};0.8;0.5.(1-\lambda);0.4-0.2\lambda)\}$<br>$\lambda\in[0;1]$ |

### 3.3 Algebraic Structures Based on Neutrosophic Recommender Systems and Their Operators

In this section, we proved that the collection of all neutrosophic recommender systems forms several types of algebraic structures such as lattices, de Morgan algebra, Kleen algebra, MV algebra, BCK-algebra, Stone algebra and Brouwerian Algebra. We also discussed some core properties of these algebraic structures.

Let F (NRS) denote the collection of all neutrosophic recommender systems. Further, we suppose that

$NRS_{\varnothing}$ is a neutrosophic recommender system which satisfies $X=\varnothing$ or $Y=\varnothing$.

**Proposition 1:** The structure $(F(NRS),\cup,\cap,NRS_{X\times Y},\mathrm{NRS}_{\varnothing})$ forms a complete lattice.

**Proof:** Let us consider $NRS_1,NRS_2,NRS_3\in F(NRS)$, then

1) From Definition (a) and (b) in Section 3.2, we have

   $NRS_1\cap NRS_2=NRS_{12}\in F(NRS)$ and $NRS_1\cup NRS_2=NRS_{12}\in F(NRS)$.

2) From Definition (a) and (b) in Section 3.2, we have

   $NRS_1\cap NRS_1=NRS_1$ and $NRS_1\cup NRS_1=NRS_1$

3) From properties (34) and (35), we see that

   $NRS_1\cap NRS_2=NRS_2\cap NRS_1$ and $NRS_1\cup NRS_2=NRS_2\cup NRS_1$

   $NRS_1\cap(NRS_2\cap NRS_3)=(NRS_1\cap NRS_2)\cap NRS_3,$
   $NRS_1\cup(NRS_2\cup NRS_3)=(NRS_1\cup NRS_2)\cup NRS_3.$

4) Also definition (31), (32), we have

$$NRS_1 \cap \left(NRS_2 \cup NRS_1\right) = NRS_1 \text{ and } NRS_1 \cup \left(NRS_2 \cap NRS_1\right) = NRS_1$$

Thus from 1) to 4), we saw that the structure $\left(F\left(NRS\right), \cup, \cap, NRS_{X\times Y}, \mathrm{NRS}_{\varnothing}\right)$ forms a lattice.

Consider a collection of neutrosophic recommender systems $\{\mathrm{NRS}_i : \mathrm{i} \in \mathrm{N}\}$ over F(NRS). We have,

$$\bigcap_{i=1}^{\infty} X_i \subseteq X, \bigcap_{i=1}^{\infty} Y_i \subseteq Y, \bigcup_{i=1}^{\infty} X_i \subseteq X, \bigcup_{i=1}^{\infty} Y_i \subseteq Y \text{ with } \mathrm{X}_i \subseteq X, Y_i \subseteq Y$$

and

$$D_l^{\infty} = \{\mathrm{R}_l^{\infty};\mathrm{T}_l^{\infty};\ \mathrm{F}_l^{\infty};\ \mathrm{I}_l^{\infty}\ \} = \{(\mathrm{R}_{lq}^{\infty};\mathrm{T}_{lq}^{\infty};\ \mathrm{F}_{lq}^{\infty};\ \mathrm{I}_{lq}^{\infty})|\ \mathrm{q}=1,2,...,\mathrm{r},\ \mathrm{l} \in \mathrm{N};\ \mathrm{k} \in \mathrm{N}\}$$
$$\mathrm{T}_{lq}^{\infty} = \max\{\mathrm{T}_{lq}^{1};\ \mathrm{T}_{lq}^{2};\ ...\};\ \mathrm{F}_{lq}^{\infty} = \min\{\mathrm{F}_{lq}^{1};\mathrm{F}_{lq}^{2};...\};\ \mathrm{I}_{lq}^{\infty} = \min\{\mathrm{I}_{lq}^{1};\mathrm{I}_{lq}^{2};...\}.$$

This implies

$$\bigcup_{i=1}^{\infty} NRS_i \subseteq F\left(NRS\right). \text{ Again, we have, } \bigcap_{i=1}^{\infty} NRS_i \subseteq F\left(NRS\right)$$

Thus we have proved that $F\left(NRS\right)$ is a complete lattice. □

**Proposition 2:** The structure $\left(F\left(NRS\right), \cup, \cap\right)$ is a bounded distributive lattice.

**Proof:** From condition 4) in above Proposition 1, we have

$$NRS_1 \cap \left(NRS_2 \cup NRS_3\right) = \left(NRS_1 \cap NRS_2\right) \cup \left(NRS_1 \cap NRS_2\right)$$

and

$$NRS_1 \cup \left(NRS_2 \cap NRS_3\right) = \left(NRS_1 \cup NRS_2\right) \cap \left(NRS_1 \cup NRS_2\right) \text{ for all } NRS_1; NRS_2; NRS_3 \in F\left(NRS\right).$$

This completes the proof. □

We can clearly see that $\left(F\left(NRS\right), \cup, \cap\right)$ is a dual lattice, so all the properties and structural configurations hold dually in an understood manner.

**Proposition 3: de Morgan Laws:** Let $NRS_1$ and $NRS_2 \in F\left(NRS\right)$. Then the following conditions hold.

1) $\left(NRS_1 \cup NRS_2\right)^c = NRS_1^c \cap NRS_2^c$,

2) $\left(NRS_1 \cap NRS_2\right)^c = NRS_1^c \cup NRS_2^c$.

**Proof:** Here we only prove 1).

1. Since, we have

$(NRS_1 \cup NRS_2)^c = NRS_{12}^c$, where

$$NRS_{12}^c = \{X_{12}^c; Y_{12}^c; \{D_l^{12c}\} | \ l=1,2,...,k\}$$
$$X_{12}^c = (X_1 \cup X_2)^c = X_1^c \cap X_2^c$$
$$Y_{12}^c = Y_1^c \cap Y_2^c$$
$$\{D_l^{12c}\} = (R_l^{12c}; T_l^{12c}; F_l^{12c}; I_l^{12c}) = \{(R_{lq}^{12c}; T_{lq}^{12c}; F_{lq}^{12c}; I_{lq}^{12c}) | \ q=1,2,...,r, \ l \in N; \ k \in N\}$$
$$T_{lq}^{12c} = F_{lq}^{12} = \min\{F_{lq}^1; F_{lq}^2\}; \ F_{lq}^{12c} = I_{lq}^{12} = \min\{I_{lq}^1; I_{lq}^2\}; \ I_{lq}^{12c} = T_{lq}^{12} = \max\{T_{lq}^1; T_{lq}^2\}$$

From definition of $NRS_1^c \cap NRS_2^c$ the proposition is proved. □

2) Can be proved on the same lines.

**Proposition 4:** $\left(F\left(NRS\right), \cup, \cap\right)$ forms a de Morgan algebra.

**Proof:** The proof is followed from Proposition 2 and 3. □

**Proposition 5:** $\left(F\left(NRS\right), \cup, \cap, ^c\right)$ forms a Boolean algebra.

**Proof:** From proposition 2 and proposition 3, we have $\left(F\left(NRS\right), \cup, \cap, ^c\right)$ is a bounded distributive lattice and $NRS_1 \in F\left(NRS\right)$ with its complement $NRS_1^c \in F\left(NRS\right)$ which completes the proof. □

**Proposition 6:** $\left(F\left(NRS\right), \cup, \cap, ^c, NRS^{\varnothing}\right)$ forms Kleen algebra.

**Proof:** From Proposition 4, $\left(F\left(NRS\right), \cup, \cap, ^c, NRS^{\varnothing}\right)$ forms de Morgan algebra. Moreover

$NRS_1 \cap NRS_1^c = NRS^{\varnothing} \subseteq NRS_2 \cup NRS_2^c$ with $NRS_1, NRS_2 \in F\left(NRS\right)$.

Thus by definition $\left(F\left(NRS\right), \cup, \cap, ^c, NRS^{\varnothing}\right)$ is a Kleen algebra. □

**Proposition 7:** $\left(F\left(NRS\right),\cap^{c},\mathrm{NRS}_{X\times Y}\right)$ is an MV – algebra.

**Proof:** To prove $\left(F\left(NRS\right),\cap,^{c},\mathrm{NRS}_{X\times Y}\right)$ is an MV – algebra. We have to prove the following 4 conditions:

**MV1:** $\left(F\left(NRS\right),\cap\right)$ is a commutative monoid. This proven is straightforward.

**MV2**: with every $NRS_1 \in F\left(NRS\right)$, we have $(NRS_1^c)^c = NRS_1$ which be implied from definition (33)

**MV3:** $NRS_1; NRS_2$ $(NRS_{X\times Y})^c \cap NRS_1 = NRS_{\varnothing} = (NRS_{X\times Y})^c$.

**MV4:** Since,

$$\left(NRS_1^c \cap NRS_2\right)^c \cap NRS_3 = ((NRS_1^c)^c \cup NRS_2^c) \cap \mathrm{NRS}_2$$
$$= (NRS_1 \cup NRS_2^c) \cap \mathrm{NRS}_2$$

$$= (\mathrm{NRS}_1 \cap \mathrm{NRS}_2) \cup \left(NRS_2^c \cap NRS_2\right)$$
$$= (\mathrm{NRS}_1 \cap \mathrm{NRS}_2) \cup NRS_{\varnothing}$$
$$= (\mathrm{NRS}_2 \cap \mathrm{NRS}_1) \cup \left(NRS_1^c \cap NRS_1\right)$$
$$= (\mathrm{NRS}_2 \cap \mathrm{NRS}_1^c) \cup NRS_1$$

for all $NRS_1, NRS_2 \in F\left(NRS\right)$. Thus $\left(F\left(NRS\right),\cap,^{c},\mathrm{NRS}_{X\times Y}\right)$ is an MV-algebra. □

**Proposition 8:** $\left(F\left(NRS\right),\cup^{c},\mathrm{NRS}_{\varnothing}\right)$ also forms an MV- algebra

**Proof:** **MV1, MV2** and **MV3** are straightforward. We prove **MV4**: Since,

$$\left(NRS_1^c \cup NRS_2\right)^c \cup NRS_2 = ((NRS_1^c)^c \cap NRS_2^c) \cup \mathrm{NRS}_2$$

$$= (NRS_1 \cup NRS_2) \cap (NRS_2^c \cup NRS_2)$$
$$= \left(NRS_1 \cup NRS_2\right) \cap NRS_{X\times Y}$$
$$= (NRS_1 \cup NRS_2) \cap (NRS_1^c \cup NRS_1)$$
$$= (NRS_1 \cup NRS_2) \cap (NRS_1 \cup NRS_1^c)$$
$$= (\mathrm{NRS}_2 \cap NRS_1^c) \cup \mathrm{NRS}_1$$
$$= (\mathrm{NRS}_2^c \cup NRS_1)^c \cup \mathrm{NRS}_1$$

for all $NRS_1, NRS_2 \in F(NRS)$. Thus $\left(F(NRS), \cup, ^c, \mathrm{NRS}_\varnothing\right)$ is MV- algebra. □

**Proposition 9:** $\left(F(NRS), |-|, \mathrm{NRS}_\varnothing\right)$ is a bounded BCK- algebra.

**Proof.** For any $NRS_1, NRS_2, NRS_3 \in F(NRS)$,

**BCI-1:**

$$\left(\left(NRS_1 |-| \mathrm{NRS}_2\right) |-| \left(NRS_1 |-| \mathrm{NRS}_3\right) |-| \left(NRS_2 |-| NRS_3\right)\right)$$
$$= NRS_\varnothing$$

**BCI-2:**

$$\left(NRS_1 |-| \left(NRS_1 |-| NRS_2\right)\right) |-| NRS_2$$
$$= NRS_\varnothing$$

**BCI-3:**

$$NRS_1 |-| NRS_2 = NRS_\varnothing.$$

**BCI-4:** Let

$$NRS_1 |-| NRS_2 = NRS_\varnothing; NRS_2 |-| NRS_1 = NRS_\varnothing \text{ and this implies that } NRS_1 = NRS_2$$

**BCI-5:**

$$NRS_\varnothing |-| NRS_1 = NRS_\varnothing$$

Thus $\left(F(NRS), |-|, \mathrm{NRS}_\varnothing\right)$ is BCK- algebra. Now $NRS_{X\times Y}$ is such that:

$$NRS_1 |-| NRS_{X\times Y} = NRS_\varnothing \text{ for all } NRS_1 \in F(NRS).$$

Therefore $\left(F(NRS), |-|, \mathrm{NRS}_\varnothing\right)$ is a bounded BCK- algebra. □

**Definition 16:** Let $\left(F(NRS), \cup, \cap, ^c\right)$ be a bounded lattice and $NRS_1 \in F(NRS)$. Then an element $NRS_1^c$ is called a pseudo-complement of $NRS_1$, if $NRS_1 \cap NRS_1^c = NRS_\varnothing$ and $NRS_2 \subseteq NRS_1^c$ whenever $NRS_1 \cap NRS_2 = NRS_\varnothing$.

If every element of a lattice F(NRS) is pseudo-complement, then F(NRS) is said to be pseudo-complemented. The equation $NRS_1^c \cup NRS = NRS_{X\times Y}$ is called Stone’s identity.

**Definition 17:** A Stone algebra is a pseudo-complemented, distributive lattice satisfying Stone's identity.

**Lemma 1:** Let $NRS_1, NRS_2 \in F(NRS)$ . Then the pseudo-complement of $NRS_1$ relative to $NRS_2$ exists in $F(NRS)$ .

**Lemma 2:** Let $NRS_1, NRS_2 \in F(NRS)$ . Then pseudo-complement of $NRS_1$ relative to $NRS_2$ exists in $F(NRS)$.

**Proposition 10:** $\left(F(NRS), \cup, \cap, ^c\right)$ forms a Brouwerian lattices.

**Proof:** The proof follows from Lemma 1 and 2. □

**3.4: Similarity Measures of Neutrosophic Recommender System Based Algebraic Operators**

In this section, we introduced several similarity measures based on the algebraic operations.

**Definition 18:** Let $F(NRS)$ be a family of neutrosophic recommender systems and $NRS_i, NRS_j$ are subsets of $F(NRS)$ where $i, j = 1, 2, ..., n$ . Then the similarity measure based on algebraic union and intersection operations of $NRS_i$ and $NRS_j$ denoted as $\$_{NRS_{ij}}$ and is defined below:

$$\$_{NRS_{ij}} = \bigcup_{i,j=1}^{n} \left\{ \left( \cap \left( S_{X_{ij}}, S_{\Upsilon_{ij}} \right) \right) \cup \left( \cap \left( S_{\Upsilon_{ij}}, S_{D^{ij}} \right) \right) \right\}, \tag{60}$$

where

$$S_{X_{ij}} = \frac{1}{r} \sum_{i,j=1}^{n} \left[ \frac{\left|T_{X_i}(x) - T_{X_j}(x)\right| \vee \left|I_{X_i}(x) - I_{X_j}(x)\right| \vee \left|F_{X_i}(x) - F_{X_j}(x)\right|}{2} \right], \tag{61}$$

$$S_{\Upsilon_{ij}} = \frac{1}{r} \sum_{i,j=1}^{n} \left[ \frac{\left|T_{\Upsilon_i}(y) - T_{\Upsilon_j}(y)\right| \vee \left|I_{\Upsilon_i}(y) - I_{\Upsilon_j}(y)\right| \vee \left|F_{\Upsilon_i}(y) - F_{\Upsilon_j}(y)\right|}{2} \right], \tag{62}$$

$$S_{D^{ij}} = \frac{1}{r} \sum_{i,j=1}^{n} \left[ \frac{\left|T_{D^i}(d) - T_{D^j}(d)\right| \vee \left|I_{D^i}(d) - I_{D^j}(d)\right| \vee \left|F_{D^i}(d) - F_{D^j}(d)\right|}{2} \right] \tag{63}$$

**Eq. (60),** is the similarity measure of the neutrosophic recommender systems whereas $\bigcup, \bigcap$ are union and intersection algebraic operations respectively. **Eqs. (61), (62)** and **(63)** are the similarity measures of the users or features of the patients $\mathrm{X}_i, \mathrm{X}_j$, items or characteristics of the symptoms $\Upsilon_i, \Upsilon_j$ and the ratings or diseases $D^i, D^j$ respectively. The variable $'r'$ in the **Eqs. (61), (62)** and **(63)** is the number of linguistic labels.

**Proposition 11:** The similarity measure $\$_{NRS_{ij}}$ defined in **Eq. (60)** satisfies the following conditions.

1. $0 \le \$_{NRS_{ij}} \le 1$;
2. $\$_{NRS_{ij}} = 0$ if and only if $i = j$;
3. $\$_{NRS_{ij}} = \$_{NRS_{ji}}$;
4. If $NRS_k$ is another subset in $F(NRS)$ such that $NRS_i \le NRS_j \le NRS_k$, then $\$_{NRS_{ik}} \le \$_{NRS_{ij}}$ and $\$_{NRS_{ik}} \le \$_{NRS_{jk}}$ where $i, j, k = 1, 2, \ldots, n$.

**Proof:**

**1.** Since, the values of truth membership function, indeterminacy membership function and falsehood membership functions belongs to the unit interval $[0,1]$, so therefore, we have $0 \le S_{\mathrm{X}_{ij}} \le 1$, $0 \le S_{\Upsilon_{ij}} \le 1$ and $0 \le S_{D^{ij}} \le 1$. In this way, we also have, $0 \le \bigcap\left(S_{\mathrm{X}_{ij}}, S_{\Upsilon_{ij}}\right) \le 1, 0 \le \bigcap\left(S_{\Upsilon_{ij}}, S_{D^{ij}}\right) \le 1$ which implies that $0 \le \left\{\left(\cap\left(S\left(\mathrm{X}_{ij}\right), S\left(\Upsilon_{ij}\right)\right)\right) \cup \left(\cap\left(S\left(\Upsilon_{ij}\right), S\left(D^{ij}\right)\right)\right)\right\} \le 1$,

$$0 \le \bigcup_{i,j=1}^{n}\left\{\left(\cap\left(S\left(\mathrm{X}_{ij}\right), S\left(\Upsilon_{ij}\right)\right)\right) \cup \left(\cap\left(S\left(\Upsilon_{ij}\right), S\left(D^{ij}\right)\right)\right)\right\} \le 1,$$

Thus $0 \le \$_{NRS_{ij}} \le 1$. ∎

**2.** Suppose that $i = j$ which gives $\mathrm{X}_i = \mathrm{X}_j$ that implies

$$T_{l\mathrm{X}}^{i}(x)=T_{l\mathrm{X}}^{j}(x), I_{l\mathrm{X}}^{i}(x)=I_{l\mathrm{X}}^{j}(x), F_{l\mathrm{X}}^{i}(x)=F_{l\mathrm{X}}^{j}(x)$$

and

$$\left|T_{l\mathrm{X}}^{i}(x)-T_{l\mathrm{X}}^{j}(x)\right|=0, \left|I_{l\mathrm{X}}^{i}(x)-I_{l\mathrm{X}}^{j}(x)\right|=0, \left|F_{l\mathrm{X}}^{i}(x)-F_{l\mathrm{X}}^{j}(x)\right|=0.$$

Thus $S_{\mathrm{X}_{ij}}=0$.

Next we consider $\Upsilon_i=\Upsilon_j$ implies that

$$T_{l\Upsilon}^{i}(y)=T_{l\Upsilon}^{j}(y), I_{l\Upsilon}^{i}(y)=I_{l\Upsilon}^{j}(y), F_{l\Upsilon}^{i}(y)=F_{l\Upsilon}^{j}(y)$$

and

$$\left|T_{l\Upsilon}^{i}(y)-T_{l\Upsilon}^{j}(y)\right|=0, \left|I_{l\Upsilon}^{i}(y)-I_{l\Upsilon}^{j}(y)\right|=0, \left|F_{l\Upsilon}^{i}(y)-F_{l\Upsilon}^{j}(y)\right|=0.$$

Therefore, $S_{\Upsilon_{ij}}=0$

Finally by considering $D^i=D^j$ gives that

$$T_{lD}^{i}(d)=T_{lD}^{j}(d), I_{lD}^{i}(d)=I_{lD}^{j}(d), F_{lD}^{i}(d)=F_{lD}^{j}(d)$$

which implies that

$$\left|T_{lD}^{i}(d)-T_{lD}^{j}(d)\right|=0, \left|I_{lD}^{i}(d)-I_{lD}^{j}(d)\right|=0, \left|F_{lD}^{i}(d)-F_{lD}^{j}(d)\right|=0,$$

and hence we have $S_{D^{ij}}=0$. Consequently we have $\$_{NRS_{ij}}=0$. □

Similarly the converse can be proved on the same lines.

**3.** This is straightforward. □

**4.** If $X_i \le X_j \le X_k$ which implies that $T_{lX}^{i}(x) \le T_{lX}^{j}(x) \le T_{lX}^{k}(x)$, $I_{lX}^{i}(x) \ge I_{lX}^{j}(x) \ge I_{lX}^{k}(x)$ and $F_{lX}^{i}(x) \ge F_{lX}^{j}(x) \ge F_{lX}^{k}(x)$ for all $i, j, k, l = 1, 2, ..., n$.

Therefore, we have

$$\left|T_{lX}^{i}(x) - T_{lX}^{j}(x)\right| \le \left|T_{lX}^{i}(x) - T_{lX}^{k}(x)\right|,\ \left|T_{lX}^{j}(x) - T_{lX}^{k}(x)\right| \le \left|T_{lX}^{i}(x) - T_{lX}^{k}(x)\right|,$$

$$\left|I_{lX}^{i}(x) - I_{lX}^{j}(x)\right| \le \left|I_{lX}^{i}(x) - I_{lX}^{k}(x)\right|,\ \left|I_{lX}^{2}(x) - I_{lX}^{3}(x)\right| \le \left|I_{lX}^{1}(x) - I_{lX}^{3}(x)\right|,$$

$$\left|F_{lX}^{i}(x) - F_{lX}^{j}(x)\right| \le \left|F_{lX}^{i}(x) - F_{lX}^{k}(x)\right|,\ \left|F_{lX}^{j}(x) - F_{lX}^{k}(x)\right| \le \left|F_{lX}^{i}(x) - F_{lX}^{k}(x)\right|.$$

Thus $S_{X_{13}} \le S_{X_{12}}$ and $S_{X_{13}} \le S_{X_{23}}$.

Similarly we can easily show that $S_{\Upsilon_{ik}} \le S_{\Upsilon_{ij}}$, $S_{\Upsilon_{ik}} \le S_{\Upsilon_{jk}}$ and $S_{D^{ik}} \le S_{D^{ij}}$, $S_{D^{ik}} \le S_{D^{jk}}$.

Next,

$$\bigcap\left(S_{X_{ik}}, S_{\Upsilon_{ik}}\right) \le \bigcap\left(S_{X_{ij}}, S_{\Upsilon_{ij}}\right) \text{ and } \bigcap\left(S_{\Upsilon_{ik}}, S_{D^{ik}}\right) \le \bigcap\left(S_{ik}, S_{D^{ij}}\right).$$

Also,

$$\left\{\left(\bigcap\left(S_{X_{ik}}, S_{\Upsilon_{ik}}\right)\right) \bigcup \left(\bigcap\left(S_{\Upsilon_{ik}}, S_{D^{ik}}\right)\right)\right\} \le \left\{\left(\bigcap\left(S_{\Upsilon_{ij}}, S_{D^{ij}}\right)\right) \bigcup \left(\bigcap\left(S_{X_{ij}}, S_{\Upsilon_{ij}}\right)\right)\right\},$$

$$\bigcup_{i,k=1}^{n}\left\{\left(\bigcap\left(S_{X_{ik}}, S_{\Upsilon_{ik}}\right)\right) \bigcup \left(\bigcap\left(S_{\Upsilon_{ik}}, S_{D^{ik}}\right)\right)\right\} \le \bigcup_{i,j=1}^{n}\left\{\left(\bigcap\left(S_{\Upsilon_{ij}}, S_{D^{ij}}\right)\right) \bigcup \left(\bigcap\left(S_{X_{ij}}, S_{\Upsilon_{ij}}\right)\right)\right\}.$$

This shows that $\$_{NRS_{ik}} \le \$_{NRS_{ij}}$. □

Similarly, we can prove $\$_{NRS_{ik}} \le \$_{NRS_{jk}}$.

**Example 5:** Consider Table 5 from Example 3. Then the similarity measures of users $S_{X_{ij}}$, items $S_{Y_{ij}}$ and ratings $S_{D^{ij}}$ are calculated in the following table.

**Table 7. Similarity measures of patients, symptoms and diseases**

| SIM of Users $S_{X_{ij}}$ | SIM of Items $S_{Y_{ij}}$ | SIM of Ratings $S_{D^{ij}}$ |
|---|---|---|
| $S(Alex, Linda) = 0.4316,$ | $S(4^{\circ}C, 15^{\circ}C) = 0.20833,$ | $S(L_{Alex}, L_{Linda}) = 0.2666,$ |
| $S(Alex, Bill) = 0.39833,$ | $S(4^{\circ}C, 22^{\circ}C) = 0.11166,$ | $S(L_{Alex}, L_{Bill}) = 0.15833,$ |
| $S(Alex, John) = 0.29,$ | $S(4^{\circ}C, 28^{\circ}C) = 0.2866,$ | $S(L_{Alex}, L_{John}) = 0.28333,$ |
| $S(Linda, Bill) = 0.15833$ | $S(15^{\circ}C, 22^{\circ}C) = 0.2$ | $S(L_{Linda}, L_{Bill}) = 0.31666,$ |
| $S(Lind, John) = 0.30833$ | $S(15^{\circ}C, 28^{\circ}C) = 0.23$ | $S(L_{Linda}, L_{John}) = 0.25833,$ |
| $S(Bill, John) = 0.2333$ | $S(22^{\circ}C, 28^{\circ}C) = 0.23$ | $S(L_{Bill}, L_{John}) = 0.25666,$ |

The column $S_{X_{ij}}$ is the similarity measures of users (patients) which is calculated by Eq. (61), The column $S_{Y_{ij}}$ is the similarity measures of items (symptoms) which is calculated by Eq. (62), and finally, The column $S_{D^{ij}}$ is the similarity measures of ratings (diseases) calculated by Eq. (63) where $i, j = 1, 2, 3, 4$.

Now using Eq. (60), we can find the similarity matrix as in Table 8. Here in this table, $\bigcup$ is the algebraic union operation, $\$_{NRS_{ij}}$ is the similarity measures of $NRS_i$ and $NRS_j$ where $i, j = 1, 2, 3, 4, 5, 6$. The bold highlighted values in this table indicate the largest similarity measure of the two rows in Table 7. The main advantage of this similarity measure is to provide an identical largest value among all the rows in Table 7 because of the algebraic union operation used in the similarity measure. But we are unable to get largest distinct values of the similarity measure between two rows in Table 7 if we have to predict about several diseases of a patient which is the big failure of this similarity measure.

This similarity measure is not give accurate result as it lacks to provide us enough large value as compared to the similarity measures in Eqs. (65), (67), (69), and (71) respectively. This similarity measure can be used in those datasets where we have to find several patients with the same disease.

**Table 8. Similarity measures of $NRS_i$ and $NRS_j$ based on union and intersection**

| $\cup$ | $\$_{NRS_{i1}}$ | $\$_{NRS_{i2}}$ | $\$_{NRS_{i3}}$ | $\$_{NRS_{i4}}$ | $\$_{NRS_{i5}}$ | $\$_{NRS_{i6}}$ |
|---|---|---|---|---|---|---|
| $\$_{NRS_{1j}}$ | 0.20833 | 0.20833 | **0.2866** | 0.20833 | 0.23 | 0.23 |
| $\$_{NRS_{2j}}$ | 0.20833 | 0.11166 | **0.2866** | 0.2 | 0.23 | 0.23 |
| $\$_{NRS_{3j}}$ | **0.2866** | **0.2866** | **0.2866** | **0.2866** | **0.2866** | **0.2866** |
| $\$_{NRS_{4j}}$ | 0.20833 | 0.2 | **0.2866** | 0.2 | 0.23 | 0.23 |
| $\$_{NRS_{5j}}$ | 0.23 | 0.23 | **0.2866** | 0.23 | 0.23 | 0.23 |
| $\$_{NRS_{6j}}$ | 0.23 | 0.23 | **0.2866** | 0.23 | 0.23 | 0.23 |

**Definition 18:** Let $\$_{NRS_{ij}}$ be a similarity measure of $NRS_i$ and $NRS_j$ in Eq. (60) which is based on algebraic union and intersection operations. The weighted similarity measures can be defined as follows:

$$\$_{w\left(NRS_{ij}\right)} = \bigcup_{i,j=1}^{n} \left\{ \left( w_1 \times \left( \cap \left( S_{X_{ij}}, S_{\Upsilon_{ij}} \right) \right) \right) \cup \left( w_2 \times \left( \cap \left( S_{\Upsilon_{ij}}, S_{D^{ij}} \right) \right) \right) \right\} \tag{64}$$

with $w_1 + w_2 = 1$ where $i, j = 1, 2, ..., n$.

**Definition 19:** Let $F\left(NRS\right)$ be a family of neutrosophic recommender systems and $NRS_i, NRS_j$ are subsets of $F\left(NRS\right)$ where $i, j = 1, 2, ..., n$. We define the following similarity measures based on algebraic operations.

1. The similarity measure based on algebraic union, intersection and probabilistic sum is defined as:

$$\$_{NRS_{ij}} = \sum_{i,j=1}^{n}\left\{\left(\left(S_{\mathrm{X}_{ij}} + S_{\Upsilon_{ij}} - S_{\mathrm{X}_{ij}}.S_{\Upsilon_{ij}}\right)\right)\cap\left(\left(S_{\Upsilon_{ij}} + S_{D^{ij}} - S_{\Upsilon_{ij}}.S_{D^{ij}}\right)\right)\right\} \tag{65}$$

The weighted similarity measure of Eq. (65) is defined below:

$$\$_{w\left(NRS_{ij}\right)} = \sum_{i,j=1}^{n}\left\{\left(w_1\times\left(S_{\mathrm{X}_{ij}} + S_{\Upsilon_{ij}} - S_{\mathrm{X}_{ij}}.S_{\Upsilon_{ij}}\right)\right)\cap\left(w_2\times\left(S_{\Upsilon_{ij}} + S_{D^{ij}} - S_{\Upsilon_{ij}}.S_{D^{ij}}\right)\right)\right\} \tag{66}$$

with $w_1 + w_2 = 1$ where $i, j = 1, 2, ..., n$.

2. The similarity measure based on algebraic intersection and bold sum of $NRS_i$ and $NRS_j$ is defined as:

$$\$_{NRS_{ij}} = \prod_{i,j=1}^{n}\left\{\left(\min\left(1, S_{\mathrm{X}_{ij}} + S_{\Upsilon_{ij}}\right)\right)\cap\left(\min\left(1, S_{\Upsilon_{ij}} + S_{D^{ij}}\right)\right)\right\} \tag{67}$$

The weighted similarity measure of Eq. (67) is as following:

$$\$_{w\left(NRS_{ij}\right)} = \prod_{i,j=1}^{n}\left\{\left(w_1\times\left(\min\left(1, S_{\mathrm{X}_{ij}} + S_{\Upsilon_{ij}}\right)\right)\right)\cap\left(w_2\times\left(\min\left(1, S_{\Upsilon_{ij}} + S_{D^{ij}}\right)\right)\right)\right\} \tag{68}$$

with $w_1 + w_2 = 1$ where $i, j = 1, 2, ..., n$.

3. The similarity measure based on algebraic union and bounded difference of $NRS_i$ and $NRS_j$ is defined below:

$$\$_{NRS_{ij}} = \sum_{i,j=1}^{n}\left\{\left(\max\left(0, S_{\mathrm{X}_{ij}} - S_{\Upsilon_{ij}}\right)\right)\cup\left(\max\left(0, S_{\Upsilon_{ij}} - S_{D^{ij}}\right)\right)\right\} \tag{69}$$

The weighted similarity measure of Eq. (69) can be defined as following:

$$\$_{w\left(NRS_{ij}\right)} = \sum_{i,j=1}^{n}\left\{\left(w_1\times\left(\max\left(0, S_{\mathrm{X}_{ij}} - S_{\Upsilon_{ij}}\right)\right)\right)\cup\left(w_2\times\left(\max\left(0, S_{\Upsilon_{ij}} - S_{D^{ij}}\right)\right)\right)\right\} \tag{70}$$

with $w_1 + w_2 = 1$ where $i, j = 1, 2, ..., n$.

4. The similarity measure based on the operation of algebraic symmetrical difference of $NRS_i$ and $NRS_j$ is defined below:

$$\$_{NRS_{ij}} = \sum_{i,j=1}^{n} \left\{ \left( S_{X_{ij}} - S_{\Upsilon_{ij}} \right) + \left( S_{\Upsilon_{ij}} - S_{D^{ij}} \right) \right\} \tag{71}$$

The weighted similarity measure of Eq. (71) can be defined as following:

$$\$_{w(NRS_{ij})} = \sum_{i,j=1}^{n} \left\{ w_1 \times \left( S_{X_{ij}} - S_{\Upsilon_{ij}} \right) + w_2 \times \left( S_{\Upsilon_{ij}} - S_{D^{ij}} \right) \right\} \tag{72}$$

with $w_1 + w_2 = 1$ where $i, j = 1, 2, ..., n$.

Here in Definition $\$_{NRS_{ij}}$ denote the similarity measure between $NRS_i, NRS_j$ and $S_{X_{ij}}, S_{\Upsilon_{ij}}, S_{D^{ij}}$ are calculated from Eqs. (61), (62) and (63) respectively.

**Proposition 12:** The similarity measures $\$_{NRS_{ij}}$ defined in Eq. (65), (67), (69) and (71) satisfies the following conditions.

1. $0 \leq \$_{NRS_{ij}} \leq 1$;
2. $\$_{NRS_{ij}} = 0$ if and only if $i = j$;
3. $\$_{NRS_{ij}} = \$_{NRS_{ji}}$;
4. If $NRS_k$ is another subset of $F(NRS)$ such that $NRS_i \leq NRS_j \leq NRS_k$, then $\$_{NRS_{ik}} \leq \$_{NRS_{ij}}$ and $\$_{NRS_{ik}} \leq \$_{NRS_{jk}}$ where $i, j, k = 1, 2, ..., n$.

**Proof:** The proof of these is straightforward. □

**Example 5:** Consider $S_{X_{ij}}, S_{\Upsilon_{ij}}$ and $S_{D^{ij}}$ calculated in Table 7. From Eq. (65), (67), (69) and (71), we have calculated the following similarity matrix in Table 9, 10, 11, and 12 respectively.

**Table 9. Similarity measures of $NRS_i$ and $NRS_j$ based on intersection and probabilistic sum**

| $\sum$ | $\$_{NRS_{i1}}$ | $\$_{NRS_{i2}}$ | $\$_{NRS_{i3}}$ | $\$_{NRS_{i4}}$ | $\$_{NRS_{i5}}$ | $\$_{NRS_{i6}}$ |
|---|---|---|---|---|---|---|
| $\$_{NRS_{1j}}$ | 0.83878 | 0.67171 | **0.90812** | 0.74606 | 0.84832 | 0.82904 |
| $\$_{NRS_{2j}}$ | 0.67171 | 0.50464 | **0.74105** | 0.57899 | 0.68124 | 0.66197 |
| $\$_{NRS_{3j}}$ | **0.90812** | **0.74105** | **0.97746** | **0.8154** | **0.91765** | **0.89838** |
| $\$_{NRS_{4j}}$ | 0.74606 | 0.57899 | **0.8154** | 0.65334 | 0.75559 | 0.73632 |
| $\$_{NRS_{5j}}$ | 0.84832 | 0.68124 | **0.91765** | 0.75559 | 0.85784 | 0.83857 |
| $\$_{NRS_{6j}}$ | 0.82904 | 0.66197 | **0.89838** | 0.73632 | 0.83857 | 0.8193 |

**Table 10. Similarity measures of $NRS_i$ and $NRS_j$ based on intersection and bold sum**

| $\prod$ | $\$_{NRS_{i1}}$ | $\$_{NRS_{i2}}$ | $\$_{NRS_{i3}}$ | $\$_{NRS_{i4}}$ | $\$_{NRS_{i5}}$ | $\$_{NRS_{i6}}$ |
|---|---|---|---|---|---|---|
| $\$_{NRS_{1j}}$ | 0.22555 | 0.12822 | **0.27067** | 0.17018 | 0.23192 | 0.22003 |
| $\$_{NRS_{2j}}$ | 0.12822 | 0.07289 | **0.15387** | 0.09674 | 0.13184 | 0.12508 |
| $\$_{NRS_{3j}}$ | **0.27067** | **0.15387** | **0.32482** | **0.20422** | **0.27831** | **0.26404** |
| $\$_{NRS_{4j}}$ | 0.17018 | 0.09674 | **0.20422** | 0.12840 | 0.17498 | 0.16601 |
| $\$_{NRS_{5j}}$ | 0.23192 | 0.13184 | **0.27831** | 0.17498 | 0.23846 | 0.22624 |
| $\$_{NRS_{6j}}$ | 0.22003 | 0.12508 | **0.26404** | 0.16601 | 0.22624 | 0.21464 |

**Table 11. Similarity measures of $NRS_i$ and $NRS_j$ based on union and bounded difference**

| $\sum$ | $\$_{NRS_{i1}}$ | $\$_{NRS_{i2}}$ | $\$_{NRS_{i3}}$ | $\$_{NRS_{i4}}$ | $\$_{NRS_{i5}}$ | $\$_{NRS_{i6}}$ |
|---|---|---|---|---|---|---|
| $\$_{NRS_{1j}}$ | 0.44654 | **0.50994** | 0.22667 | 0.2237 | 0.3016 | 0.22657 |
| $\$_{NRS_{2j}}$ | **0.50994** | **0.57334** | **0.29007** | **0.28667** | **0.365** | **0.28997** |
| $\$_{NRS_{3j}}$ | 0.22667 | **0.29007** | 0.0068 | 0.0034 | 0.08173 | 0.0067 |
| $\$_{NRS_{4j}}$ | 0.2237 | **0.28667** | 0.0034 | 0 | 0.07833 | 0.0033 |
| $\$_{NRS_{5j}}$ | 0.3016 | **0.365** | 0.08173 | 0.07833 | 0.15666 | 0.08163 |
| $\$_{NRS_{6j}}$ | 0.22657 | **0.28997** | 0.0067 | 0.0033 | 0.08163 | 0.006 |

**Table 12. Similarity measures of $NRS_i$ and $NRS_j$ based on symmetrical difference**

| $\sum$ | $\$_{NRS_{i1}}$ | $\$_{NRS_{i2}}$ | $\$_{NRS_{i3}}$ | $\$_{NRS_{i4}}$ | $\$_{NRS_{i5}}$ | $\$_{NRS_{i6}}$ |
|---|---|---|---|---|---|---|
| $\$_{NRS_{1j}}$ | 0.33 | **0.405** | 0.17167 | 0.00667 | 0.215 | 0.14164 |
| $\$_{NRS_{2j}}$ | **0.405** | **0.48** | **0.24667** | **0.08167** | **0.29** | **0.21664** |
| $\$_{NRS_{3j}}$ | 0.17167 | **0.24667** | 0.01334 | -0.15166 | 0.05667 | -0.01669 |
| $\$_{NRS_{4j}}$ | 0.00667 | **0.08167** | -0.15166 | -0.31666 | -0.10833 | -0.18169 |
| $\$_{NRS_{5j}}$ | 0.215 | **0.29** | 0.05667 | -0.10833 | 0.1 | 0.02664 |
| $\$_{NRS_{6j}}$ | 0.14164 | **0.21664** | -0.01669 | -0.18169 | 0.02664 | -0.04672 |

The symbol $\sum$ refers to the aggregation operator in the Tables 9, 11, 12 while $\prod$ refers to the geometric operator in Table 10 and the bold values indicate the largest similarity measures in these tables. The similarity measure in Table 9 provides largest distinct values as compared to all other similarity measures in Tables (7), (9), (10) and (11). Further, this similarity measure provides us an accurate result with respect to other defined similarity measures in Eqs. (60), (67), (69), and (71) respectively.

As the similarity measure in Eq. (65) gives us accurate largest distinct results, so we will use this similarity measure to find the predicting formula for calculating the levels of diseases of the patients in this paper.

### 3.5 Prediction Formula and Non-linear Regression Model

**Definition 20:** Let $S_h$ be the symptoms of the patient $p_i$ whose diseases are $(d_1, d_2, ...., d_k)$ in a multi-criteria neutrosophic recommender system (MC-NRS) where $i = 1, 2, 3, ..., \mathrm{n}$ and $h = 1, 2, 3, ..., \mathrm{m}$. The linguistic labels of the patient $p_i$ can be predicted by the following formula:

$$T_{D_l}^{\ p_i}(d_h) = \frac{\sum_{j=1}^{n} \$_{NRS_{ij}} \times T_{D_l}^{\ p_j}(d_h)}{\sum_{j=1}^{n} \$_{NRS_{ij}}} \tag{73}$$

$$I_{D_l}^{\ p_i}(d_h) = T_{D_l}^{\ p_i}(d_h) + \frac{\sum_{j=1}^{n} \$_{NRS_{ij}} \times I_{D_l}^{\ p_j}(d_h)}{\sum_{j=1}^{n} \$_{NRS_{ij}}} \tag{74}$$

$$F_{D_l}^{\ p_i}(d_h) = I_{D_l}^{\ p_i}(d_h) + \frac{\sum_{j=1}^{n} \$_{NRS_{ij}} \times F_{D_l}^{\ p_j}(d_h)}{\sum_{j=1}^{n} \$_{NRS_{ij}}} \tag{75}$$

where $\$_{NRS_{ij}}$ is the similarity measure of Eq. (65) in Definition 16 and for all $h = 1, 2, 3, ..., \mathrm{m}$, $i = 1, 2, 3, ..., \mathrm{n}$ and for all $l = 1, 2, 3, ..., \mathrm{s}$. Eq. (73) refers to predictive truth membership function while Eq. (74) refers predictive indeterminate membership function and Eq. (75) is the predictive false membership function of the linguistic labels of the patient $p_i$.

**Theorem 1:** The predictive values of Eqs. (73), (74), and (75) in Definition 17 are neutrosophic values.

**Proof:** Since we have

$$T_{D_l}{}^{p_i}(d_h)+I_{D_l}{}^{p_i}(d_h)+F_{D_l}{}^{p_i}(d_h)=\frac{\sum_{j=1}^{n}\$_{NRS_{ij}}\times T_{D_l}{}^{p_j}(d_h)}{\sum_{j=1}^{n}\$_{NRS_{ij}}}+\left(T_{D_l}{}^{p_i}(d_h)+\frac{\sum_{j=1}^{n}\$_{NRS_{ij}}\times I_{D_l}{}^{p_j}(d_h)}{\sum_{j=1}^{n}\$_{NRS_{ij}}}\right)+\left(I_{D_l}{}^{p_i}(d_h)+\frac{\sum_{j=1}^{n}\$_{NRS_{ij}}\times F_{D_l}{}^{p_j}(d_h)}{\sum_{j=1}^{n}\$_{NRS_{ij}}}\right) \tag{76}$$

$$=\frac{\sum_{j=1}^{n}\$_{NRS_{ij}}\times T_{D_l}{}^{p_j}(d_h)}{\sum_{j=1}^{n}\$_{NRS_{ij}}}+\left(\frac{\sum_{j=1}^{n}\$_{NRS_{ij}}\times T_{D_l}{}^{p_j}(d_h)}{\sum_{j=1}^{n}\$_{NRS_{ij}}}+\frac{\sum_{j=1}^{n}\$_{NRS_{ij}}\times I_{D_l}{}^{p_j}(d_h)}{\sum_{j=1}^{n}\$_{NRS_{ij}}}\right)+\left(\left(\frac{\sum_{j=1}^{n}\$_{NRS_{ij}}\times T_{D_l}{}^{p_j}(d_h)}{\sum_{j=1}^{n}\$_{NRS_{ij}}}+\frac{\sum_{j=1}^{n}\$_{NRS_{ij}}\times I_{D_l}{}^{p_j}(d_h)}{\sum_{j=1}^{n}\$_{NRS_{ij}}}\right)+\frac{\sum_{j=1}^{n}\$_{NRS_{ij}}\times F_{D_l}{}^{p_j}(d_h)}{\sum_{j=1}^{n}\$_{NRS_{ij}}}\right) \tag{77}$$

$$=\frac{\sum_{j=1}^{n}\$_{NRS_{ij}}\times T_{D_l}{}^{p_j}(d_h)}{\sum_{j=1}^{n}\$_{NRS_{ij}}}+\frac{\sum_{j=1}^{n}\$_{NRS_{ij}}}{\sum_{j=1}^{n}\$_{NRS_{ij}}}\left[T_{D_l}{}^{p_j}(d_h)+I_{D_l}{}^{p_j}(d_h)\right]+\frac{\sum_{j=1}^{n}\$_{NRS_{ij}}}{\sum_{j=1}^{n}\$_{NRS_{ij}}}\left[\left(T_{D_l}{}^{p_j}(d_h)+I_{D_l}{}^{p_j}(d_h)\right)+F_{D_l}{}^{p_j}(d_h)\right] \tag{78}$$

$$=\frac{\sum_{j=1}^{n}\$_{NRS_{ij}}}{\sum_{j=1}^{n}\$_{NRS_{ij}}}\left[\left(3T_{D_l}{}^{p_j}(d_h)+2I_{D_l}{}^{p_j}(d_h)\right)+F_{D_l}{}^{p_j}(d_h)\right] \tag{79}$$

As we know from Definition 2 and Proposition 2 respectively that

$$0\le T_{D_l}{}^{p_j}(d_h)+I_{D_l}{}^{p_j}(d_h)+F_{D_l}{}^{p_j}(d_h)\le 3 \tag{80}$$

$$0\le \$_{NRS_{ij}}\le 1 \tag{81}$$

This become obvious that

$$T_{D_l}{}^{p_j}\left(d_h\right)+I_{D_l}{}^{p_j}\left(d_h\right)+F_{D_l}{}^{p_j}\left(d_h\right)\geq 0 \tag{82}$$

Therefore,

$$T_{D_l}{}^{p_i}\left(d_h\right)+I_{D_l}{}^{p_i}\left(d_h\right)+F_{D_l}{}^{p_i}\left(d_h\right)\leq \frac{\sum_{j=1}^{n}\$_{NRS_{ij}}}{\sum_{j=1}^{n}\$_{NRS_{ij}}}\left[\left(3T_{D_l}{}^{p_j}\left(d_h\right)+2I_{D_l}{}^{p_j}\left(d_h\right)\right)+F_{D_l}{}^{p_j}\left(d_h\right)\right]\leq 3 \tag{83}$$

This completes the proof. □

### 3.6 Non-linear Regression Model

We have proposed a non-linear regression model for the prediction task shown in Fig. 1. The problem with linear regression is that it can model curves, but it might not be able to model the specific curve that exists in the data. A non-linear model can possibly be infinite number of functions which can't be easy to setup. But it can best fit a mathematical model to some data. Moreover, we used a neutrosophication and de-neutrosophication which convert data from crisp inputs into neutrosophic input and again from neutrosophic set to crisp input respectively which is shown in Fig 1.

## 4. Evaluation

In this section, the experimental environment as well as the database for this experiment is presented. In the subsection 4.1, we describe the experimental set up while the subsection 4.2 is dedicated to the experimental results.

### 4.1 Experimental Set up

This part is dedicated to the experimental environments which are described below:

**Experimental tools**: The proposed algorithm has been implemented in addition to the methods of ICSM [59], DSM [61], and CARE [10], and CFMD [22]. We run our proposed method as well as all these algorithms and Variants in the Matlab 2015a programming language and executed them on a PC Intel(R) Core (TM) 2 Dual CPU T6400@2.00 GHz (2CPUs), 2048MB RAM and the operating system is windows7 Professional 32 bits. The following **Table 13** on the next page provides us the description of experimental datasets. Finally we analyzed the strength of all 8 algorithms by ANOVA test.

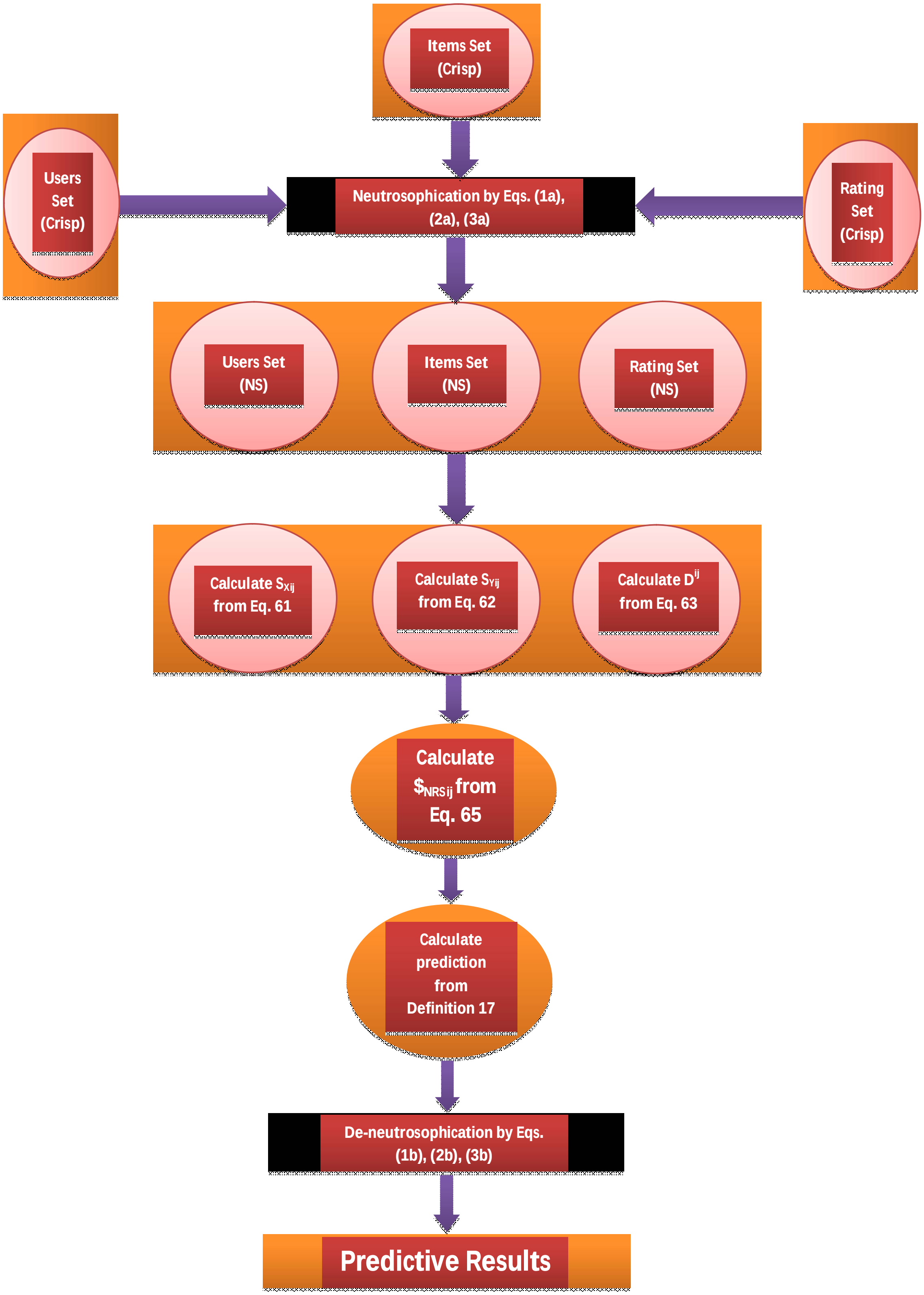

**Fig.1. Non-linear regression model**

**Table 13. The descriptions of experimental datasets**

| Dataset | No. elements | No. attributes | No. classes |
|---|---|---|---|
| RHC | 5736 | 5 | 3 |
| Diabetes | 404 | 4 | 2 |
| Breast | 3304 | 5 | 3 |
| DMD | 201 | 3 | 1 |
| Heart | 271 | 4 | 2 |

**Experimental datasets:** The benchmark dataset HEART has been taken from UCI Machine Learning Repository (University of California) while the remaining 4 benchmark datasets RHC (Right Heart Catheterization), Diabetes, Breast Cancer, DMD (Duchenne Muscular Dystrophy) have been taken from (Department of Biostatistics, Vanderbilt University). Table 13 gives an overview of all those datasets.

**Objectives:**

- To evaluate the qualities of algorithms through validity indices. In this regard, experiments on the computational time of algorithms are also considered;
- To validate the performance of algorithms by various cases of parameters.

### 4.2 The comparison of proposed algorithm quality

The following Table **14** presents the average MSE and computational time (Sec.) of our proposed method with ICSM [59], DSM [61], CARE [10], CFMD [22], Variant 67, Variant 69, and Variant 71 on the medical datasets of Heart, RHC, Diabetes, Breast and DMD. For more detailed information, we refer to the **Table 14**.

In this **Table 14**, Mean Square Error (MSE) and Computational time (Sec) of all 8 algorithms have been computed on the 5 data sets Heart, RHC, Diabetes, Breast and DMD respectively. It is clearly seen that the MSE of our proposed method is better than ICSM, DSM, CARE and CFMD on the Heart data set while it does not give a reasonable change in MSE with the variants 67, 69, and 71. Specifically in **Table 13**, the average MSE of ICSM, DSM, CARE, CFMD, Proposed method, Variant 67, Variant 69 and Variant 71 are **0.3407, 0.3407, 0.2502, 0.2525, 0.236052 ± 1.4393e-007, 0.235093 ± 1.8290e-007, 0.236079 ± 1.7144e-007, 0.234787 ± 2.2378e-007** respectively.

Table 14. Mean Square Error (MSE) and Computational time (Sec)

| Dataset | Mean Square Error (MSE) | | | | | | | |
|---|---|---|---|---|---|---|---|---|
| | ICSM | DSM | CARE | CFMD | Proposed method | Variant (67) | Variant (69) | Variant (71) |
| Heart | 0.3407 | 0.3407 | 0.2502 | 0.2525 | 0.236052 ± 1.4393e-007 | 0.235093 ± 1.8290e-007 | 0.236079 ± 1.7144e-007 | 0.234787 ± 2.2378e-007 |
| RHC | 0.1780 | 0.1780 | 0.3658 | 0.1896 | 0.25 | 0.250000 | 0.250000 | 0.250000 |
| Diabetes | 0.1085 | 0.1085 | 0.1253 | 0.0472 | 0.086841 ± 2.3494e-005 | 0.029185 ±1.5699e-005 | 0.029185 ± 2.1085e-005 | 0.078113 ± 5.2897e-005 |
| Breast | 0.1984 | 0.1984 | 0.1494 | 0.1909 | 0.030004 ± 6.3095e-006 | 0.030545 ± 7.8554e-006 | 0.030545 ± 1.6957e-005 | 0.029496 ± 1.4263e-005 |
| DMD | 0.3589 | 0.3589 | 0.2439 | 0.0472 | 0.038967 ± 4.2896e-005 | 0.035131 ± 3.2675e-006 | 0.035131 ± 3.2675e-006 | 0.03482 ± 5.2897e-005 |
| | Computational time (Sec) | | | | | | | |
| | ICSM | DSM | CARE | CFMD | Proposed method | Variant 67 | Variant 69 | Variant 71 |
| Heart | 0.152517 | 0.076999 | 0.139614 | 204.127803 | 0.192975 | 0.268192 | 0.25357 | 0.270576 |
| RHC | 0.325134 | 0.222736 | 56.976828 | 3064.0 | 420.05 | 269.812746 | 269.8127 | 235.2842 |
| Diabetes | 0.189302 | 0.093844 | 0.302440 | 334.046758 | 1.4231 | 1.413615 | 1.413615 | 0.992357 |
| Breast | 0.468686 | 0.155074 | 19.274739 | 1004.13321 | 71.01342 | 71.111934 | 71.11193 | 82.96327 |
| DMD | 0.113384 | 0.021726 | 0.078052 | 565.664369 | 1.17265 | 1.117552 | 1.117552 | 0.991103 |

The MSE of our proposed algorithm is clearly better than CARE for the data set RHC but it does not provide a significant result as compared to ICSM, DSM, CFMD and Variant 67, 69, 71. Their average values in **Table 13** for the data set RHC are **0.1780, 0.1780, 0.3658, 0.1896, 0.25, 0.250000, 0.250000,** and **0.250000** respectively. Analogously, the proposed method has better MSE values over ICSM, DSM and CARE which are **0.086841±2.3494e-005, 0.1085, 0.1085, 0.1253** respectively whereas the Variants 67, 69 and 71 have the MSE average values are **0.029185±1.5699e-005, 0.029185±2.1085e-005,** and **0.078113±5.2897e-005** respectively on the data set Diabetes. Similarly the proposed algorithm is advantageous for the remaining two data sets Breast and DMD because the MSE values are more accurate over all other algorithms and variants. These values in the **Table 13** calculated for Breast data set are singly **0.1984, 0.1984, 0.1494, 0.1909, 0.030004±6.3095e-006, 0.030545±7.8554e-006, 0.030545±1.6957e-005,** and **0.029496±1.4263e-005**. The average MSE values computed on the DMD data set are **0.3589, 0.3589, 0.2439, 0.0472, 0.038967±4.2896e-005, 0.035131±3.2675e-006, 0.035131±3.2675e-006** and **0.03482±5.2897e-005** respectively. Overall, the average MSE values of our proposed algorithm are better than the other algorithms. These can be demonstrated clearly in the following **Figs. 1, 2, 3, 4, 5, 6,** and **7.**

The computational time of our proposed algorithm is also advantageous here in the data sets of all size. There is no such large difference in the computational time taken by our proposed method and other mentioned algorithms. From **Table 14**, it is clear that the computational time of the algorithms ICSM [59], DSM [61], CARE [10], CFMD [22], the proposed method, Variant 67, 69 and 71 are respectively **0.152517, 0.076999, 0.139614, 204.127803, 0.192975, 0.268192, 0.25357, and** 0.270576 on the data set of Heart. This scenario can also be seen on the data sets of Diabetes, and DMD. On the other hand, on the large data set RHC, the time taken our proposed method by calculation is quite large (**420.05 sec)** as compared to **ICSM (0.325134), DSM (0.222736), CARE (56.976828), Variant 67 (269.812746), Variant 69 (269.8127),** and **Variant 71 (**235.2842**).** This can be seen in **Table 14.** The analogous situation appeared for the data set of Breast.

Here in **Fig. 1,** we choose ICSM and the proposed method for the sake of simplicity because we can easily extend the rest of the algorithms by replacing the ICSM algorithm with one by one.

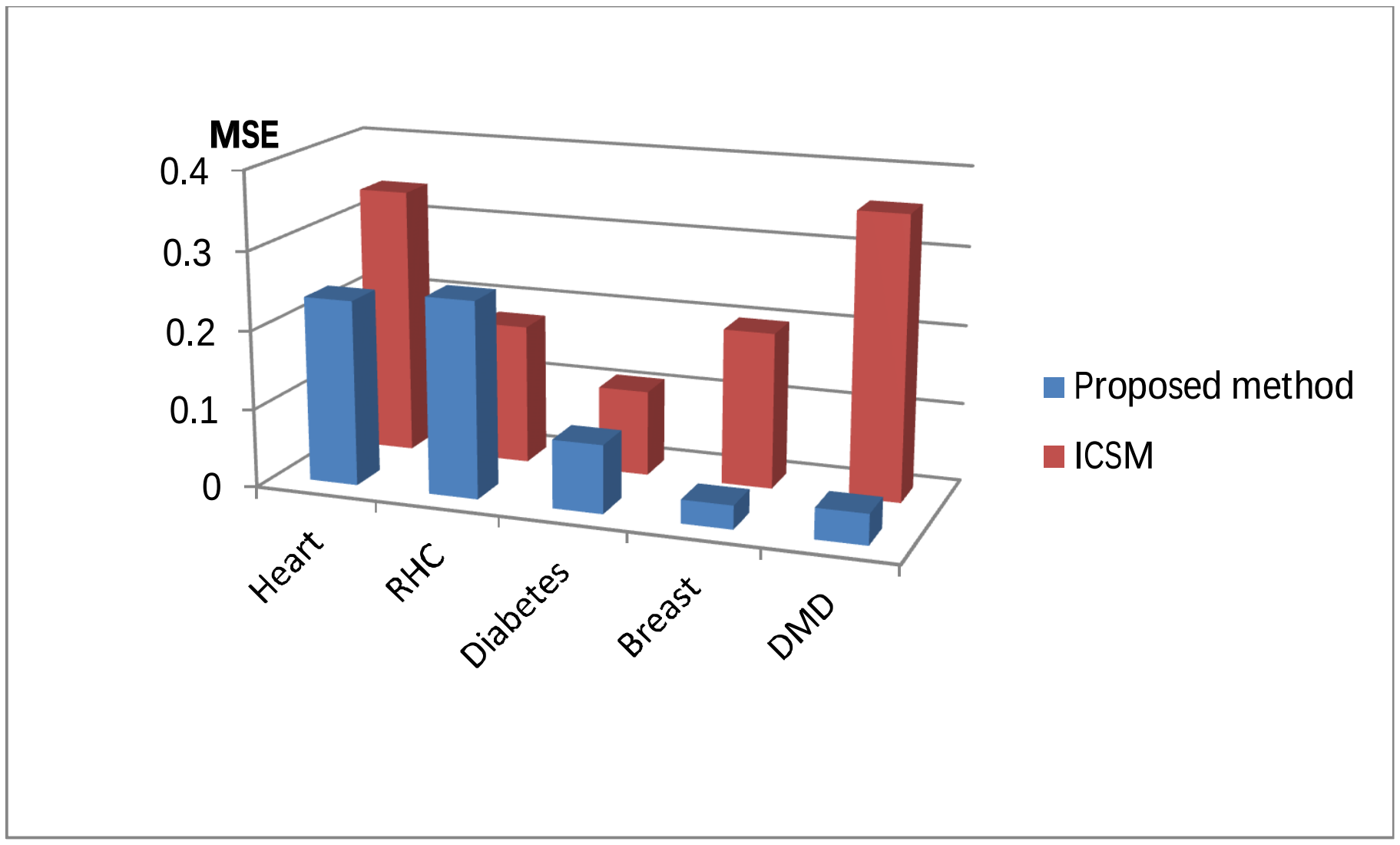

**Fig. 2. MSE (Mean Square Error) of ICSM and Proposed method**

Fig. 2 describes the MSE between **ICSM** and our **proposed method (65)** on the datasets of **Heart, RHC, Diabetes, Breast** and **DMD**. The blue bars show the MSE of **our proposed method** while the brown ones demonstrate the MSE of our **proposed ICSM**. We can see clearly that the MSE of our proposed algorithm on each dataset is smaller than the MSE of **ICSM**. Thus **Fig. 2** provides us the evidence that our proposed method can be applied to diagnose diseases in a better way with accuracy.

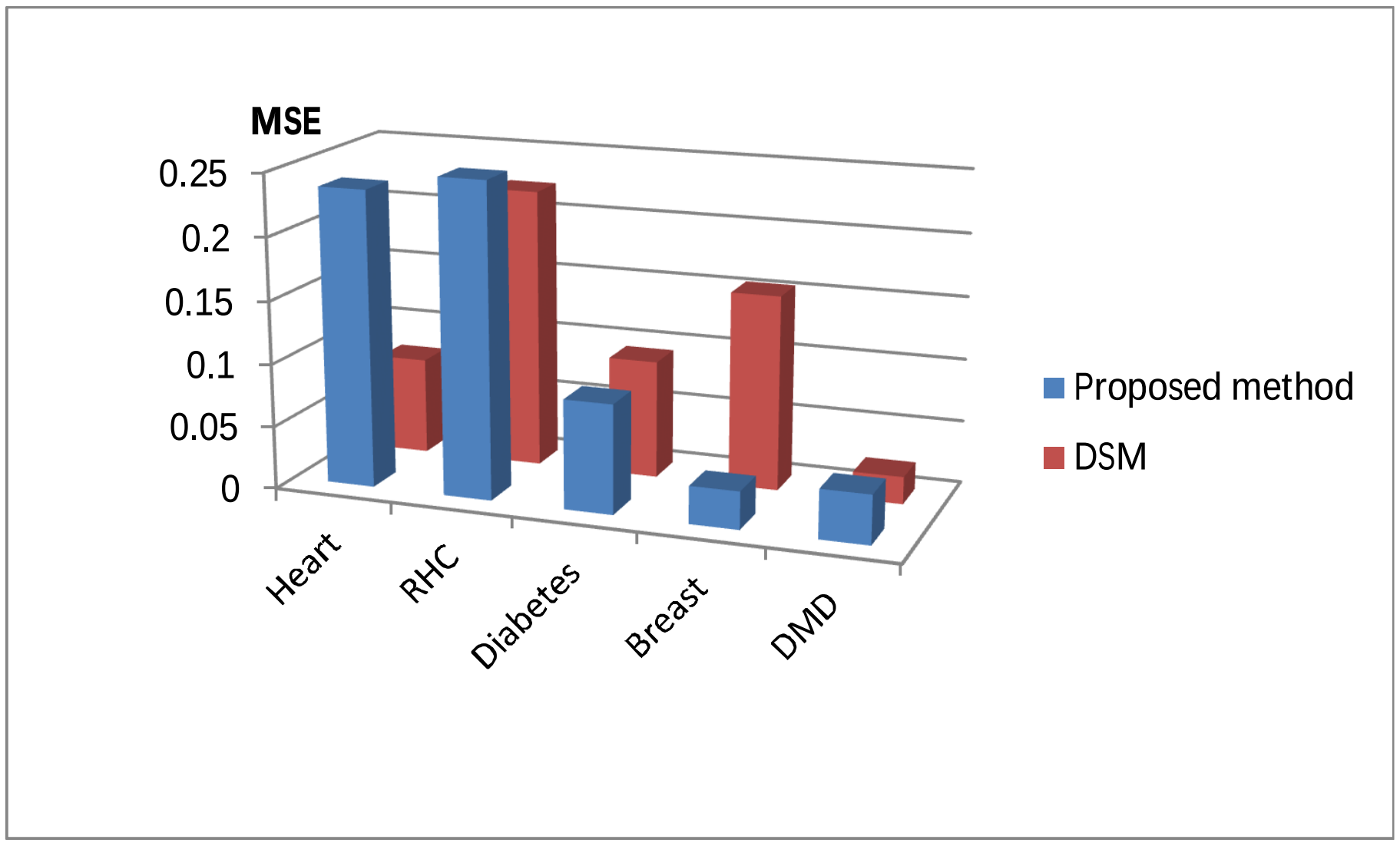

**Fig. 3. MSE (Mean Square Error) of DSM and Proposed method**

In **Fig. 3,** we can see that the MSE values of **DSM** and our **proposed method** on the datasets of **Heart, RHC, Diabetes, Breast** and **DMD**. We can see clearly that the MSE of our proposed algorithm on each dataset is better (smaller) than the MSE of **DSM**.

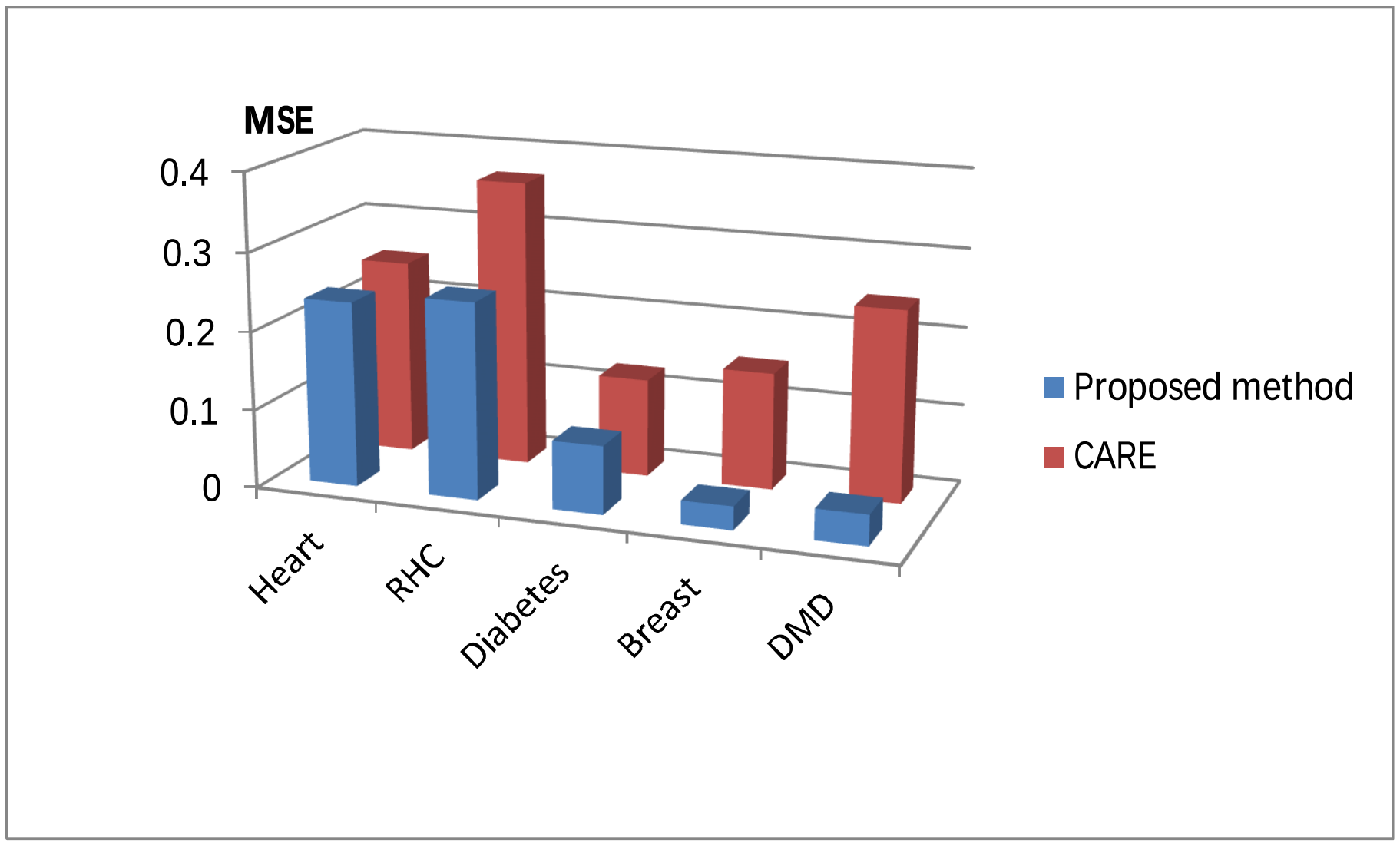

**Fig. 4. MSE (Mean Square Error) of CARE and Proposed method**

Again it is clear that the MSE of our proposed algorithm on each dataset is better (smaller) than the MSE of **CARE** which can be seen in the **Fig. 4.**

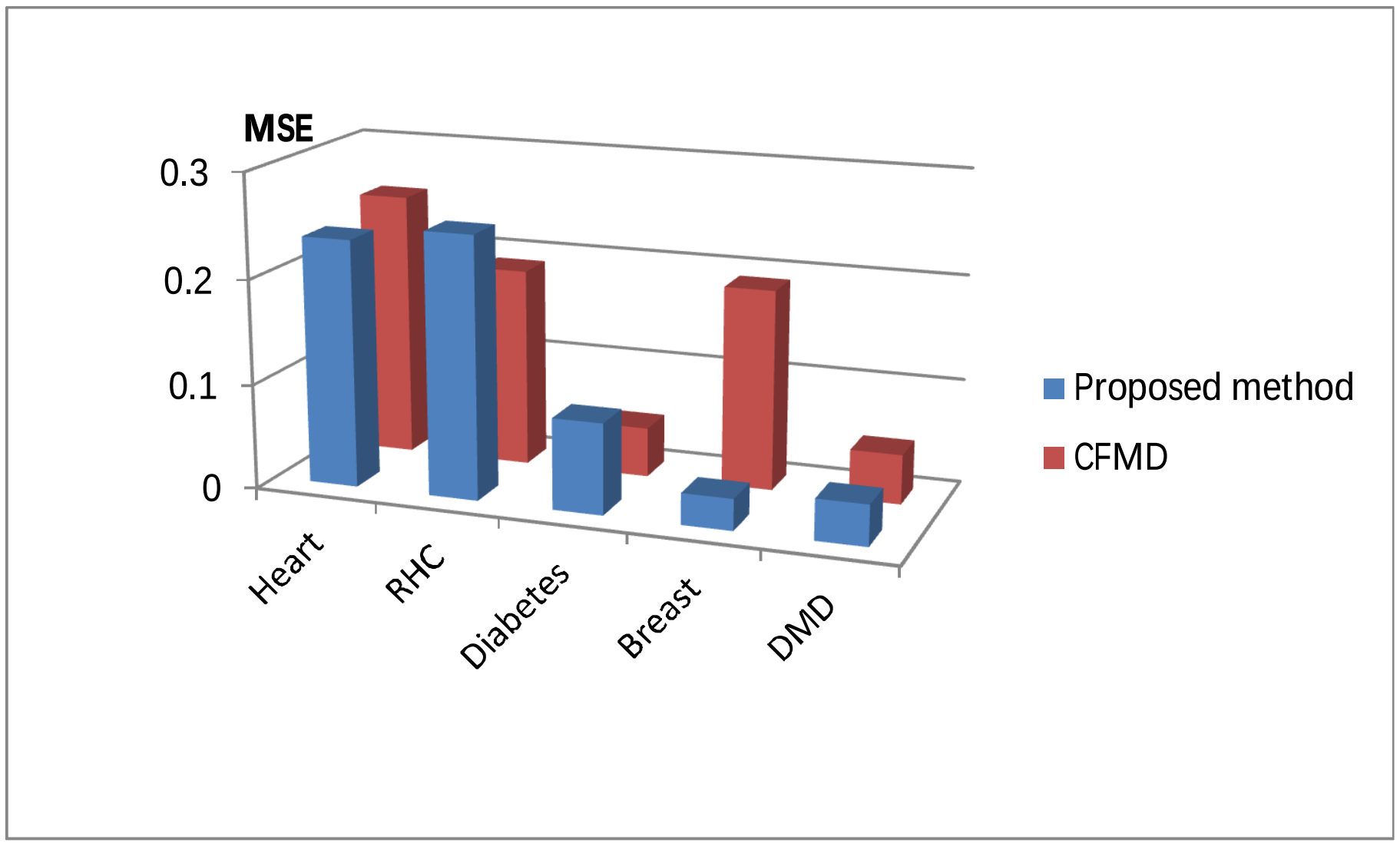

**Fig. 5. MSE (Mean Square Error) of CFMD and Proposed method**

Similarly the result is same for the CFMD as the comparison clearly demonstrates this fact in the **Fig. 5** of CFMD and proposed algorithm in the blue and brown bars.

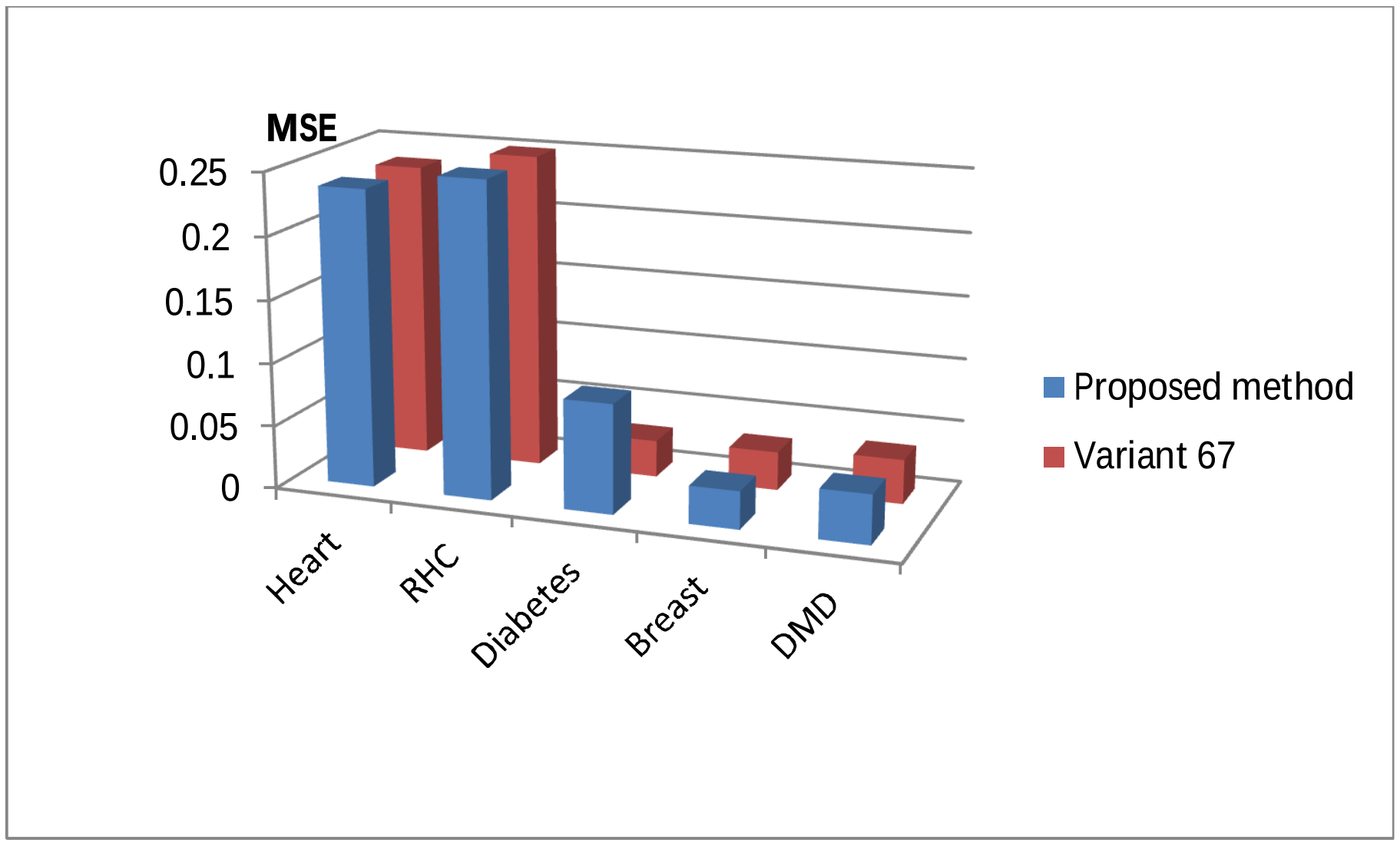

**Fig. 6. MSE (Mean Square Error) of Variant 67 and Proposed method**

In **Figs. 6, 7, 8,** the situation is almost same for the **Variant 67, 69, 71** and the **proposed method** on the datasets of **Heart, RHC, Diabetes, Breast** and **DMD**. Here the MSE values are almost equal in these algorithms which do not give any significant difference with the comparison of our proposed method.

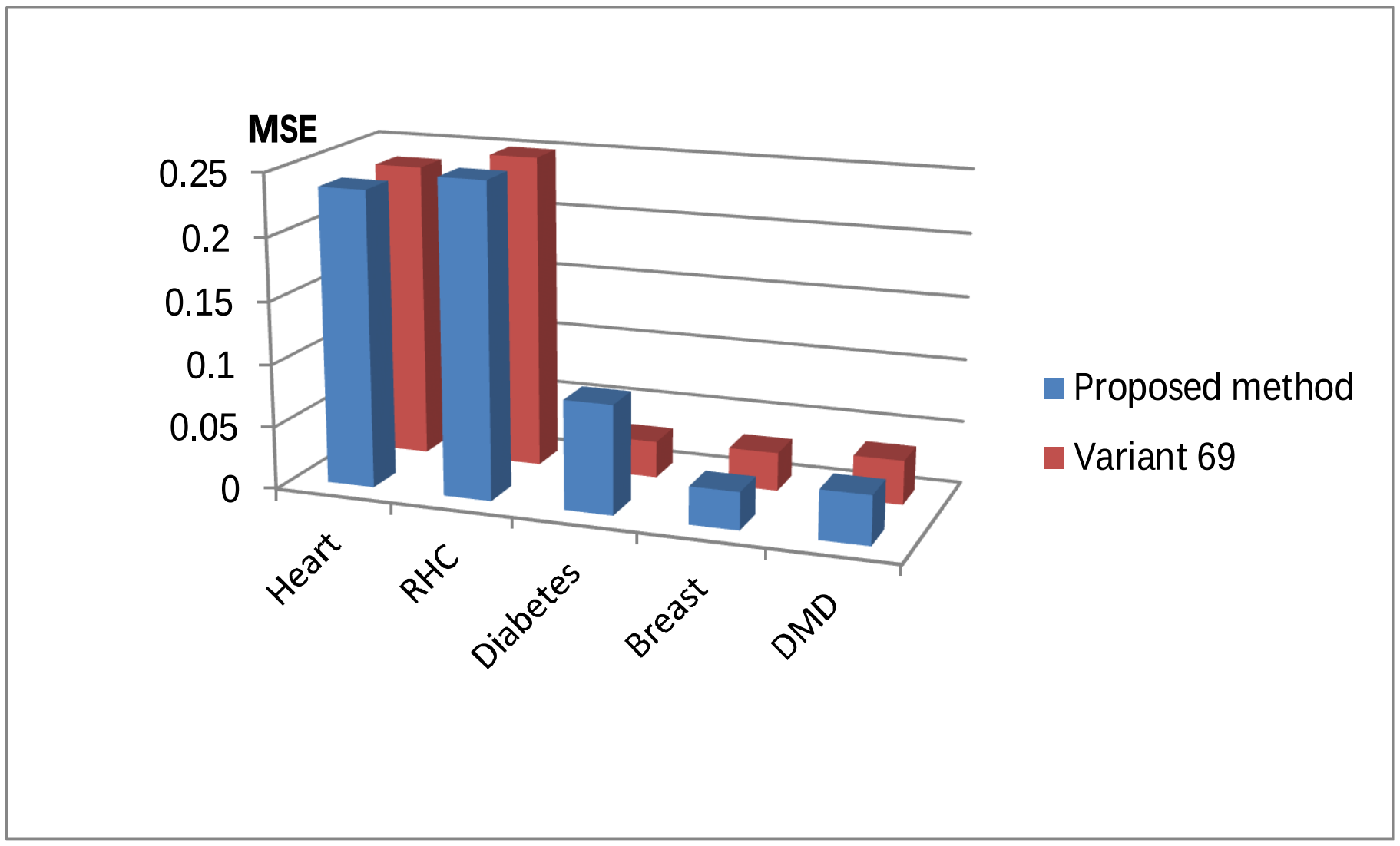

**Fig. 7. MSE (Mean Square Error) of Variant 69 and Proposed method**

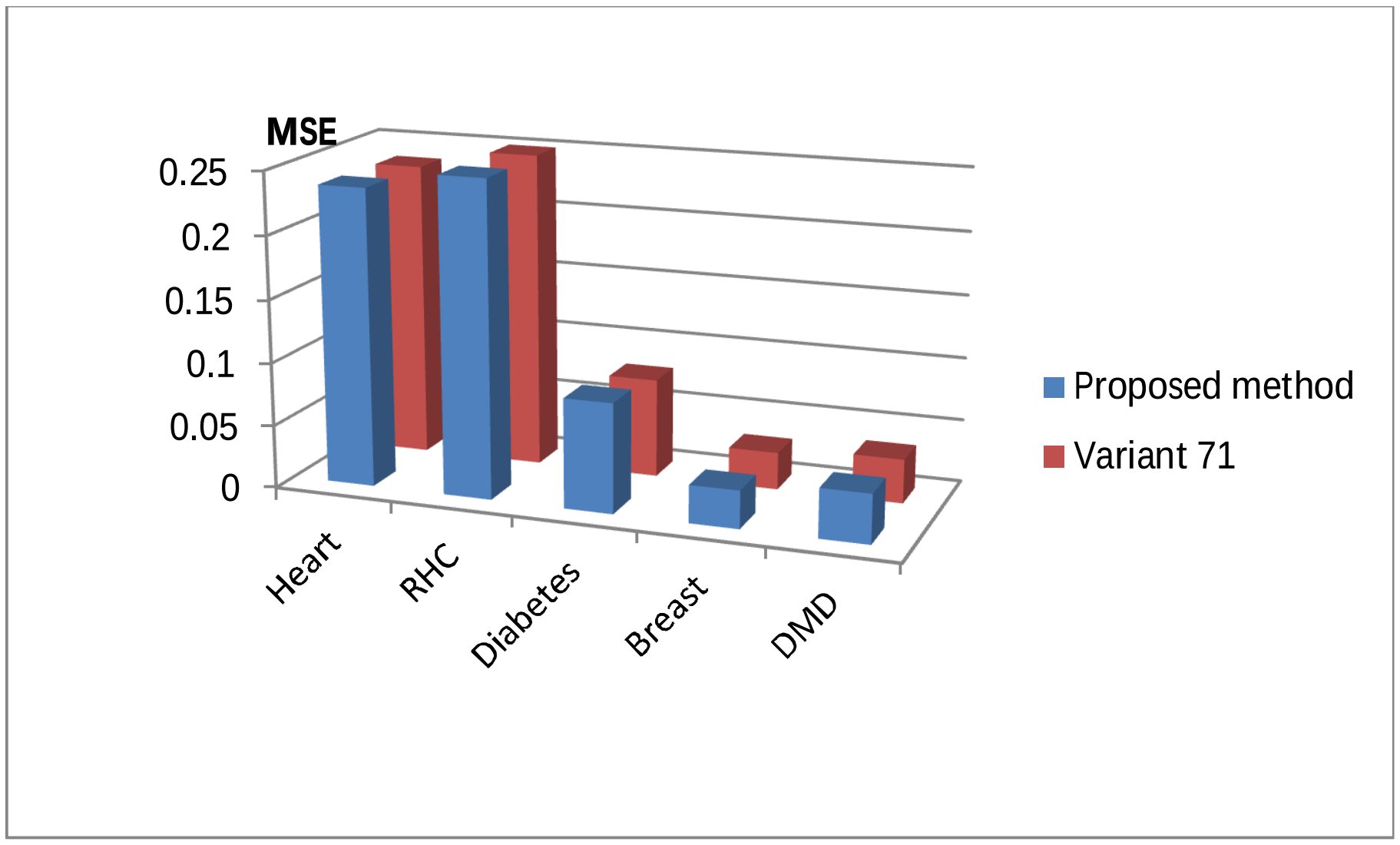

**Fig. 8. MSE (Mean Square Error) of Variant 71 and Proposed method**

In the next **Fig. 8**, we have combined all the 8 algorithms which show the MSE values and give a good comparison among them.

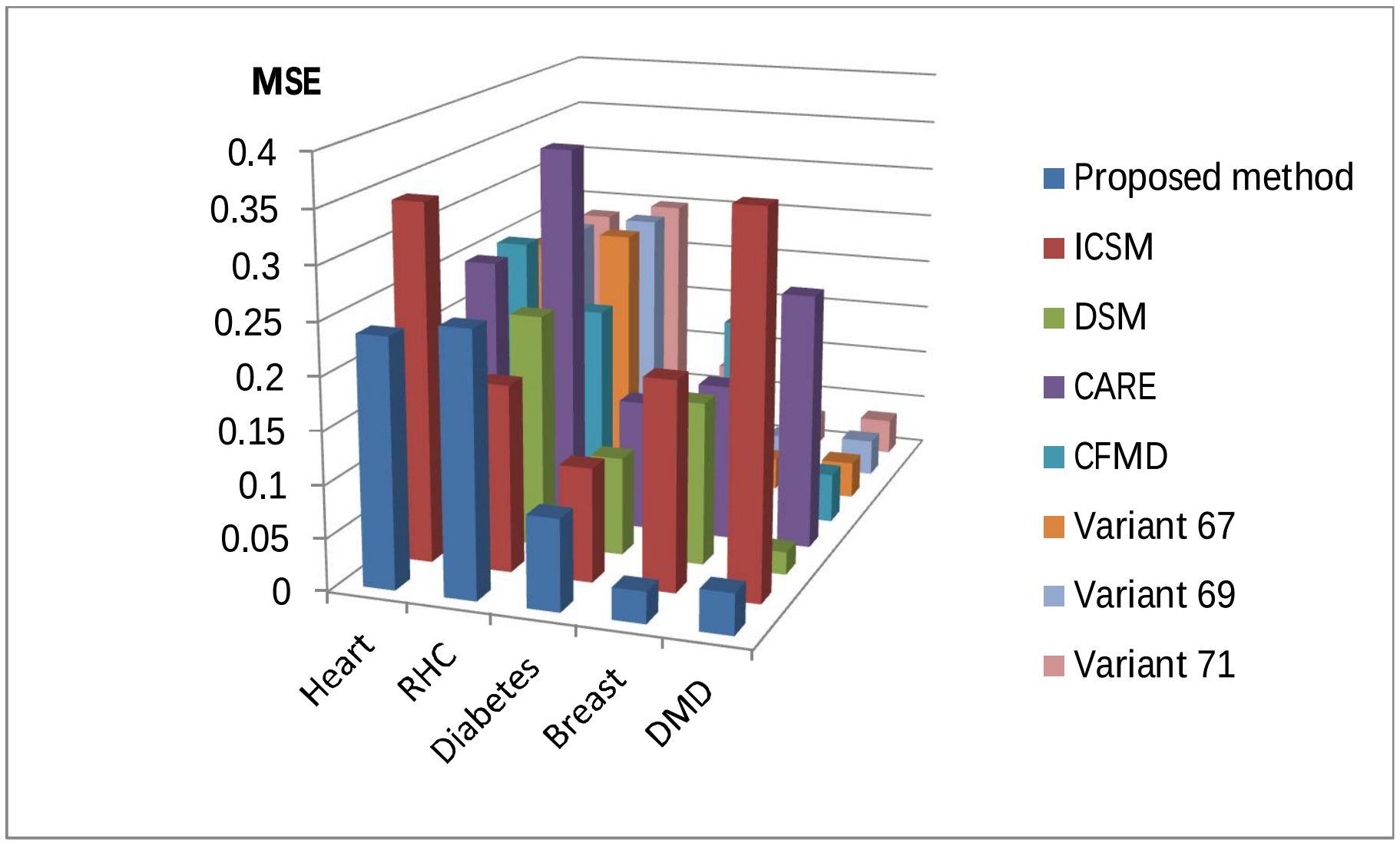

**Fig. 9. MSE (Mean Square Error) of all the 8 algorithms**

By observing this chart, one can easily see that the largest MSE value is approximately 0.35 which is provide by the algorithms ICMS and CARE while the proposed method has the least MSE value which is approximately 0.04. The Variants has also provided us some good results as compared to ICSM, DSM, CARE, and CFMD.

In the next figures, we have shown the average MSE values of all the 7 algorithms on the 5 datasets in the line graphs. First, we gave the comparison of our proposed method with other 7 algorithms one by one in the **Figs. 10, 11, 12, 13, 14, 15** respectively and then combine them in one line graph in the **Fig. 16**.

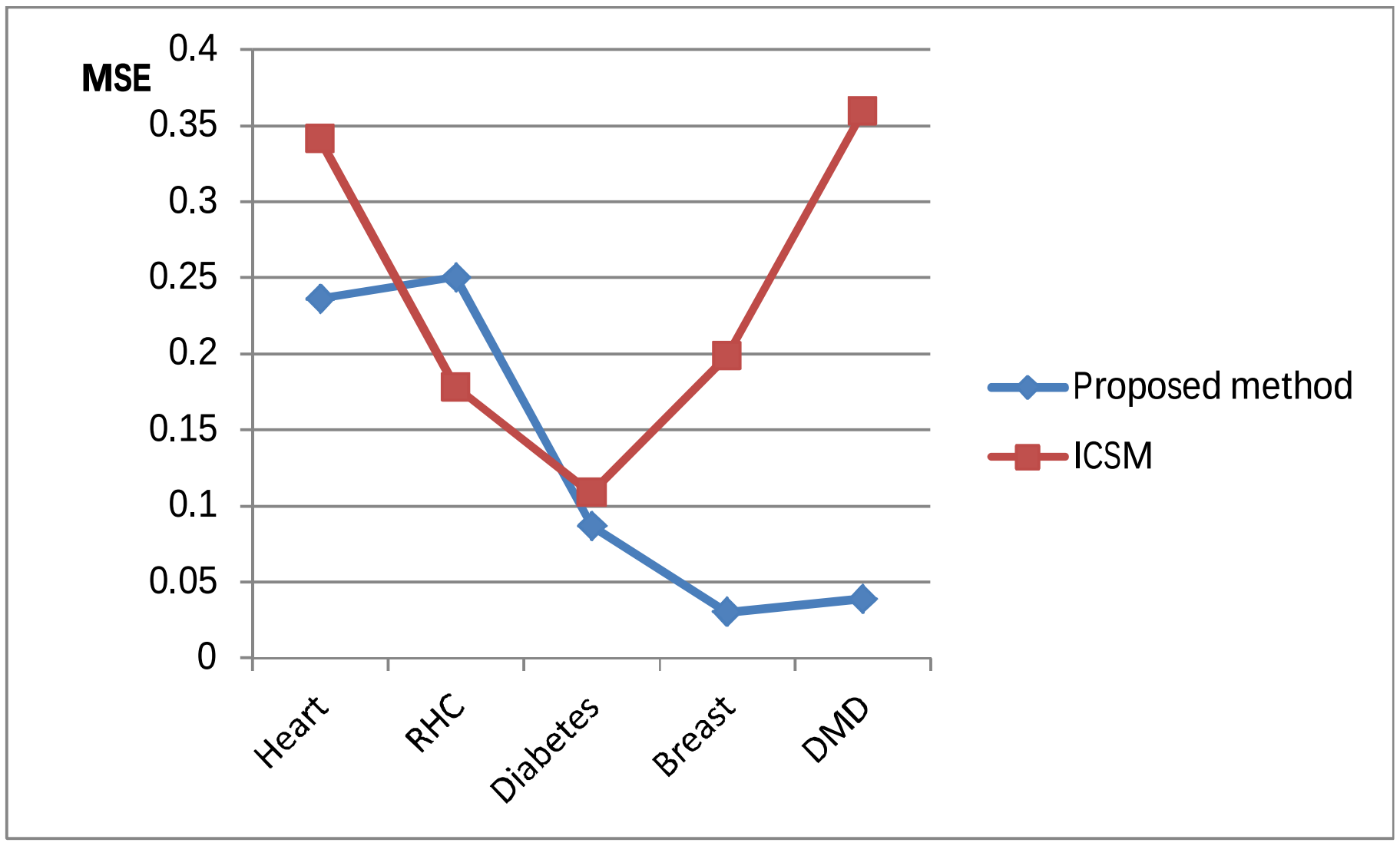

**Fig. 10. MSE (Mean Square Error) of ICSM and Proposed method**

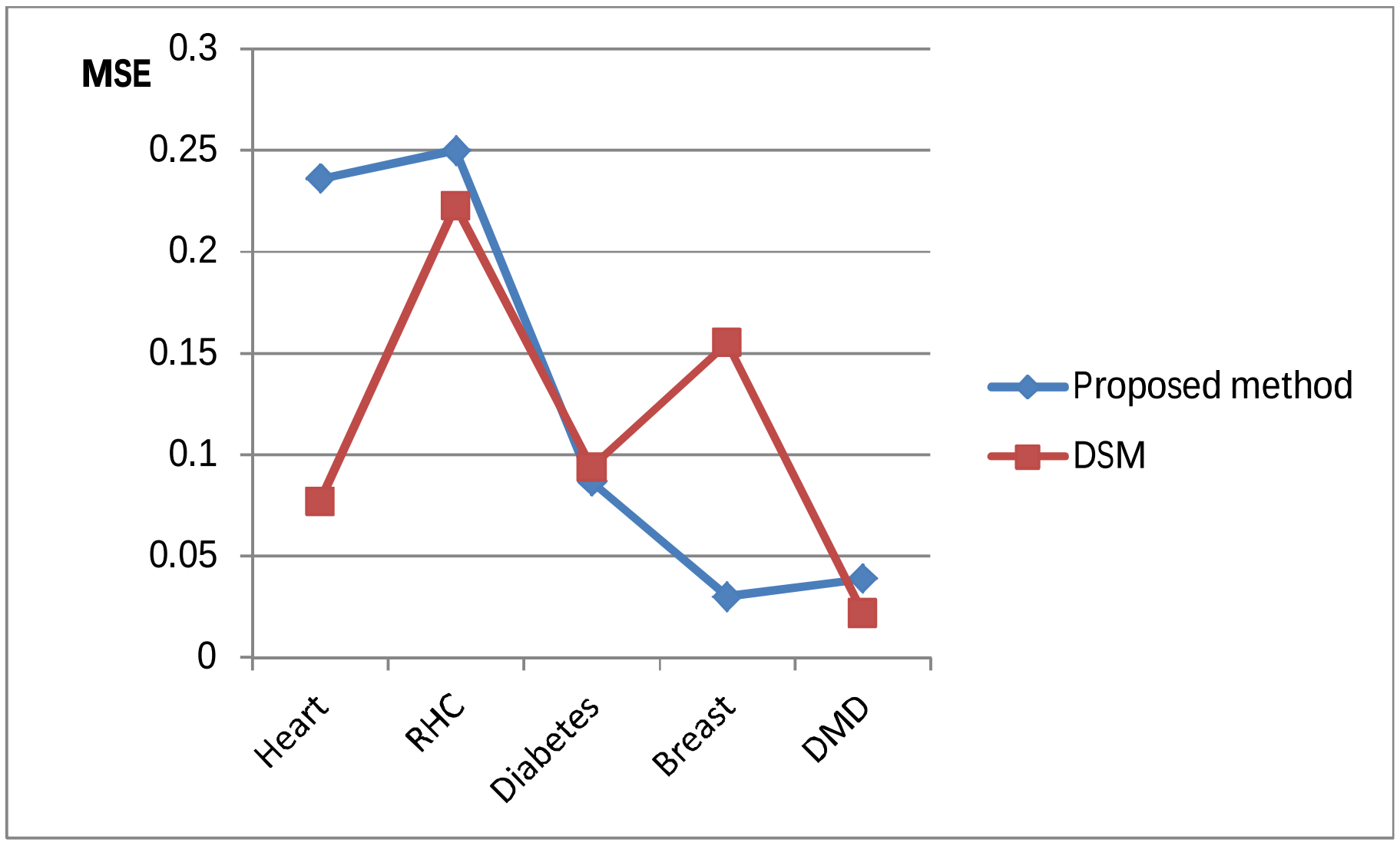

**Fig. 11. MSE (Mean Square Error) of DSM and Proposed method**

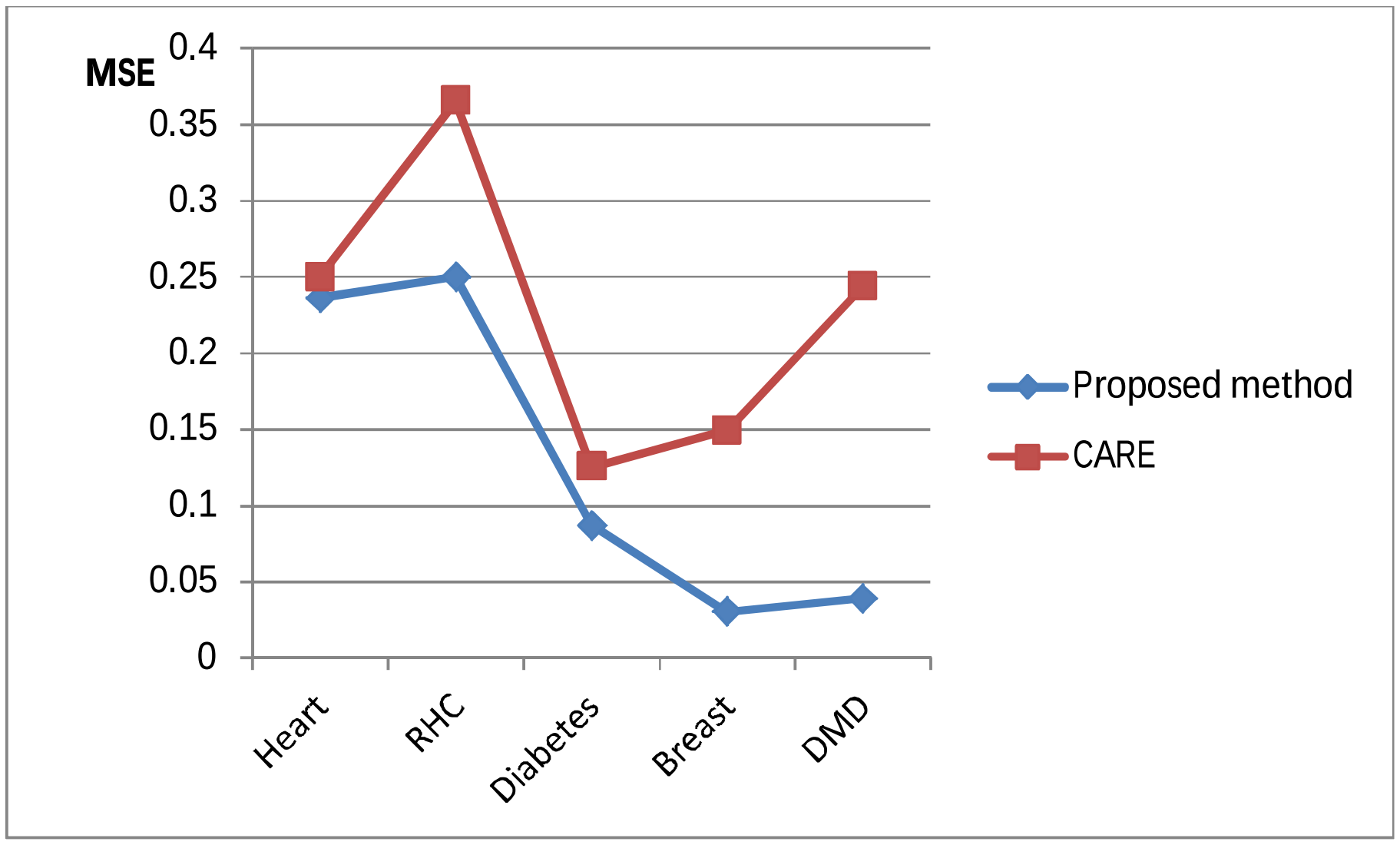

**Fig. 12. MSE (Mean Square Error) of CARE and Proposed method**

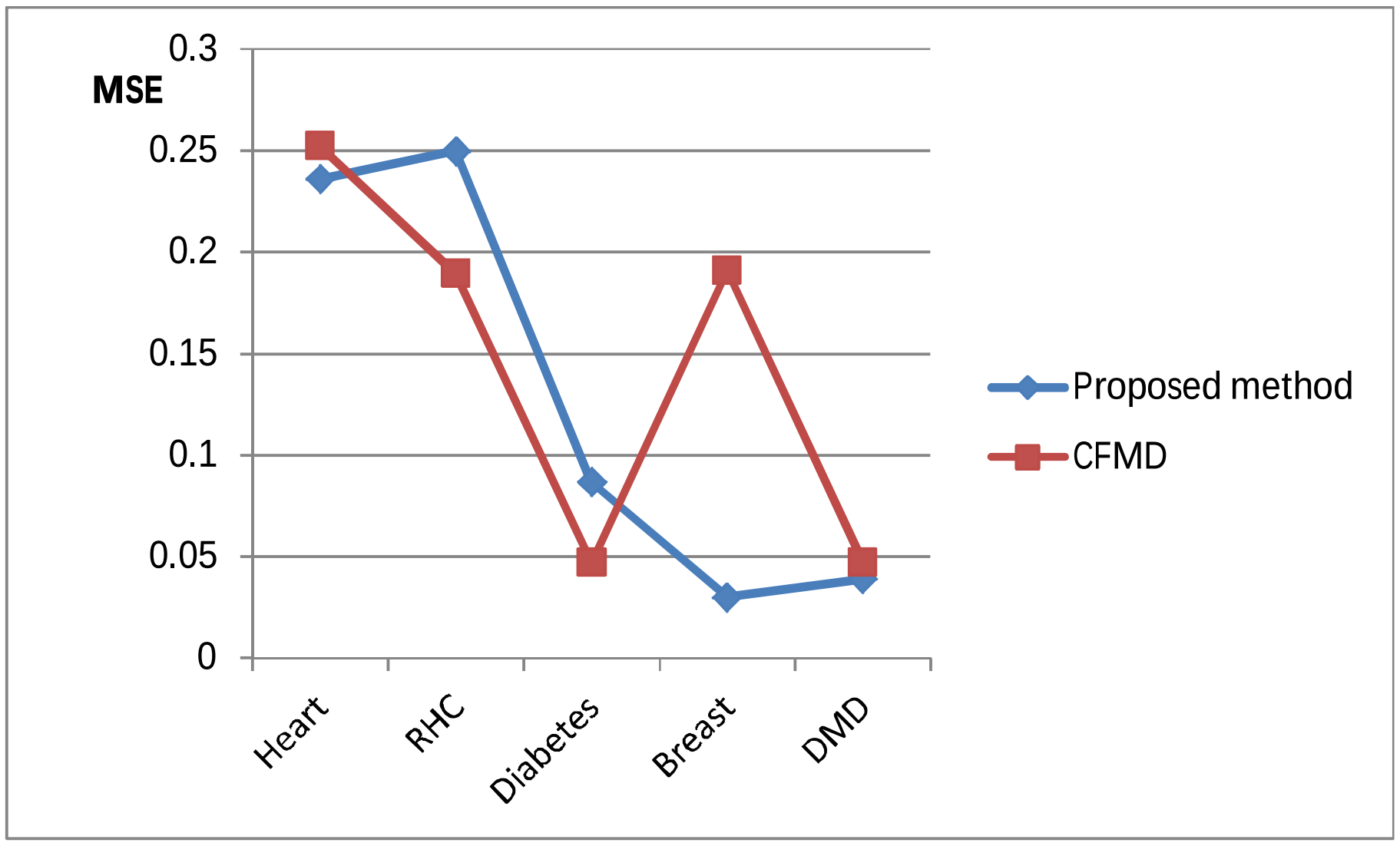

**Fig. 13. MSE (Mean Square Error) of CFMD and Proposed method**

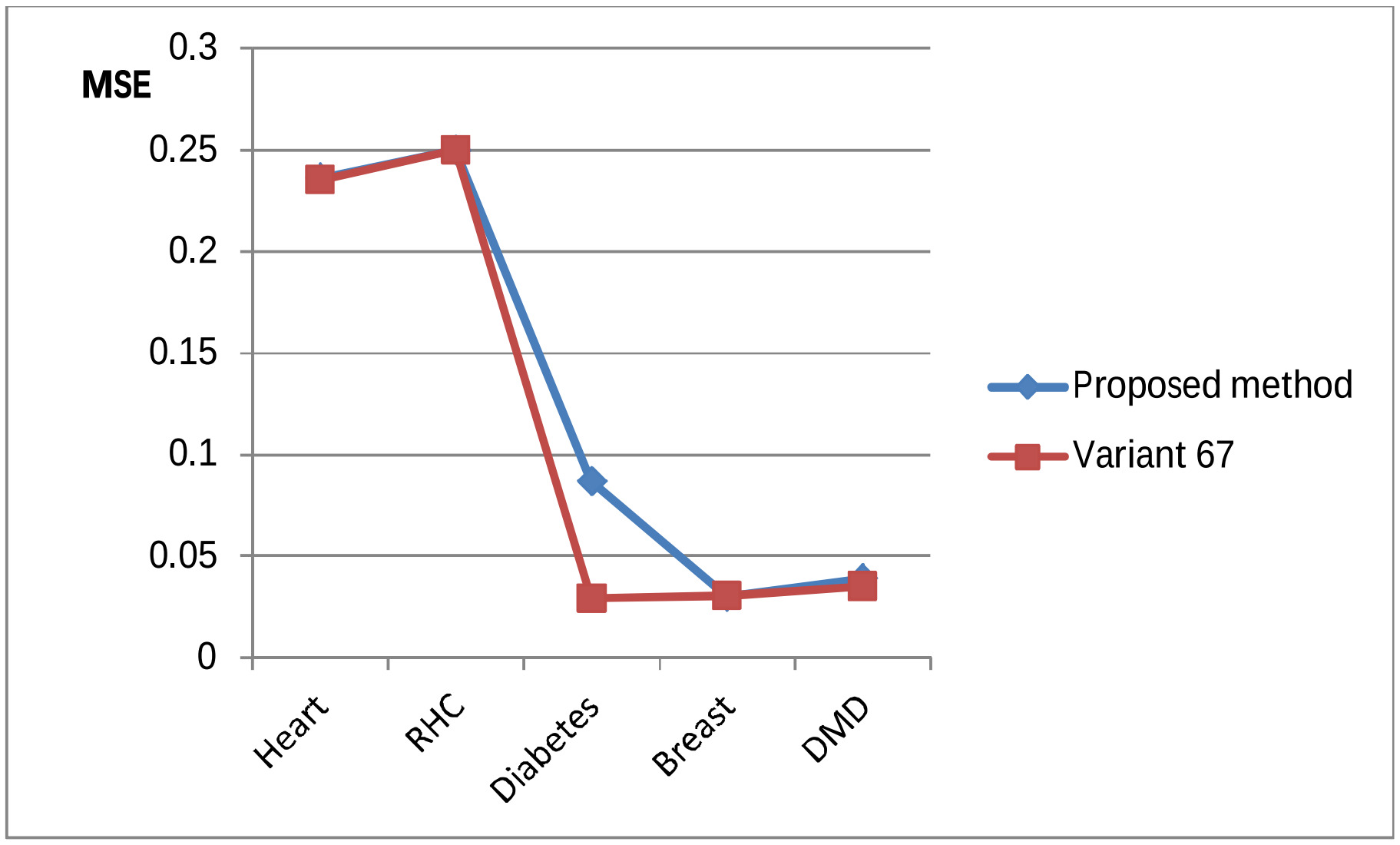

**Fig. 14. MSE (Mean Square Error) of Variant 67 and Proposed method**

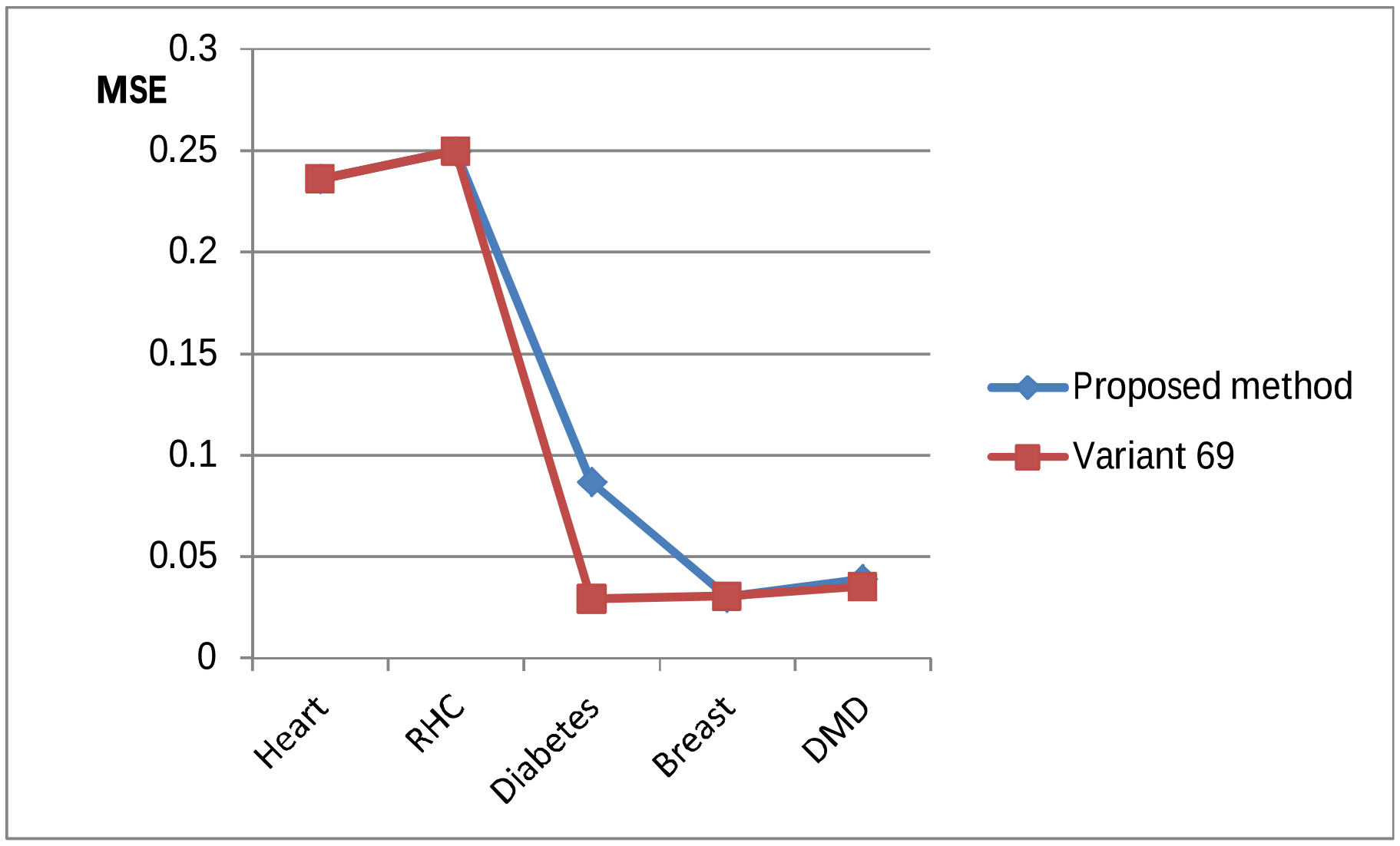

**Fig. 15. MSE (Mean Square Error) of Variant 69 and Proposed method**

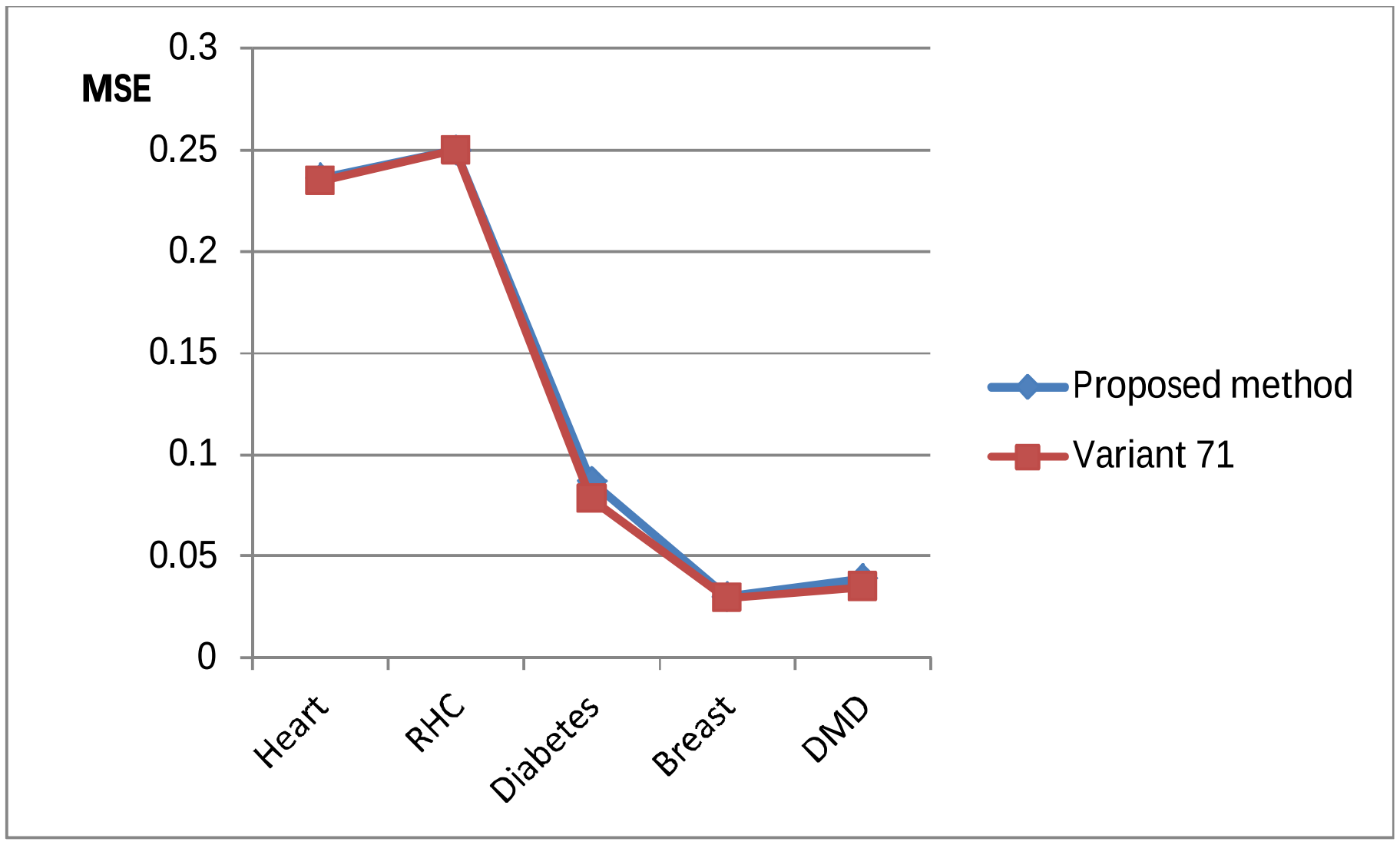

**Fig. 16. MSE (Mean Square Error) of Variant 71 and Proposed method**

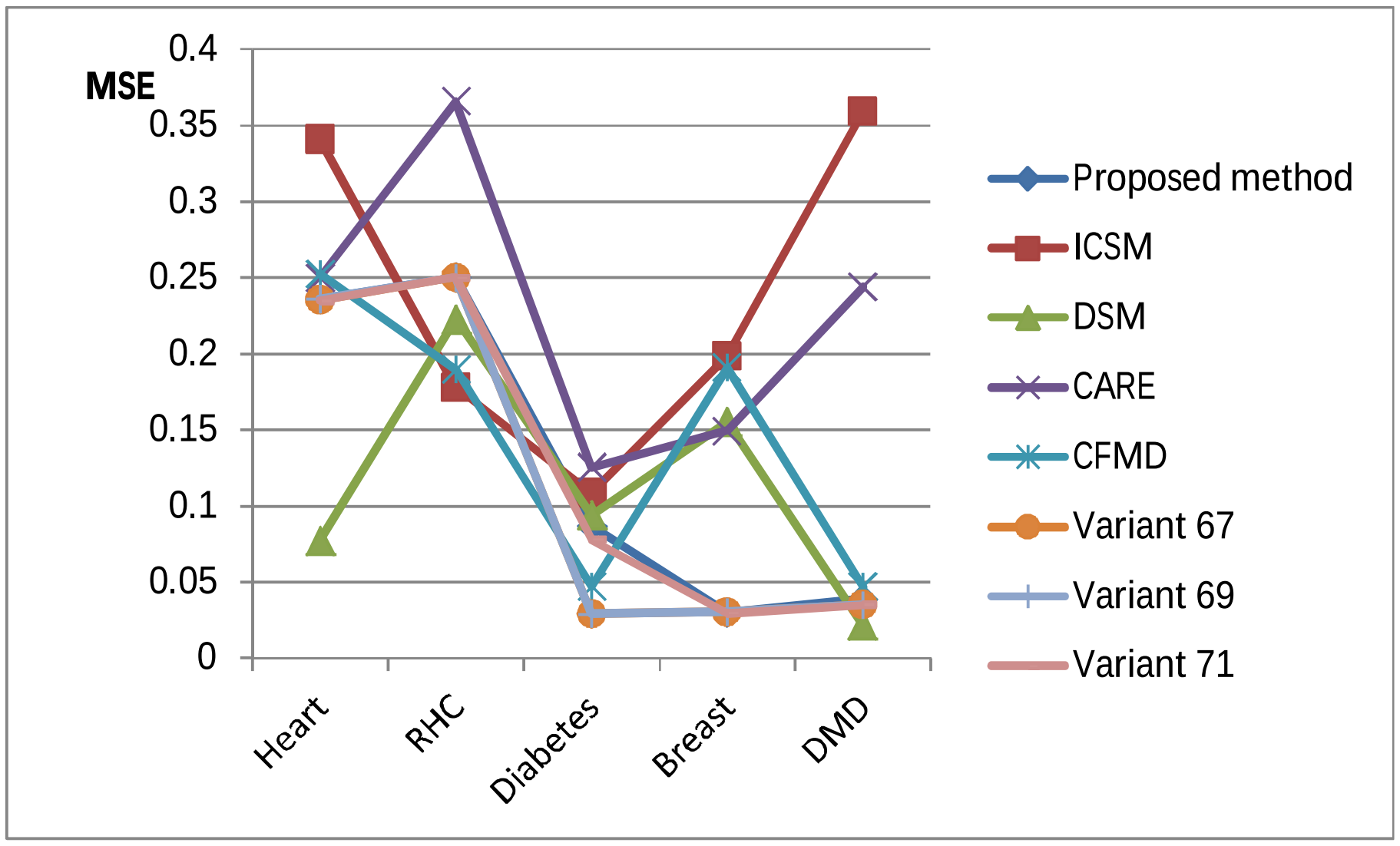

**Fig. 17. MSE (Mean Square Error) of all the 8 algorithms**

Now to analyze these algorithms deeply, we provide some more detail by drawing the average MSE values of all the 8 algorithms on a single data set. In the **Figs. 18, 19, 20, 21, 22** and **23**, we presented the MSE of all datasets for 8 algorithms including our proposed one to analyze them in a more detailed and comprehensive way.

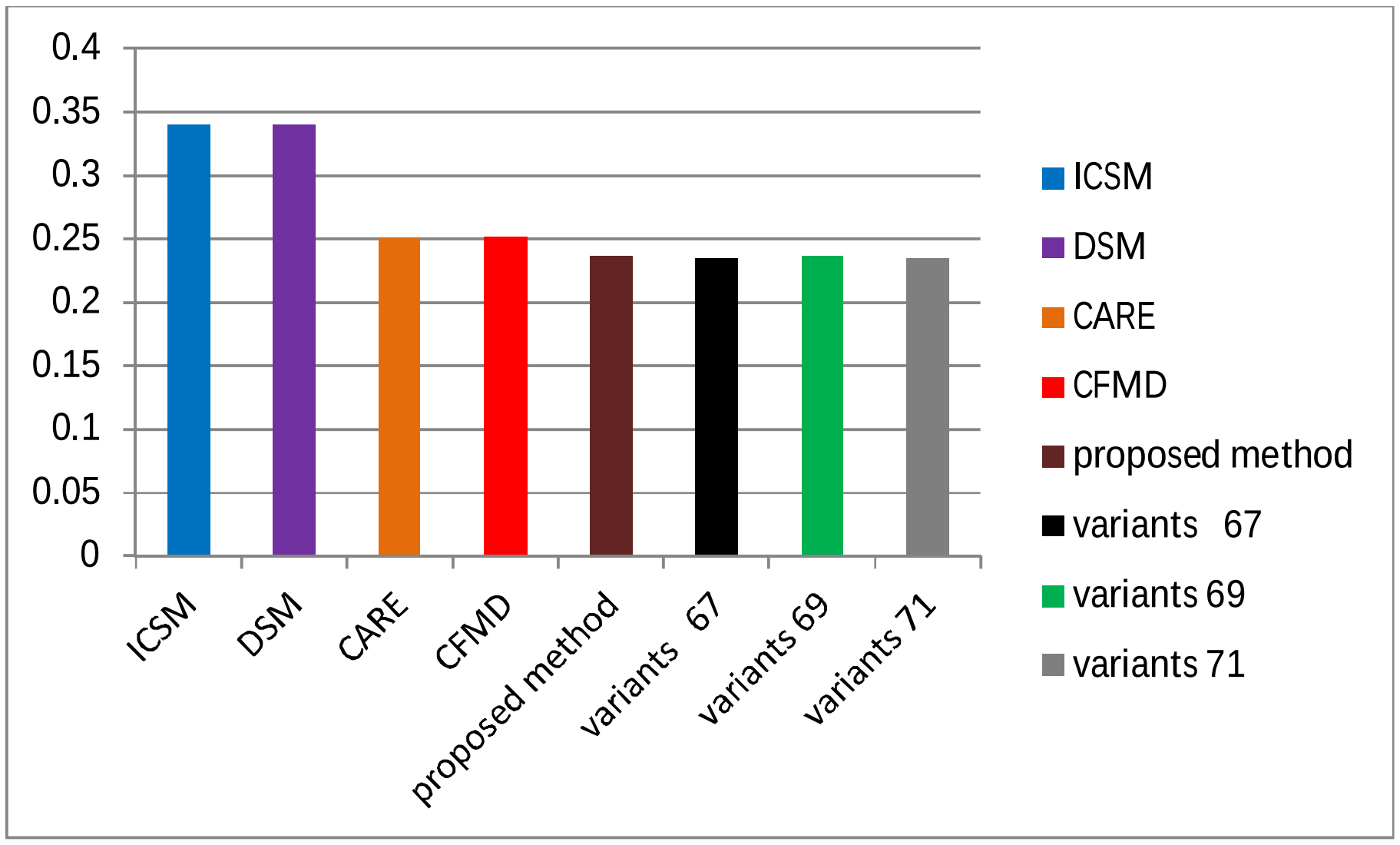

**Fig. 18. MSE of Heart dataset between 8 algorithms**

From **Table 14** and **Fig. 18**, we can easily see that the MSE of our proposed method (approximately 0.23) is much smaller than the MSE of ICSM, DSM (approximately 0.35), CARE and CFMD (approximately 0.25) on the Heart dataset. Therefore our proposed algorithm is advantageous over the previous mentioned algorithms on the Heart dataset. On the other hand, the proposed algorithm has not significance MSE variation as compared to the Variants 67, 69 and 71 (approximately 0.24).

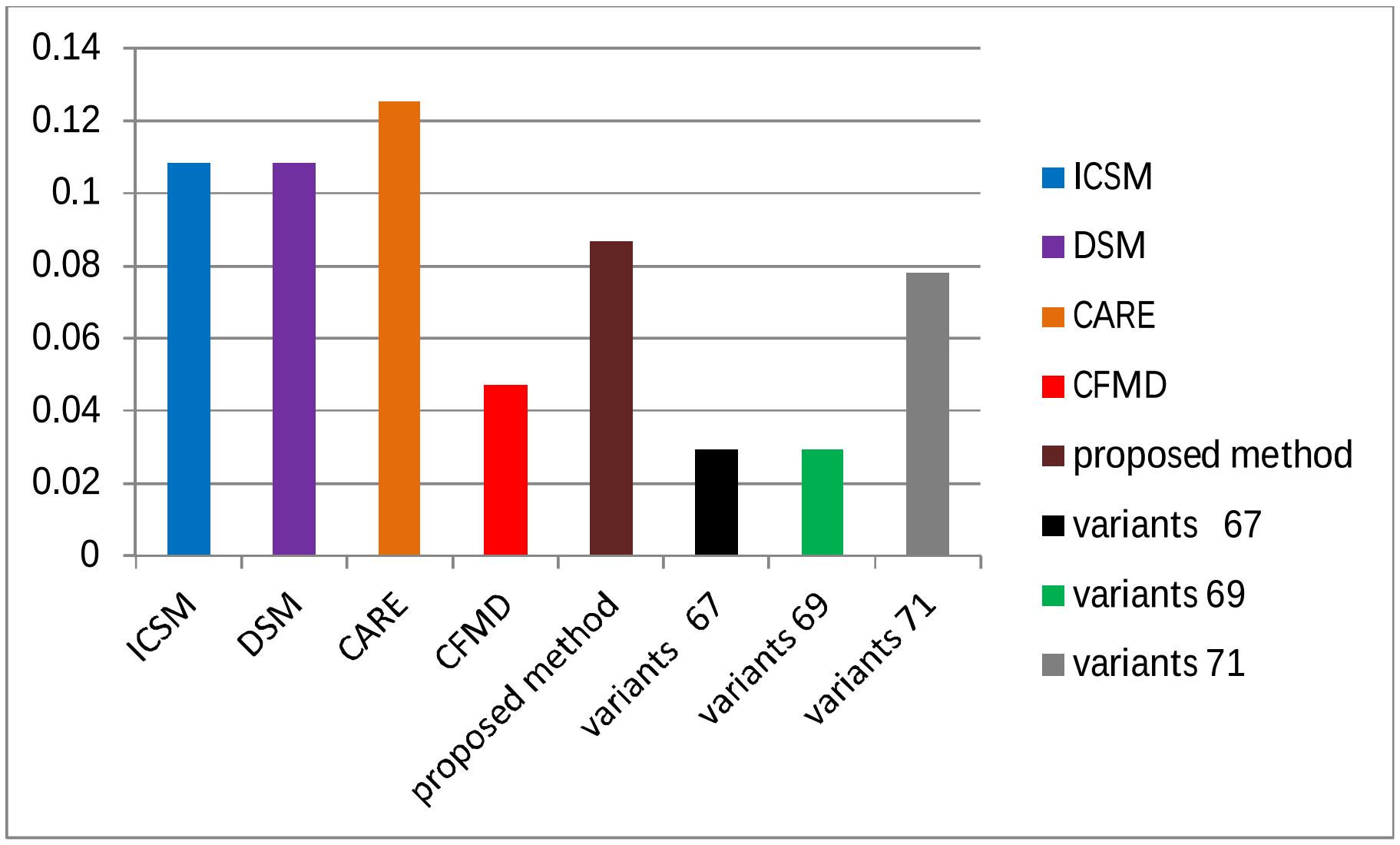

**Fig. 19. MSE of Diabetes dataset between 8 algorithms**

Here in **Fig. 19**, the result is same that the proposed algorithm (approx. 0.09) has MSE is smaller than ICSM, DSM, and CARE (approx. 0.12) while the situation is reverse in the case of CFMD (approx.. 0.05), Variant 67, Variant 69 (approx. 0.03) and Variant 71(approx. 0.08) on the dataset of Diabetes.

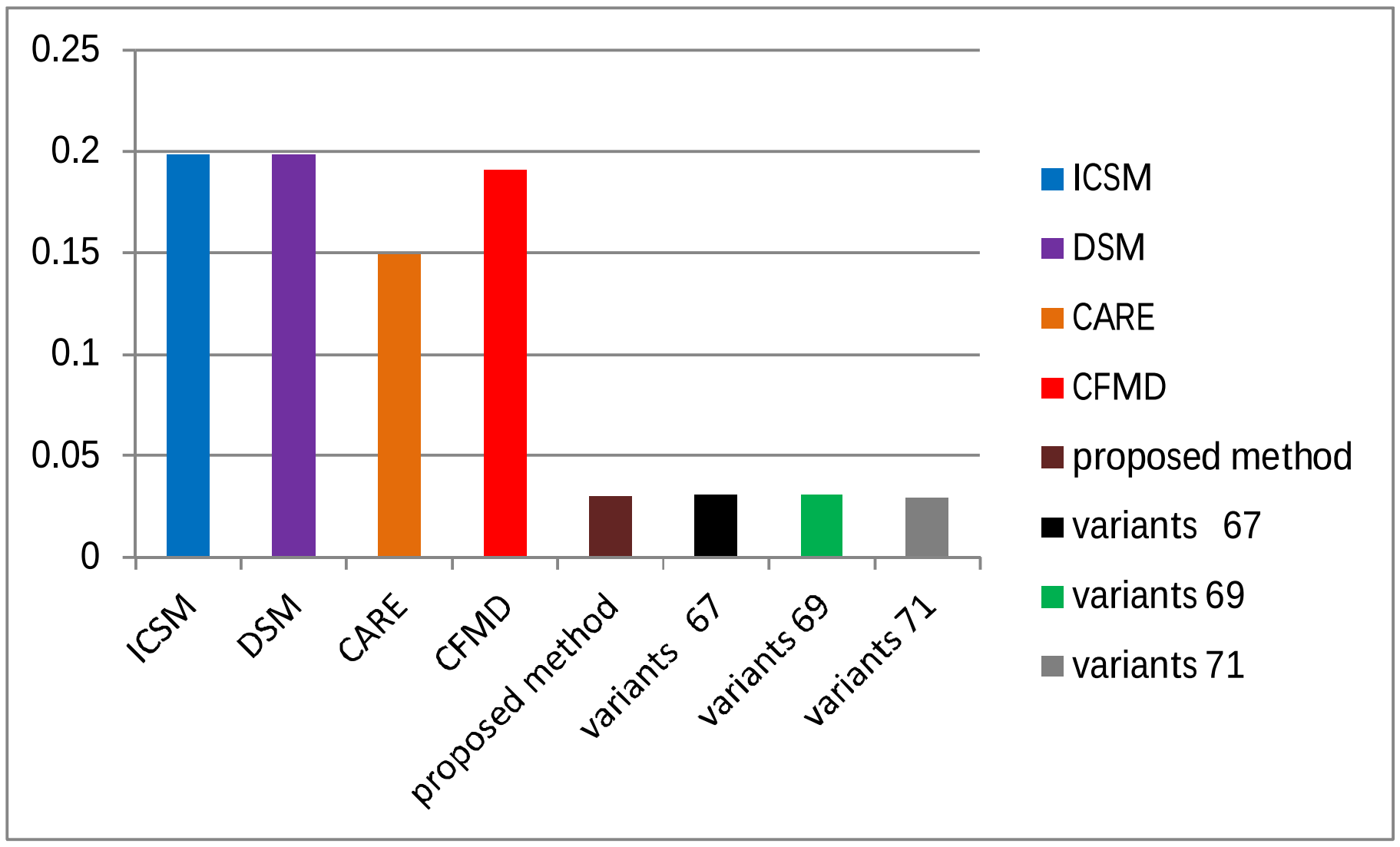

**Fig. 20. MSE of Breast dataset between 8 algorithms**

From **Fig. 20**, it is clear that the MSE of our proposed method and the variants (approximately 0.03) is much smaller than the MSE of ICSM, DSM, CARE and CFMD (approximately 0.2) on the Heart dataset. Again our proposed algorithm is advantageous over ICSM, DSM, CARE and CFMD on the Breast dataset.

**Fig. 21** demonstrates that again the proposed algorithm surpasses over other methods. The MSE of proposed method and the variants (approximately 0.03) is much smaller as compared to the MSE of ICSM, DSM (approx. 0.35), CARE (approx. 0.25) whereas CFMD (approximately 0.05) is comparably the same MSE as that of proposed method on the DMD dataset.

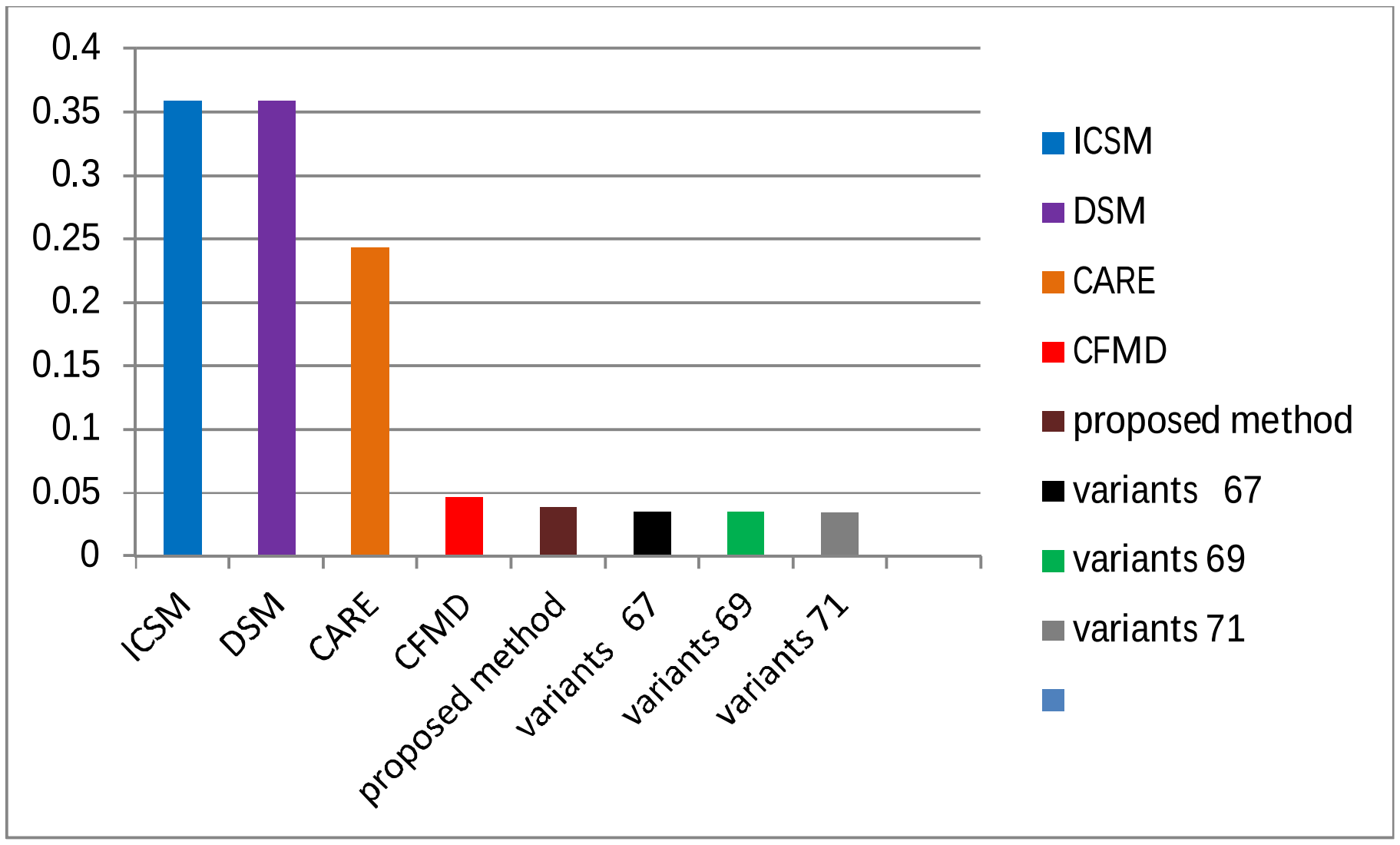

**Fig. 21. MSE of DMD dataset between 8 algorithms**

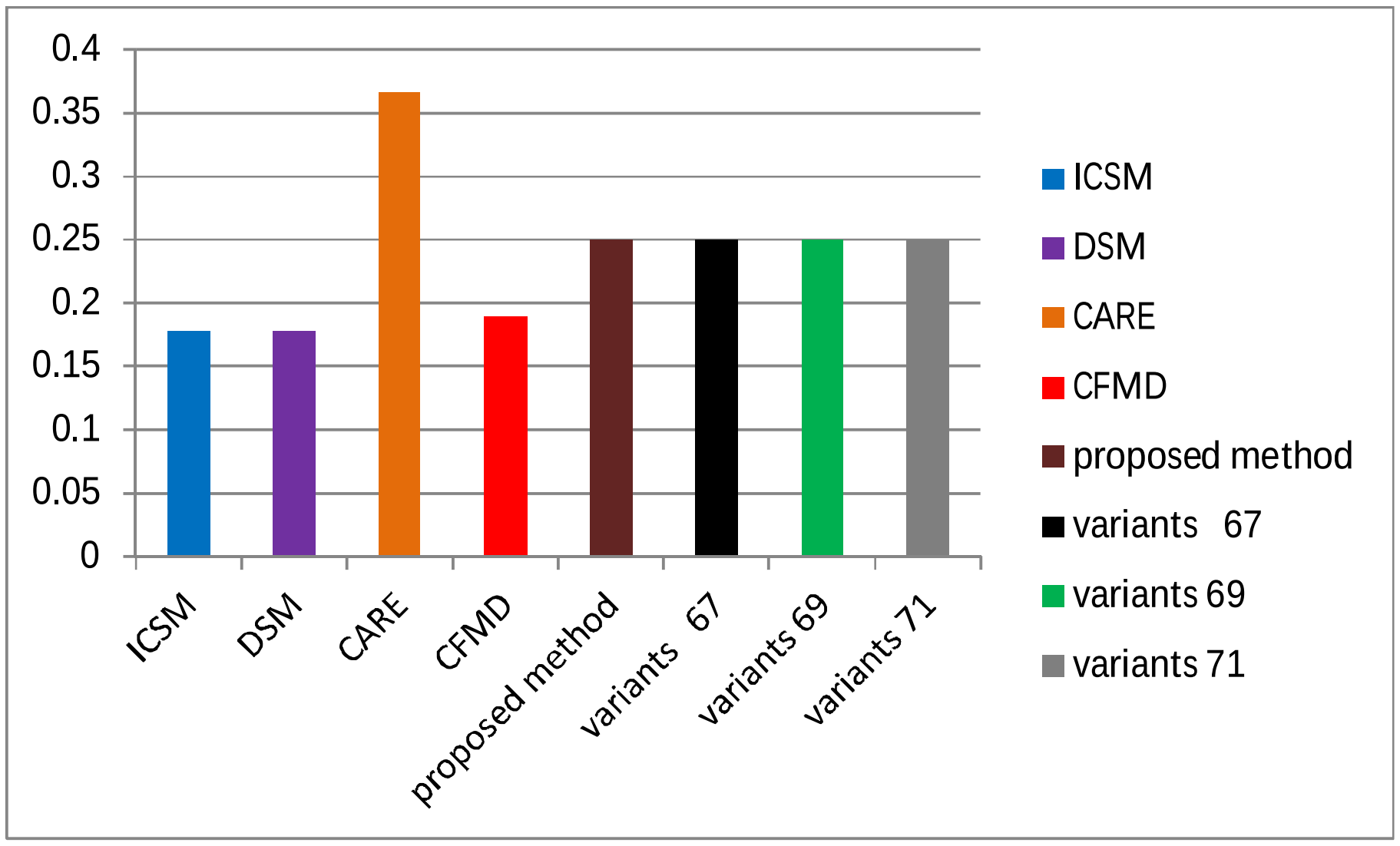

**Fig. 22. MSE of RHC dataset between 8 algorithms**

On the other, the large dataset RHC, one can see that MSE of ICSM, DSM and CFMD (approx. 0.18) is smaller than the MSE of our proposed method and the variants (approx. 0.25) in **Fig. 22**, while CARE has quite larger MSE value as compared to all other 7 algorithms in this **Fig. 22** on the large dataset RHC.

Overall in all 5 typical datasets, we can observed that the MSE result of RHC dataset is not very well due to its large size which has much noise and choosing random parameters  is not fit for our proposed model. This is a big drawback of our proposed method.

In the next **figures**, we have analyze all of the 8 algorithms including the proposed method on the 5 datasets Heart, RHC, Breast Cancer, Diabetes and DMD of average MSE values in the line graphs which provide a more detailed explanation to see the behavior of the algorithms.

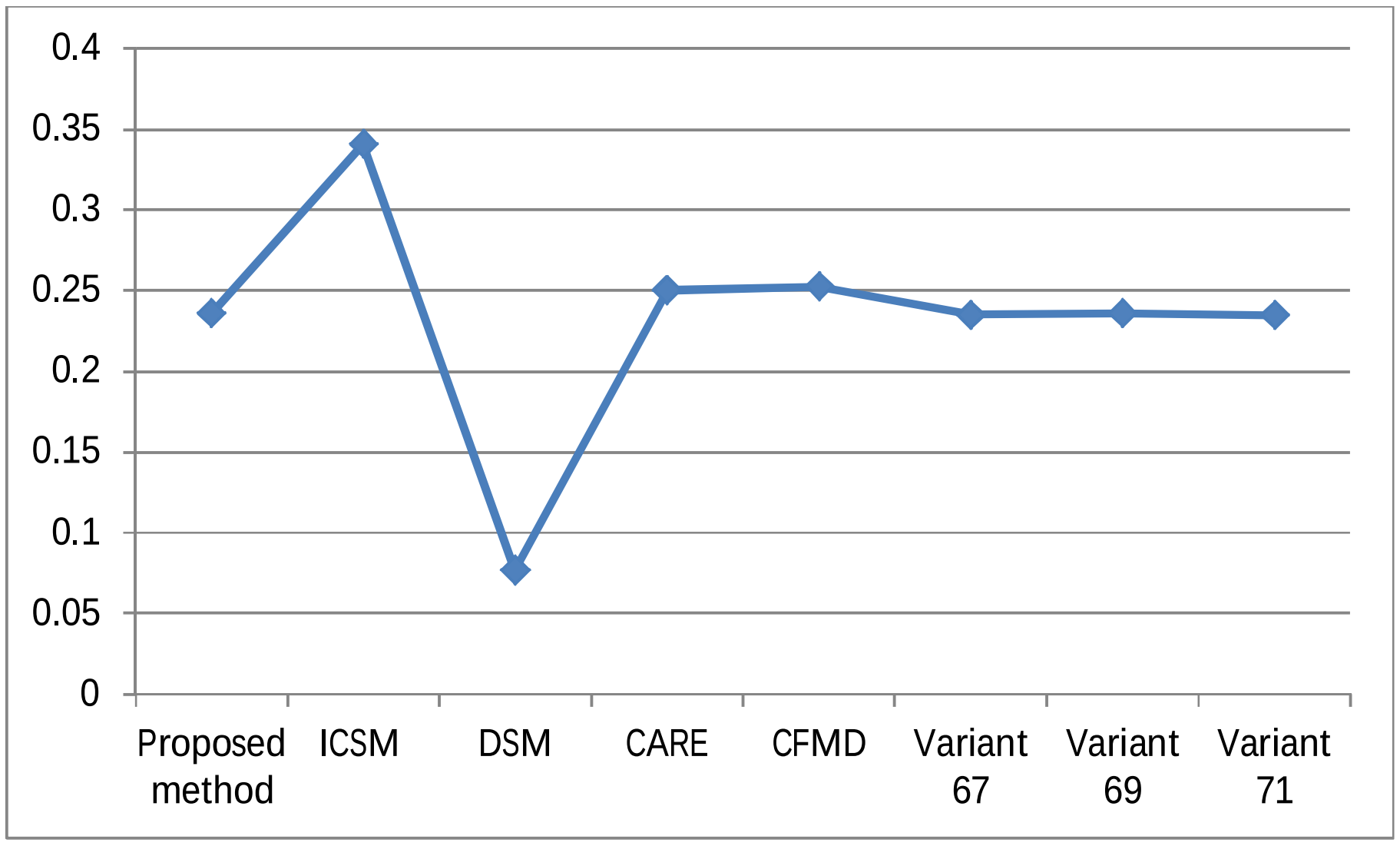

**Fig. 23.  MSE of Heart dataset between 8 algorithms**

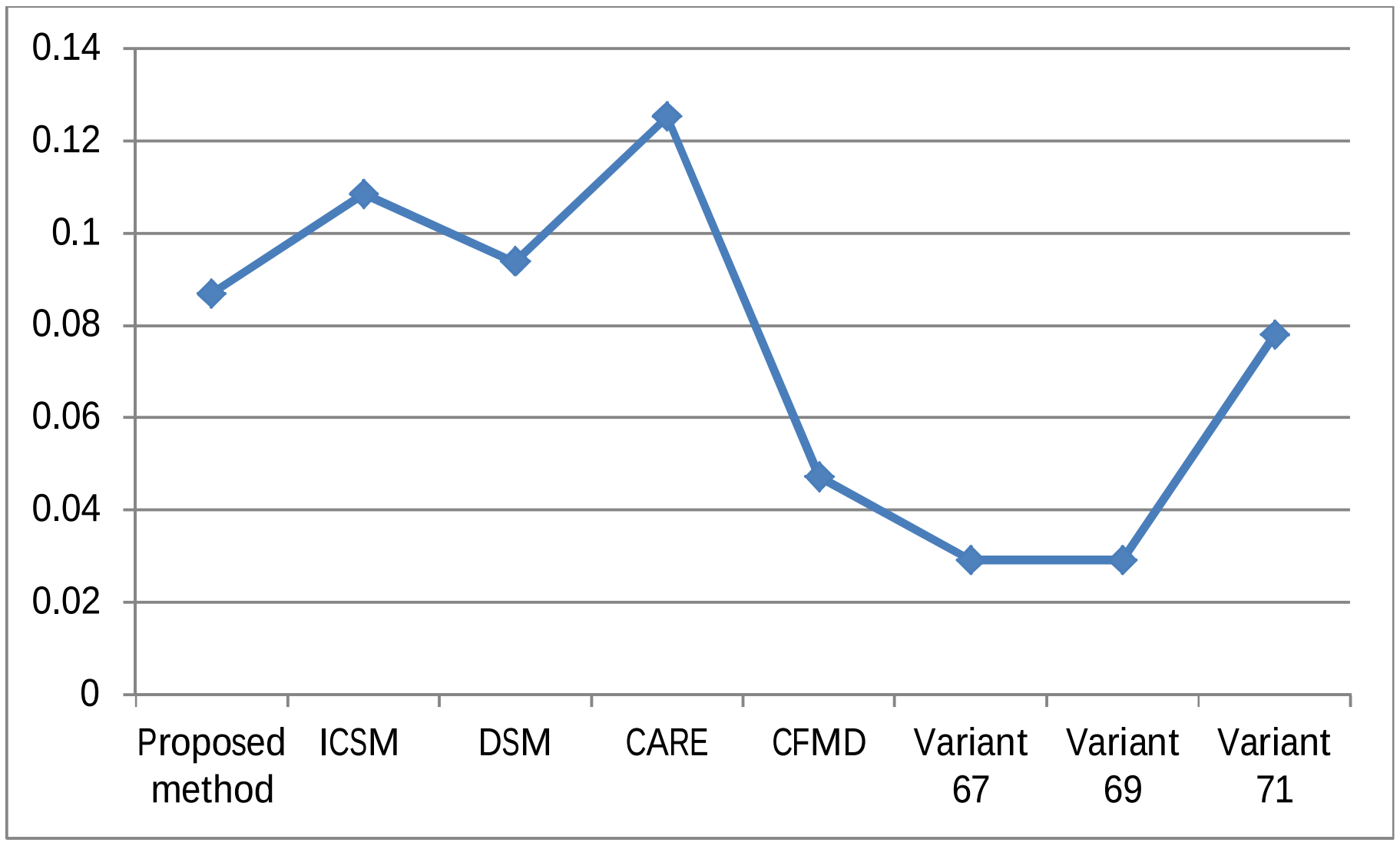

**Fig. 24. MSE of Diabetes dataset between 8 algorithms**

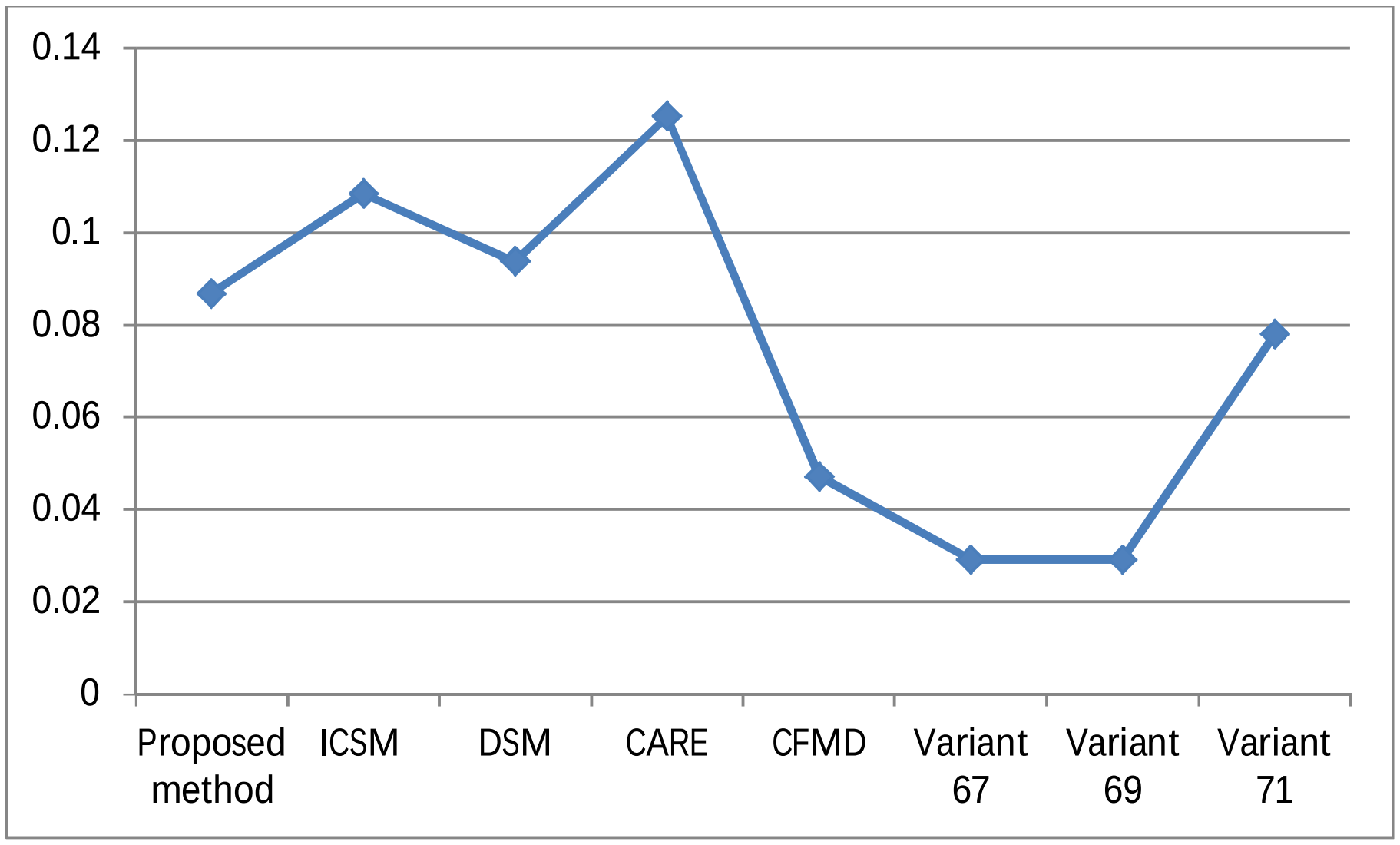

**Fig. 25. MSE of Breast dataset between 8 algorithms**

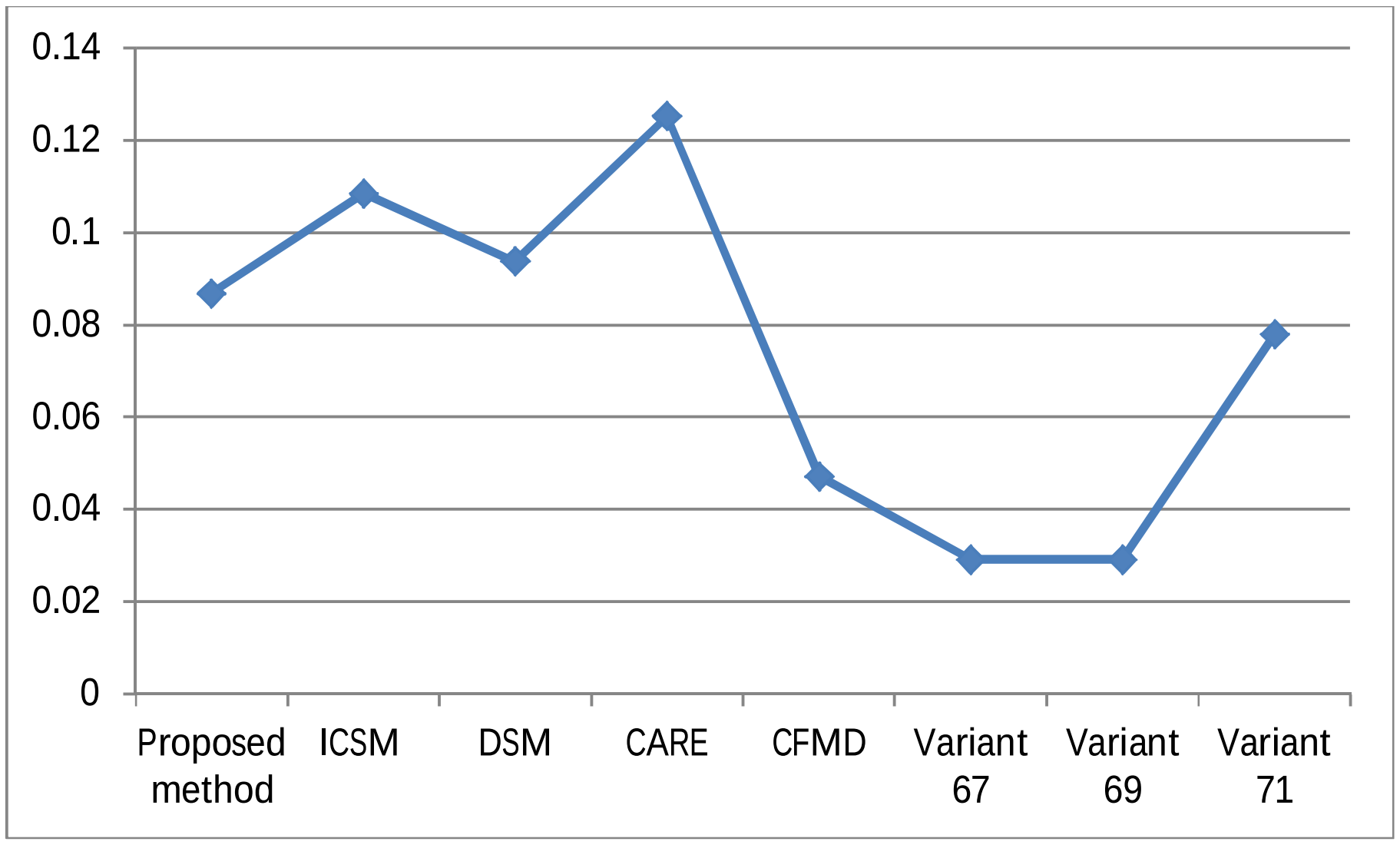

**Fig. 26. MSE of DMD dataset between 8 algorithms**

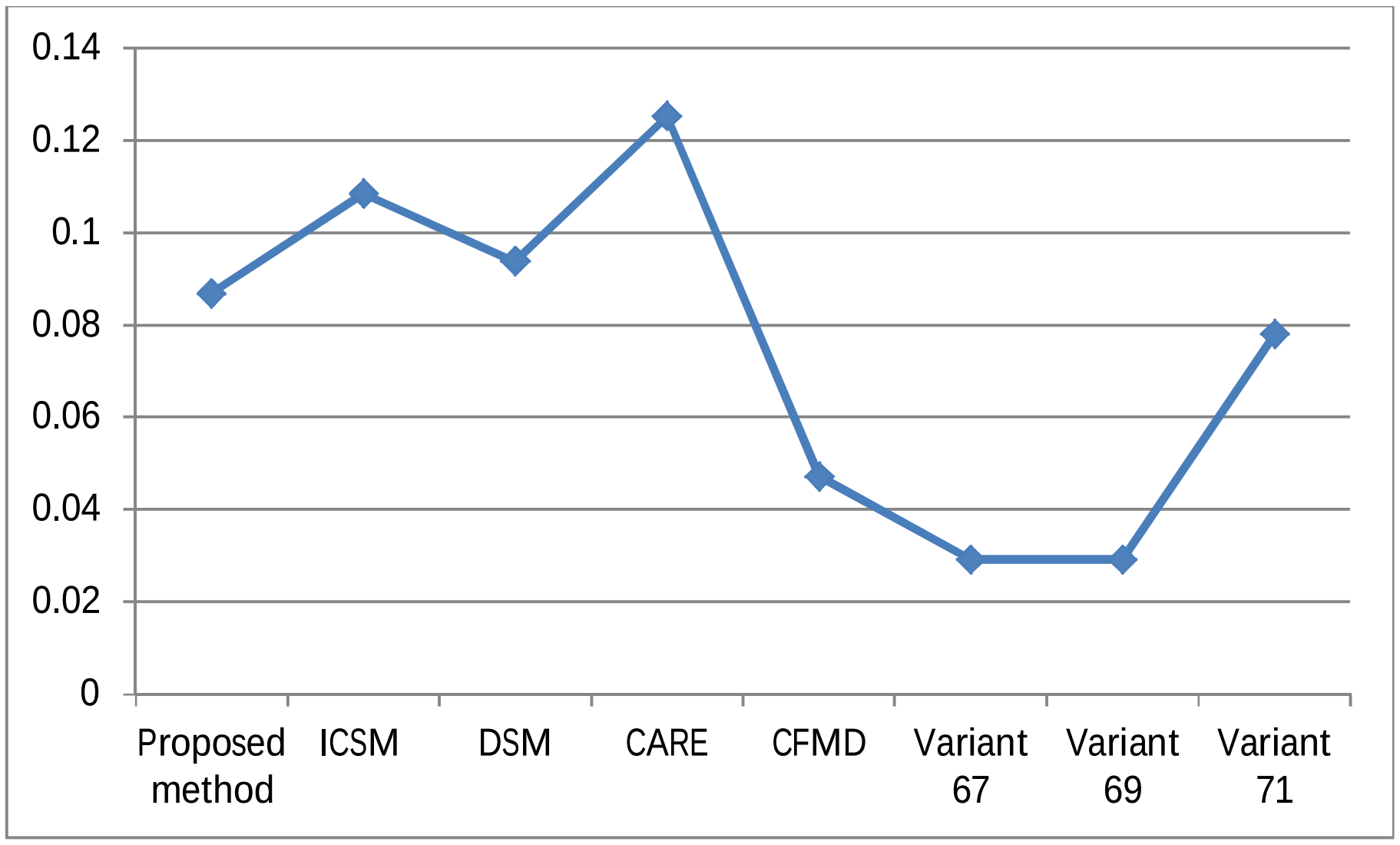

**Fig. 27. MSE of RHC dataset between 8 algorithms**

The MSE values of each algorithm have been plotted on a single dataset in each figure to check the deviation among them.

**Important Note:** Since our model has 36 numbers of parameters in neutrosophication process. Therefore, it is very difficult to present them in tabular form due to excessive number of pages. Thus, we have uploaded all the codes of our paper on the site in the Appendix.

Here, we only presented the deneutrosophication process in which we checked the result of MSE and computational time (Sec.) by changing the values of parameters $\alpha, \beta, \gamma$. Although the values of parameters are different but there is no much difference in the values of MSE and this confirming the stability of our proposed method.

**Table 15. The results of proposed method by parameters of deneutrosophication step**

| Dataset | Deneutrosophication parameters | | | | |
|---|---|---|---|---|---|
| A. Heart | | | | | |
| | $\alpha$ | $\beta$ | $\gamma$ | MSE | Time (sec) |
| | 0.2 | 0.3 | 0.5 | **0.236009** | **0.331406** |
| | 0.3 | 0.2 | 0.5 | **0.235996** | **0.231169** |
| | 0.5 | 0.3 | 0.2 | **0.236031** | **0.206800** |
| | 0.5 | 0.2 | 0.3 | **0.236010** | **0.212506** |
| | 0.3 | 0.5 | 0.2 | **0.236043** | **0.184897** |
| | 0.2 | 0.5 | 0.3 | **0.236036** | **0.196846** |
| **B. RHC** | | | | | |
| | 0.2 | 0.3 | 0.5 | **0.250000** | **71.887356** |
| | 0.3 | 0.2 | 0.5 | **0.250000** | **72.612241** |
| | 0.5 | 0.3 | 0.2 | **0.250000** | **73.629676** |
| | 0.5 | 0.2 | 0.3 | **0.250000** | **77.334187** |
| | 0.3 | 0.5 | 0.2 | **0.250000** | **74.391425** |
| | 0.2 | 0.5 | 0.3 | **0.250000** | **73.131208** |
| **C. Diabetes** | | | | | |
| | 0.2 | 0.3 | 0.5 | **0.055733** | **0.409223** |
| | 0.3 | 0.2 | 0.5 | **0.038069** | **0.402370** |
| | 0.5 | 0.3 | 0.2 | **0.030184** | **0.414937** |
| | 0.5 | 0.2 | 0.3 | **0.042295** | **0.421679** |
| | 0.3 | 0.5 | 0.2 | **0.042529** | **0.399683** |
| | 0.2 | 0.5 | 0.3 | **0.034698** | **0.391552** |
| **D. Breast** | | | | | |
| | 0.2 | 0.3 | 0.5 | **0.033555** | **35.283942** |
| | 0.3 | 0.2 | 0.5 | **0.033579** | **36.248034** |
| | 0.5 | 0.3 | 0.2 | **0.033472** | **36.425809** |
| | 0.5 | 0.2 | 0.3 | **0.033522** | **35.224594** |
| | 0.3 | 0.5 | 0.2 | **0.033464** | **35.228633** |
| | 0.2 | 0.5 | 0.3 | **0.033488** | **36.703062** |
| **E. DMD** | | | | | |
| | 0.2 | 0.3 | 0.5 | **0.250000** | **0.136135** |
| | 0.3 | 0.2 | 0.5 | **0.250000** | **0.126467** |
| | 0.5 | 0.3 | 0.2 | **0.250000** | **0.117475** |
| | 0.5 | 0.2 | 0.3 | **0.250000** | **0.132521** |
| | 0.3 | 0.5 | 0.2 | **0.250000** | **0.111520** |
| | 0.2 | 0.5 | 0.3 | **0.25000** | **0.127623** |
| | | | | | |

In the deneutrosophication process, the values of MSE are almost remains the same in each datasets by changing randomly the values of parameters $\alpha$, $\beta$ and $\gamma$ in all the 5 medical datasets. For example, the MSE values are approximately the same **(0.236009, 0.235996, 0.030184, 0.236010, 0.236043, 0.236036) respectively** in the Heart dataset by taking the values of $\alpha = 0.2, 0.3, 0.5, 0.5, 0.3, 0.2$, $\beta = 0.3, 0.2, 0.3, 0.2, 0.5, 0.5$ and $\gamma = 0.5, 0.5, 0.2, 0.3, 0.2, 0.3$. However, in the computational time (Sec.) some noticeable changes can be seen respectively **(0.331406, 0.231169, 0.206800, 0.212506, 74.391425, 0.196846).** The analogous scenario can be seen for the other 4 datasets of RHC, Diabetes, Breast and DMD where the values of MSE remains the same in each datasets while the computational time (Sec.) varies by changing the values of parameters which can be checked in the above **Table 15**.

### 4.3 Analyzing the strength of algorithms by ANOVA test

Next, we have tested all the 8 algorithms using ANOVA one-way test and Kruskal-Wallis test of variance by considering the MSE values among all 8 algorithms on the same dataset. This scenario can be seen in **Figs. 27, 28** and **Tables 16** and **17** respectively.

With regards to the ANOVA one-way test, in the **Fig. 27**, the blue bar represents the comparison interval for mean strength for our propose algorithm while the red bar represents the comparison interval for mean strength for the rest of 7 algorithms. There is no overlap between the blue bar and res bar which designate that the mean strength for proposed algorithm is significantly different from the rest of the algorithms.

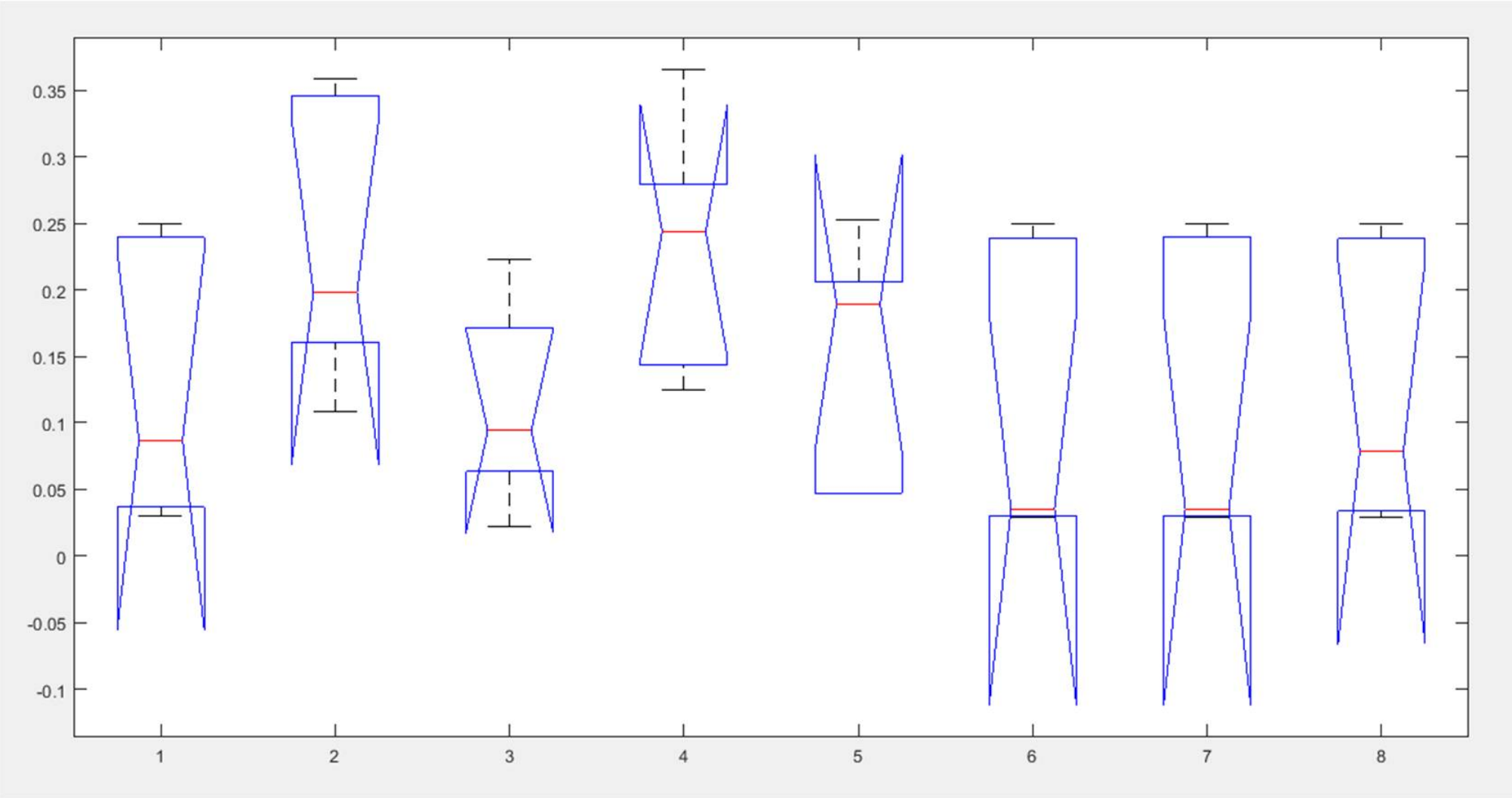

**Fig. 27. Analyze variance one- way for 8 algorithms**

The ANOVA **Table 16** demonstrates the between-groups variation (column) and within-groups variation (error). Here $df$ denote the total degrees of freedom which mean total number of observation minus 1. In this case we have $df = 8 - 1 = 7$. $SS$ is the sum of squares due to each source which is 0.09069 whereas, $MS$ represents the mean squared error which is equal to $MS = {}^{SS}\!/_{df} = 0.01296$. $F = F$ -statistic which is the ratio of mean square and we have 1.21 and finally Prob > F means the probability that the $F$ -statistic can take a value greater than the computed test-statistic value. In this case, we have the probability of 0.3252. **Table 16** summarizes all these results of ANOVA test.

**Table 16. Result of analyze variance one- way for 8 algorithms by ANOVA test**

| Source | SS | df | MS | F | Prob>F |
|---|---|---|---|---|---|
| **Columns** | 0.09069 | 7 | 0.01296 | 1.21 | 0.3252 |
| **Error** | 0.34232 | 32 | 0.0107 | | |
| **Total** | 0.43301 | 39 | | | |

The Kruskal-Wallis is a non-parametric test which is the classical version of one-way ANOVA test.

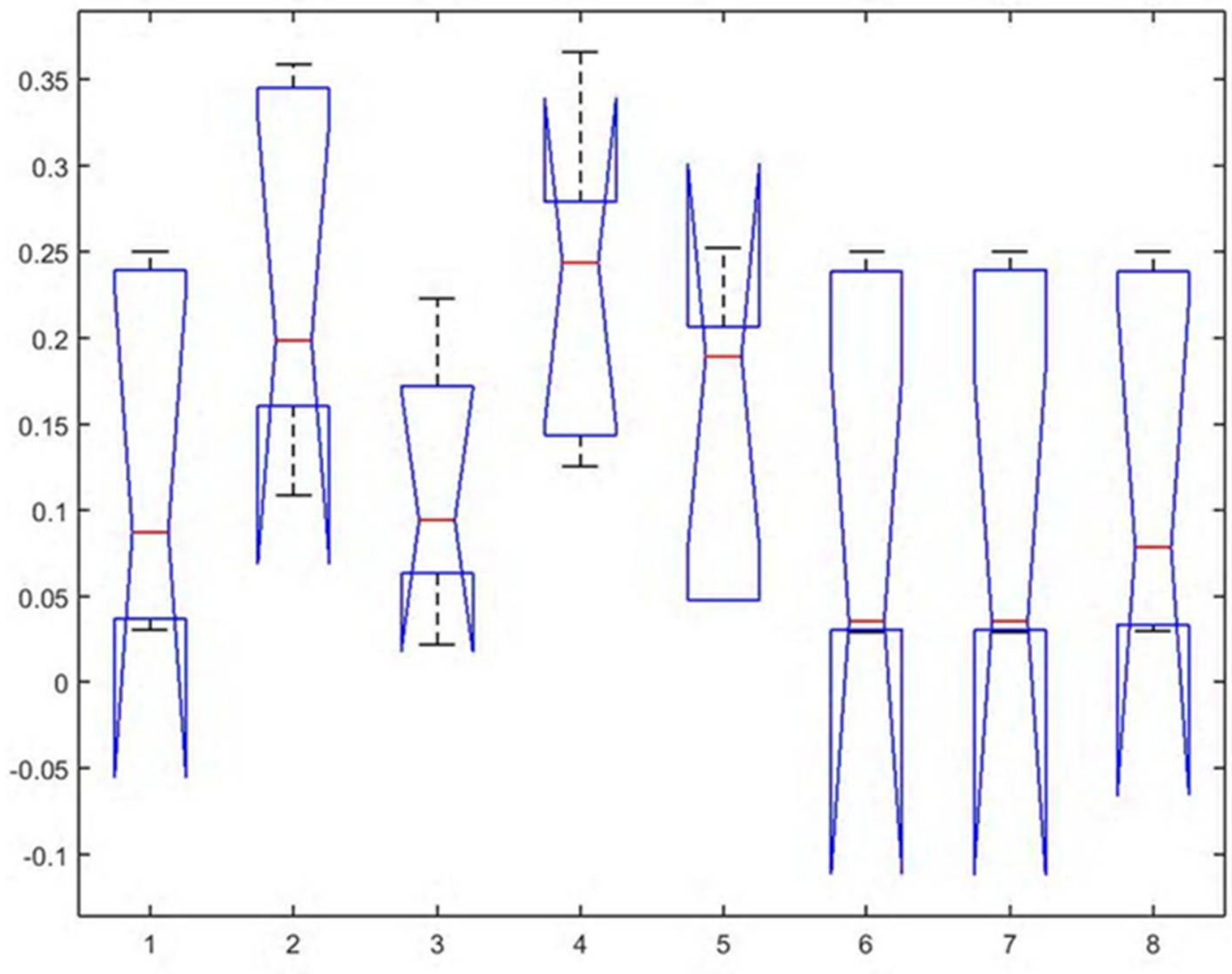

**Fig 28: Result of analyze Kruskal-Wallis test for 8 algorithms**

In the Kruskal-Wallis test, the $F$ -statistic value in ANOVA one-way test is replaced by Chi-square statistic. All the results of Kruskal-Wallis can be seen in the **Table 17** below to analyze the 8 algorithms by this test as well.

**Table 17. Result of analyze Kruskal-Wallis test for 8 algorithms**

| Source | SS | df | MS | Chi-sq | Prob>Chi-sq |
|---|---|---|---|---|---|
| **Columns** | 1052.5 | 7 | 150.357 | 1.71 | 0.3587 |
| **Error** | 4270.5 | 32 | 133.453 | | |
| **Total** | 5323 | 39 | | | |

## 5. Conclusion

This paper is dedicated to develop a novel neutrosophic recommender system based on neutrosophic set for medical diagnosis problems which has the ability to predict more accurately during diagnosis process. First, we presented a single-criteria neutrosophic recommender system (SC-NRS) and a multi-criteria neutrosophic recommender system (MC-NRS) and studied some of their basic properties. Further, we studied some algebraic operations of neutrosophic recommender system NRS such as union, complement, intersection etc. We also showed that these algebraic operations satisfy commutative, associative and distributive property. Based on these operations, we investigated the algebraic structures such as lattices, Kleen algebra, de Morgan algebra, Brouwerian algebra, BCK algebra, Stone algebra and MV algebra. In addition, we introduced several types of algebraic similarity measures which are basically based on these algebraic operations and studied some of their theoretic properties. Using the proposed similarity measure, we developed a predication formula. We proposed a new algorithm for medical diagnosis based on neutrosophic recommender system. Some interesting properties of the proposed methodology were also investigated. Finally to check the validity of the proposed methodology, we made experiments on the datasets Heart, RHC, Breast cancer, Diabetes and DMD. At the end, we presented the MSE and computational time (Sec.) by comparing the proposed algorithm with ICSM, DSM, CARE, CFMD, Variant 67, Variant 69, and Varian 71 both in tabular and graphical form to analyze the effectiveness and accuracy. Numerical examples have been given throughout in the paper. Finally we analyzed the strength of all 8 algorithms by ANOVA one-way test and Kruskal-Wallis test.

A numerical example has been presented on neutrosophic medical diagnosis data in the evaluation section. The experiments were conducted carefully on 5 benchmark medical datasets of both the small and large size. The benchmark dataset HEART has been taken from UCI Machine Learning Repository (University of California) and the remaining 4 benchmark datasets RHC (Right Heart Catheterization), Diabetes, Breast Cancer, DMD (Duchenne Muscular Dystrophy) have been taken from (Department of Biostatistics, Vanderbilt University). The proposed methodology has the ability to predict and recommend with several kinds of datasets including neutrosophic data than other standalone algorithms such as ICSM, DSM, CARE, CFMD and neutrosophic set. Our proposed method has the more accuracy than the other algorithms as well as can handle the limitation and drawbacks of the previous work. The Mean Square Error (MSE), computational time and the accuracy of diagnosis in the proposed methodology is worth of much attention and concentration which provides us a good evidence of the usefulness of the proposed algorithm in the paper.

Due to the significance and the importance of the proposed work, some more solid research can be conducted in this area which can extend the effectiveness of the proposed work to other fields such as time series forecasting. We are planning to develop the parameter estimation for neutrosophic recommender system by Bayesian approach and a multi-characteristic neutrosophic recommender system for medical diagnosis. A hybrid algorithm between our proposed algorithm and neutrosophic clustering method which can enhance the accuracy can be considered in the near future.

**Appendix**

Source codes and experimental datasets are found at:

http://se.mathworks.com/matlabcentral/fileexchange/55239-a-neutrosophic-recommender-system-for-medical-diagnosis-based-on-algebraic-neutrosophic-measures

**References**

[1] Akhtar, N., Agarwal, N., Burjwal, A., K-mean algorithm for Image Segmentation using Neutrosophy, International Conference on Advances in Computing, Communications and Informatics (ICACCI, 2014), 2417–2421, IEEE Xplore, **DOI:** 10.1109/ICACCI.2014.6968286

[2] Ali, M., Deli, I., and Smarandache, F., The Theory of Neutrosophic Cubic Sets and Their Application in Pattern Recognition, Journal of Intelligent and Fuzzy Systems, (In press), 1-7, DOI:10.3233/IFS-151906.

[3] Ali, M., and Smarandache, F., Complex Neutrosophic Set, Neural Computing and Applications,Vol. 25, (2016), 1-18. DOI: 10.1007/s00521-015-2154-y.

[4] Ansari, A., Q., Biswas, R., and Aggarwal, S., Proposal of Applicability of Neutrosophic Set Theory in Medical AI, International Journal of Computer Applications, Volume 27, No.5, August 2011, (0975 –8887).

[5] Broumi, S. and Deli, I., Correlation measure of neutrosophic refined sets and its applications in medical diagnosis, Palestine Journal of Mathematics, Vol. 3(1) (2014), 11–19.

[6] Broumi, S., and Smarandache, S., Extended Hausdorff Distance and Similarity measure of Refined Neutrosophic Sets and their Application in Medical Diagnosis, Journal of New Theory, Vol. 1, issue 7, (2015), 64-78.

[7] Broumi, S., Deli, I,. Smarandache, F,. N-Valued Interval Neutrosophic Sets and Their Application in Medical Diagnosis, Critical Review; 2015, Vol. 10, p45.

[8] Connors, F. A., et al., The effectiveness of right heart catheterization in the initial care of critically III patients, Jama 276 (11) (1996) 889–897.

[9] Cuong, B. C., Son, L. H., Chau, H. T. M., Some context fuzzy clustering methods for classification problems, in: Proceedings of the 2010 ACM Symposium on Information and Communication Technology, 2010, pp. 34–40.

[10] Davis, D. A., Chawla, N. V., Blumm,N., Christakis,N., Barabási,A. L., Predicting individual disease risk based on

medical history, in: Proceedings of the 17th ACM Conference on Information and Knowledge Management, 2008, pp. 769–778.

[11] Duan, L., Street, W. N., Xu, E., Healthcare information systems: data mining methods in the creation of a clinical recommender system, Enterprise Inform. Syst. 5 (2) (2011) 169–181.

[12] De, S. K., Biswas, R., Roy, A. R., An application of intuitionistic fuzzy sets in medical diagnosis, Fuzzy Sets Syst. 117 (2) (2001) 209–213.

[13] Deli, I., Ali, M., and Smarandache, F., Bipolar neutrosophic sets and their application based on multi-criteria decision making problems, 2015 International Conference on Advanced Mechatronic Systems (ICAMechS), (2015), 249 – 254, IEEE Xplore, DOI:10.1109/ICAMechS.2015.7287068

[14] Department of Biostatistics, Vanderbilt University. http://biostat.mc.vanderbilt.edu/DataSets

[15] Gaber, T., Zehran, G., Anter, A., and Vaclav, S., Thermogram breast cancer detection approach based on Neutrosophic sets and fuzzy c-means algorithm, Conference: 37th Annual International Conference of the IEEE Engineering in Medicine and Biology Society (EMBC'15), Milano, Italy.

[16] George J. K., and Bo Y., Fuzzy sets and fuzzy logic: Theory and applications, Prentice Hall, Upper Saddle River, New Jersey, 1995.

[17] Ghazanfar, M. A., & Prügel-Bennett, A. (2014). Leveraging clustering approaches to solve the gray-sheep users problem in recommender systems. Expert Systems with Applications, 41(7), 3261–3275.

[18] Guo, Y., Zhou, C., Chan, H. P,. Aamer Chughtai, Jun Wei, Lubomir M. Hadjiiski, and Ella A. Kazerooni, Automated iterative neutrosophic lung segmentation for image analysis in thoracic computed tomography, Med Phys. 2013 Aug; 40(8), doi: 10.1118/1.4812679

[19] Guerram, I., Maamri, R., Sahnoun, Z., and Merazga, S., Qualitative Modelling of Complex Systems by Neutrosophic Cognitive Maps: Application to the Viral Infection, itpapers.info/acit10/Papers/f710.pdf

[20] Hanafy, M., Salama, A., A., and Mahfouz, K., M., Correlation of neutrosophic Data, International Refereed Journal of Engineering and Science, 1(2) (2012) 39-43.

[21] Hanafy, M., Salama, A., A., and Mahfouz, K., M., Correlation Coefficients of Neutrosophic Sets by Centroid Method, International Journal of Probability and Statistics, 2(1) (2013) 9-12.

[22] Hassan, S., Syed, Z., From netflix to heart attacks: collaborative filtering in medical datasets, in Proceedings of the 1st ACM International Health Informatics Symposium, 2010, pp. 128–134.

[23] Johnson, P. E.; Duran, A. S.; Hassebrock, F.; Moller, J.; Prietula, M.; Feltovich, P. J.; Swanson, D. B."Expertise and Error in Diagnostic Reasoning".Cognitive Science **5**(3): 235-283, (1981). doi:10.1207/s15516709cog0503_3

[24] Kala, R., Janghel, R. R., Tiwari, R., Shukla, A., Diagnosis of breast cancer by modular evolutionary neural networks, Int. J. Biomed. Eng. Technol. 7 (2) (2011) 194–211.

[25] Kandasamy, W., B. V., and Smarandache, F., Fuzzy Relational Maps and Neutrosophic Relational Maps, Hexis Church Rock, (2004), ISBN: 1-931233-86-1.

[26] Kharal, A., An application of neutrosophic sets in medical diagnosis, Critical Review, Vol. VII, (2014), 3-15.

[27] Kononenko,Y., Machine learning for medical diagnosis: history, state of the art and perspective, Artif. Intell. Med. 23 (1) (2001) 89–109.

[28] Mathew, J., M**., and** Simon, P., Color Texture Image Segmentation Based on Neutrosophic Set and Nonsubsampled Contourlet Transformation, Volume 8321 of the series Lecture Notes in Computer Science, (2014), 164-173, DOI:10.1007/978-3-319-04126-1_14

[29] Mahdavi, M. M., Implementation of a recommender system on medical recognition and treatment, Int. J. e-Education, e-Business, e-Management e-Learning 2 (4) (2012) 315–318.

[30] Moein, S., Monadjemi, S. A., Moallem, P., A novel fuzzy-neural based medical diagnosis system, Int. J. Biol. Med. Sci. 4 (3) (2009) 146–150.

[31] Mohan, J., Krishnaveni, V., Guo, Y., A New Neutrosophic Approach of Wiener Filtering for MRI Denoising, MEASUREMENT SCIENCE REVIEW, Volume 13, No. 4, 2013.

[32] Neog, T. J., Sut, D. K., An application of fuzzy soft sets in medical diagnosis using fuzzy soft complement, Int. J. Comput. Appl. 33 (2011) 30–33.

[33] Own, C. M., Switching between type-2 fuzzy sets and intuitionistic fuzzy sets: an application in medical diagnosis, Appl. Intell. 31 (3) (2009) 283–291.

[34] Parthiban, L., Subramanian, R., Intelligent heart disease prediction system using CANFIS and genetic algorithm, Int. J. Biol. Biomed. Med. Sci. 3 (3) (2008) 157–160.

[35] Pramanik, S., and Mondal, K., Cosine Similarity Measure of Rough Neutrosophic Sets and Its Application in Medical Diagnosis, Global Journal of Advanced Research 01/2015; 2 (1), 212-220.

[36] Palanivel, K., & Siavkumar, R. (2010). Fuzzy multi-criteria decision-making approach for collaborative recommender systems. International Journal of Computer Theory and Engineering, 2(1), 57–63.

[37] Peng J., J., Wang J., Q., Wang J., Zhang H., Y., and Chena X., H., Simplified neutrosophic sets and their applications in multi-criteria group decision-making problems, International Journal of Systems Science, DOI:10.1080/00207721.2014.994050

[38] Ricci, F., Rokach, L., Shapira, B., Introduction to Recommender Systems Handbook, Springer, US, 2011, 1–35.

[39] Samuel, A. E., Balamurugan, M., Fuzzy max–min composition technique in medical diagnosis, Appl. Math. Sci. 6 (35) (2012) 1741–1746.

[40] Sanchez, E., Resolution of composition fuzzy relation equations, Inform. Control, 30 (1976) 38–48.

[41] Shinoj, T. K., John, S. J., Intuitionistic Fuzzy Multi sets and its Application in Medical Diagnosis, World Acad. Sci. Eng. Technol. 6 (2012) 1418–1421.

[42] Smarandache, F., A Unifying Field in Logics. Neutrosophy: Neutrosophic Probability, Set and Logic, Rehoboth: American Research Press, 1998.

[43] Son, L. H., Lanzi, P. L., Cuong, B. C., Hung, H. A., Data mining in GIS: a novel contextbased fuzzy geographically weighted clustering algorithm, Int. J. Machine Learning Comput. 2 (3) (2012), 235–238.

[44] Son, L. H., Linh, N. D., Long, H. V., A lossless DEM compression for fast retrieval method using fuzzy clustering and MANFIS neural network, Eng. Appl. Artif. Intell. 29 (2014) 33–42.

[45] Son, L. H., DPFCM: a novel distributed picture fuzzy clustering method on picture fuzzy sets, Expert Syst. Appl. 42 (1) (2015) 51–66.

[46] Son, L. H., Cuong, B. C., Lanzi, P. L., Thong, N. T., A novel intuitionistic fuzzy clustering method for geo-demographic analysis, Expert Syst. Appl. 39 (10) (2012) 9848–9859.

[47] Son, L. H., Cuong, B. C., Long, H. V., Spatial interaction–modification model and applications to geo-demographic analysis, Knowl.-Based Syst. 49 (2013) 152–170.

[48] Szmidt, E., Kacprzyk, J., Intuitionistic fuzzy sets in some medical applications, in: Proceeding of Computational Intelligence: Theory and Applications, 2001, pp. 148–151.

[49] Szmidt, E., Kacprzyk, J., An intuitionistic fuzzy set based approach to intelligent data analysis: an application to medical diagnosis, in: Proceeding of Recent Advances in Intelligent Paradigms an Applications, 2003, pp. 57–70.

[50] Szmidt, E., Kacprzyk, J., A similarity measure for intuitionistic fuzzy sets and its application in supporting medical diagnostic reasoning, in Proceeding of Artificial Intelligence and Soft Computing (ICAISC 2004). 388-393.

[51] Tan, K. C., Yu, Q., Heng, C. M., Lee, T. H., Evolutionary computing for knowledge discovery in medical

diagnosis, Artif. Intell. Med. 27 (2) (2003) 129–154.

[52] Treasure, W, (2011). "Chapter 1: Diagnosis". Diagnosis and Risk Management in Primary Care: Oxford: Radcliffe. ISBN 978-1-84619-477-1

[53] University of California, UCI Repository of Machine Learning Databases, 2007. <http://archive.ics.uci.edu/ml/>.

[54] Wang, H., Smarandache , F., Sunderraman R., Zhang, Y., Q., Interval Neutrosophic Set and Logic, Theory and Applications in Computing, Hexis Arizona, (2005), ISBN: 1-931233-94-2.

[55] Wang H., Smarandache F., Zhang Y., Q., Sunderraman R. Single valued neutrosophic sets. Multispace Multistructure, 2010;4:410–3.

[56] Xiao, Z., Yang, X., Niu, Q., Dong, Y., Gong, K., Xia, S., Pang, Y., A new evaluation method based on D–S generalized fuzzy soft sets and its application in medical diagnosis problem, Appl. Math. Modell. 36 (10) (2012) 4592–4604.

[57] Yager, R.: Fuzzy Logic Methods in Recommender Systems. Fuzzy Sets and Systems, 136, 133–149, (2003).

[58] Ye, J., Multicriteria decision-making method using the correlation coefficient under single-valued neutrosophic environment, International Journal of General Systems, 42(4) (2013) 386-394.

[59] Ye, J., Improved cosine similarity measures of simplified neutrosophic sets for medical diagnoses, Artificial Intelligence in Medicine 63 (2015) 171–179, http://dx.doi.org/10.1016/j.artmed.2014.12.007.

[60] Ye, S., Fu, J., and Ye, J., Medical Diagnosis Using Distance-Based Similarity Measures of Single Valued Neutrosophic Multisets, Neutrosophic Sets and Systems, Vol. 07, (2015), 47-54.

[61] Ye, S., Ye, J., Dice Similarity Measure between Single Valued Neutrosophic Multisets and Its Application in Medical Diagnosis, Neutrosophic Sets and Systems, 6, 48–53, 2014.

[62] Ye, J., and Fu, J., Multi-period medical diagnosis method using a single valued neutrosophic similarity measure based on tangent function, Computer Methods and Programs in Biomedicine, Available online 14 October 2015, doi:10.1016/j.cmpb.2015.10.002

[63] Ye J., Vector similarity measures of simplified neutrosophic sets and their application in multi-criteria decision making. Int J Fuzzy Syst 2014;16(2):204–11.

[64] Ye, J., Trapezoidal neutrosophic set and its application to multiple attribute decision-making, Neural Computing and Applications, 2015, Volume 26, Issue 5, pp 1157-1166

[65] Zhang, H., Wang, J., Q., and Chen, X., H, Interval Neutrosophic Sets and Their Application in Multicriteria

Decision Making Problems, The Scientific World Journal, Volume 2014 (2014), Article ID 645953, 15 pages, http://dx.doi.org/10.1155/2014/645953

[66] Zhang, M., Zhang, L., Cheng, H., Segmentation of ultrasound breast images based on a neutrosophic method, Opt. Eng. 49(11) 117001, (2010), doi: 10.1117/1.3505

[67] Park, D. H., Kim, H. K., Choi, I. Y., Kim, J. K., A literature review and classification of recommender systems research, Expert Systems with Applications, 39 (2012) 10059-10072